\documentclass[lettersize,journal]{IEEEtran}
\usepackage{amsmath,amsfonts}
\usepackage{algorithmic}
\usepackage{algorithm}
\usepackage{array}
\ifCLASSOPTIONcompsoc
\usepackage[caption=false, font=normalsize, labelfont=sf, textfont=sf]{subfig}
\else
\usepackage[caption=false, font=footnotesize]{subfig}
\fi
\usepackage{textcomp}
\usepackage{stfloats}
\usepackage{url}
\usepackage{verbatim}
\usepackage{graphicx}
\usepackage{epstopdf}
\usepackage{cite}
\usepackage{color}
\usepackage{amssymb}
\usepackage{multirow}
\usepackage{bbding}
\usepackage{makecell}
\usepackage[T1]{fontenc}

\newcommand{\tablesize}{\fontsize{7.5pt}{12pt}\selectfont}

\def\Wcal{\mathcal{W}}

\def\E{\mathbf{E}}

\def\K{\mathbf{K}}

\def\M{\mathbf{M}}

\def\T{\mathbf{T}}
\def\U{\mathbf{U}}
\def\V{\mathbf{V}}
\def\W{\mathbf{W}}
\def\X{\mathbf{X}}
\def\Y{\mathbf{Y}}
\def\Z{\mathbf{Z}}

\def\u{\mathbf{u}}
\def\v{\mathbf{v}}

\def\cited#1{{\iffalse #1 \fi}}
\def\1#1{\textcolor{red}{\textbf{#1}}}
\def\2#1{\textcolor{blue}{\textbf{#1}}}
\def\3#1{\textcolor{green}{\textbf{#1}}}

\begin{document}


\makeatother

\title{Iterative Low-rank  Network for Hyperspectral Image Denoising}

\author{Jin Ye, Fengchao Xiong,~\IEEEmembership{Member,~IEEE},  Jun Zhou,~\IEEEmembership{Senior Member,~IEEE} and Yuntao Qian,~\IEEEmembership{Senior Member,~IEEE}
\thanks {This work was supported in part by  the National Natural Science Foundation of China under Grant 62371237 and the Fundamental Research Funds for the Central Universities under Grant 30923010213. (Corresponding author: Fengchao Xiong.)}
\thanks{Jin Ye and Fengchao Xiong are with the School of Computer Science and Engineering, Nanjing University of Science and Technology, Nanjing 210094, China. }
\thanks{Jun Zhou is with the School of Information and Communication Technology, Griffith University, Nathan, Australia.}
\thanks{Yuntao Qian  is with the College of Computer Science, Zhejiang University, Hangzhou 310027, China.}
}
\markboth{Journal of \LaTeX\ Class Files,~Vol.~14, No.~8, February~2024}%
{Shell \MakeLowercase{\textit{et al.}}: A Sample Article Using IEEEtran.cls for IEEE Journals}


\maketitle

\begin{abstract}
        Hyperspectral image (HSI) denoising is a crucial preprocessing step for subsequent tasks. The clean HSI usually reside in a low-dimensional subspace, which can be captured by low-rank and sparse representation, known as the physical prior of HSI. It is generally challenging to adequately use such physical properties for effective denoising while preserving image details. This paper introduces a novel iterative low-rank network (ILRNet) to address these challenges. ILRNet integrates the strengths of model-driven and data-driven approaches by embedding a rank minimization module (RMM) within a U-Net architecture. This module transforms feature maps into the wavelet domain and applies singular value thresholding (SVT) to the low-frequency components during the forward pass, leveraging the spectral low-rankness  of HSIs in the feature domain. The parameter, closely related to the hyperparameter of the singular vector thresholding algorithm, is adaptively learned from the data, allowing for flexible and effective capture of low-rankness across different scenarios. Additionally, ILRNet features an iterative refinement process that adaptively combines intermediate denoised HSIs with noisy inputs.  This  manner  ensures progressive enhancement and superior preservation of image details. Experimental results demonstrate that ILRNet achieves state-of-the-art performance in both synthetic and real-world noise removal tasks.

\end{abstract}


\section{Introduction}
Hyperspectral images (HSIs) offer rich spectral information about the observed scene, enabling their use in various fields such as medical diagnosis~\cite{Lu2014MedicalHI}, face recognition~\cite{Uzair2015PLS}, and remote sensing~\cite{Li2024TASAM,Wang2024DirectNet}. However, HSIs are often affected by noise due to factors like imaging, illumination, and weather conditions during the data capture process. This noise can degrade the quality of the HSIs and subsequently hinder their effectiveness in downstream applications. Therefore, denoising HSIs is a critical preprocessing step to maintain their practical value.

Model-driven and data-driven methods represent the two primary approaches for HSI denoising. Model-driven methods leverage the intrinsic physical properties of HSIs for noise reduction. Since HSIs are high-dimensional data, clean HSIs typically reside in a low-dimensional subspace, which can be modeled by low-rank representations~\cite{Zhang2014LRMR,Chang2017LLRT,He2018LLRSSTV,Zheng2020LRTF-DFR}. Strategies such as low-rank matrix/tensor factorization~\cite{Wang2018LRTDTV,He2022NGMeet,Zheng2020LRTF-DFR} and matrix/tensor rank minimization~\cite{Xie2022ARSSC,Chen2022WNLRATV,Chang2017LLRT} are commonly employed to investigate the low-rank structure. Low-rank decomposition involves factorizing the target matrix or tensor into flat factors whose size requires to be preset. On the other hand, rank minimization aims to reconstruct the data by imposing an additional rank constraint  via different norms  such as nuclear norm on the estimated matrices or tensors.  Accordingly, the rank is not preset but estimated from the data. Recently, tensor singular value decomposition (TSVD), defined by the number of non-zero tubes in the frontal slice of the tensor under a predefined discrete Fourier transform (DFT) along the third mode, has been successfully applied to multidimensional data recovery, such as HSIs and videos~\cite{Kernfeld,Zhou}. TSVD can be seen as low-rank representation in the transformed domain. Consequently, many works have introduced new transforms to TSVD, including discrete cosine transform (DCT) and framelet transform~\cite{Jiang}. However, most of these transforms are analytic rather than learned, lacking adaptiveness to the data.  Additionally, these low-rank methods offer clear interpretability and theoretical guarantees, yet setting hyperparameters requires substantial domain expertise. Additionally, their performance can degrade if the noise type does not align with the hand-crafted priors, and the extensive computations  can limit their practical usage.

Benefiting from the powerful representation capability of deep neural networks (DNNs), data-driven methods can utilize abundant data to learn the mapping from noisy HSI to clean HSI, without the need for manually set priors. The strong nonlinear representation abilities of convolutional neural networks (CNNs)~\cite{Zhang2017DnCNN,Chang2019HSI-DeNet,Yuan2019HSID-CNN,Wei2021QRNN3D} and the  global dependencies modeling capacities of Transformers~\cite{Dixit2023UNFOLD,Li2023SERT} provide data-driven methods with a significant advantage.  However, compared to model-driven methods, data-driven methods often operate as a "black box" with limited interpretability. In response, there is a trend toward combining both approaches to leverage the hybrid advantages such as improved performance and promising interpretability~\cite{Zhang2022SLRP-DSP,Chen2024Flex-DLD,bodrito2021T3SC,Xiong2022SMDS-Net,Xiong2022MAC-Net}. Many methods incorporate sparsity priors through deep unfolding, but few explore the use of low-rank characteristics. From low-rank matrix factorization perspective, Zhang~\emph{et al.}~\cite{Zhang2021LR-Net} introduced a low-rank module to capture the latent low-rank semantic relationships in HSIs for recovery. However, their approach relies on a manually preset rank rather than learning it from data, limiting adaptiveness.


Alternatively, the low-rank structure can be enforced within DNNs by imposing a rank minimization constraint. This approach involves the network performing singular value decomposition (SVD) on extracted feature maps and learning singular value thresholding (SVT) parameters directly from the data to achieve low-rankness. However, integrating SVD into DNNs is challenging. One major issue arises from the backpropagation formula for SVD, where the gradient explosion can occur if two singular values are very close, potentially destabilizing training and hindering convergence~\cite{Ionescu2015Matrix}. In practice, heuristic approaches such as approximate eigenvector computation~\cite{Cho2019GDWCT}, decorrelated batch normalization~\cite{Huang_2018_CVPR}, and gradient clipping are often employed to mitigate these challenges but can result in reduced performance. Power iteration (PI)~\cite{Nakatsukasa2013Stable}, which relies on an iterative procedure to approximate the dominant eigenvector of a matrix, can also face problems such as error accumulation and increased computational costs. Recently, Wang~\emph{et al.}~\cite{Wang2022Robust} established the relationships between the Taylor expansion of the SVD gradient and the gradient obtained using PI, allowing SVD to be integrated into DNNs accurately and quickly without an iterative process over the eigenvector. This advancement enables DNNs to achieve stable and precise performance in tasks that benefit from low-rankness.

Motivated by the success of low-rank DNNs, this paper introduces a novel iterative low-rank network (ILRNet) for HSI denoising.  One essential component of ILRNet is the low-rank U-Net~\cite{ronneberger2015unet} which includes an embedded rank minimization module (RMM). This module transforms feature maps into the wavelet domain and applies SVT to the low-frequency components during the forward pass, exploiting the spectral low-rankness of HSIs.  The thresholding parameters are adaptively learned from data, providing flexible handling of low-rankness in real-world scenarios. Additionally, ILRNet integrates an iterative refinement process that progressively recovers clean HSIs and enhances image detail preservation. As illustrated in Fig.~\ref{fig:ilrnet}, ILRNet consists of three primary components: a coarse estimation module, a multi-stage refinement module, and the ${\Lambda}({\cdot})$ module. The coarse estimation module employs the low-rank U-Net. The multi-stage refinement module performs multiple iterations, each involving two steps: initially, the original noisy HSI is combined with the estimation from the previous iteration to include more details. Given the reduced noise level in the current HSI compared to the original, the refinement process leverages a lightweight network. Then, the refined HSI is combined with the previous estimation to  produce the outcome in the current iteration. The weights in the refinement process are adaptively determined by the ${\Lambda}({\cdot})$ module. To evaluate the effectiveness of ILRNet, we compare it against both model-driven and data-driven methods across synthetic and real-world noise removal experiments. The experimental results demonstrate that ILRNet achieves state-of-the-art performance in HSI denoising, showcasing its robustness and efficacy.

The remainder of this paper are as follows: Section~\ref{related_work}
systematically reviews recent advances of  HSI denoising. Section~\ref{method} details the introduced ILRNet. Section~\ref{experiment} demonstrates the outstanding performance of ILRNet. This paper concludes in Section~\ref{conc}.

\section{Related Work}\label{related_work}


\subsection{Model-driven Methods for HSI Denoising}

Traditionally, model-driven methods have been used to leverage the physical properties of HSIs for noise removal. These methods commonly employ approaches such as total variation (TV) \cite{Yuan2012SSAHTV,Peng2020E-3DTV,Chen2022WNLRATV}, low-rank modeling \cite{Zhang2014LRMR,Chang2017LLRT,He2018LLRSSTV,Zheng2020LRTF-DFR}, and sparsity-based methods \cite{Xie2020NonRLRS,Zha2023LGSR} that are rooted in physical properties. TV regularization reduces noise by minimizing abrupt changes in grayscale levels between regions of the image. For example, Wang \emph{et al.}~\cite{Wang2018LRTDTV} applied TV regularization to the spatial-spectral gradients of HSIs to achieve piecewise-smooth denoising. The inherent sparsity of HSIs allows them to be represented effectively with fewer atoms. Peng~\emph{et al.} \cite{Peng2014Decomposable} employed sparse representation methods  using a small number of atoms from a predefined or learned dictionary to represent clean HSIs for denoising. Ye~\emph{et al.} \cite{Ye2015MTSNMF} enforced different band images to share the same sparse coefficients to achieve spatial denoising of HSIs, while Maggioni ~\emph{et al.} \cite{Maggioni2013BM4D} modeled similarity between non-local spatial-spectral cubes based on joint sparse representation. In addition, the global spectral correlation and local or non-local spatial correlation of HSIs have led to successful low-rank modeling in HSI denoising. Zheng~\emph{et al.} \cite{Zheng20203DlogTNN} proposed a rank minimization model and solved it using the alternating direction method of multipliers (ADMM) algorithm. Sarkar \emph{et al.}~\cite{Sarkar2021Superpatch} developed a recovery algorithm for HSIs based on their low-rank characteristics, incorporating a fidelity term measured by structural similarity index and a regularization term measured by nuclear norm. He~\emph{et al.} \cite{Xie2022ARSSC} utilized low-rank Tucker decomposition and matrix minimization methods to capture global spectral correlation and non-local spatial adaptivity, respectively. Xie~\emph{et al.} \cite{Zheng20203DlogTNN} introduced two convex regularizers, rank correction (RC) and structured sparse correction (SSC), to achieve a close approximation of the clean low-rank structure of HSIs. Despite their physical interpretability, model-driven methods often face challenges such as high computational cost and complex parameter tuning.

\subsection{Data-driven Methods for HSI Denoising}

Data-driven methods leverage powerful representation capabilities to learn the non-linear mapping between noisy and clean HSIs directly from data, eliminating the need for explicit priors. These methods have significantly advanced the state-of-the-art in HSI denoising~\cite{Yuan2019HSID-CNN,Maffei2020HSI-SDeCNN}. For instance, 3D convolutions have been utilized to capture spatial-spectral correlations in HSIs~\cite{Zhang2019SSGN}. Liu~\emph{et al.}~\cite{LiuDADCNN} extracted feature maps along spatial and spectral dimensions to expand the effective receptive field and capture more contextual information. Wei~\emph{et al.}~\cite{Wei2021QRNN3D} treated HSIs as sequential data and used a quasi-recurrent pooling module to capture global spectral correlations. Moreover, attention mechanisms have been employed to model non-local spatial correlations in HSIs. For example, Ma~\emph{et al.}~\cite{Ma2020ENCAM} applied channel attention to enhance the extraction of multi-scale features, while Shi~\emph{et al.}~\cite{Shi20213DADNet} utilized position attention to exploit spatial information and channel attention for spectral information. Fu~\emph{et al.}~\cite{Fu2022NSSNN} combined CNNs and attention mechanisms to extract global spectral correlations and non-local spatial features, and Wang~\emph{et al.}~\cite{Pang2022TRQ3DNet} employed window-based multi-head self-attention to leverage global features. Hu~\emph{et al.} \cite{HuHCANet} combined convolution and attention mechanism to model both local and global features of HSI simultaneously. Recently, a trend towards integrating model-driven and data-driven methods has emerged to enhance the interpretability of HSI restoration~\cite{bodrito2021T3SC,meng2020gapnet,cai2022degradation,Li2023PADUT,Ma2022US3RN,Liu2022SPFNet,liu2023efficient,Xiongdpnet}.  Qiu~\emph{et al.}~\cite{Qiu2021TVFFDNet} incorporated TV priors into DNNs to facilitate the capture of underlying piecewise smooth structures.  Xiong~\emph{et al.}~\cite{Xiong2022SMDS-Net} converted the optimization of  a multi-dimensional sparse model to capture the low-rankness and sparsity of HSIs. Meanwhile, Zeng \emph{et al.}~\cite{zeng2023DNA-Net} unfolded a solution based on the half-quadratic splitting algorithm, utilizing Transformers as denoisers at each stage to simultaneously capture spectral correlations and local/non-local dependencies. Zhuang~\emph{et al.} \cite{Zhuangeigen} utilized the spectral low-rank property and spatial correlation of HSI through eigendecomposition and convolutional neural network (CNN), respectively. Tan~\emph{et al.} \cite{TanHyLoRa} used low-rank prompts for transformers to more accurately represent spatial-spectral correlations in the feature space. Following this trend, we incorporate the low-rank representation into DNNs for enhanced denoising.


\begin{figure*}[th]
        \centering
        \subfloat{\includegraphics[width=1\linewidth]{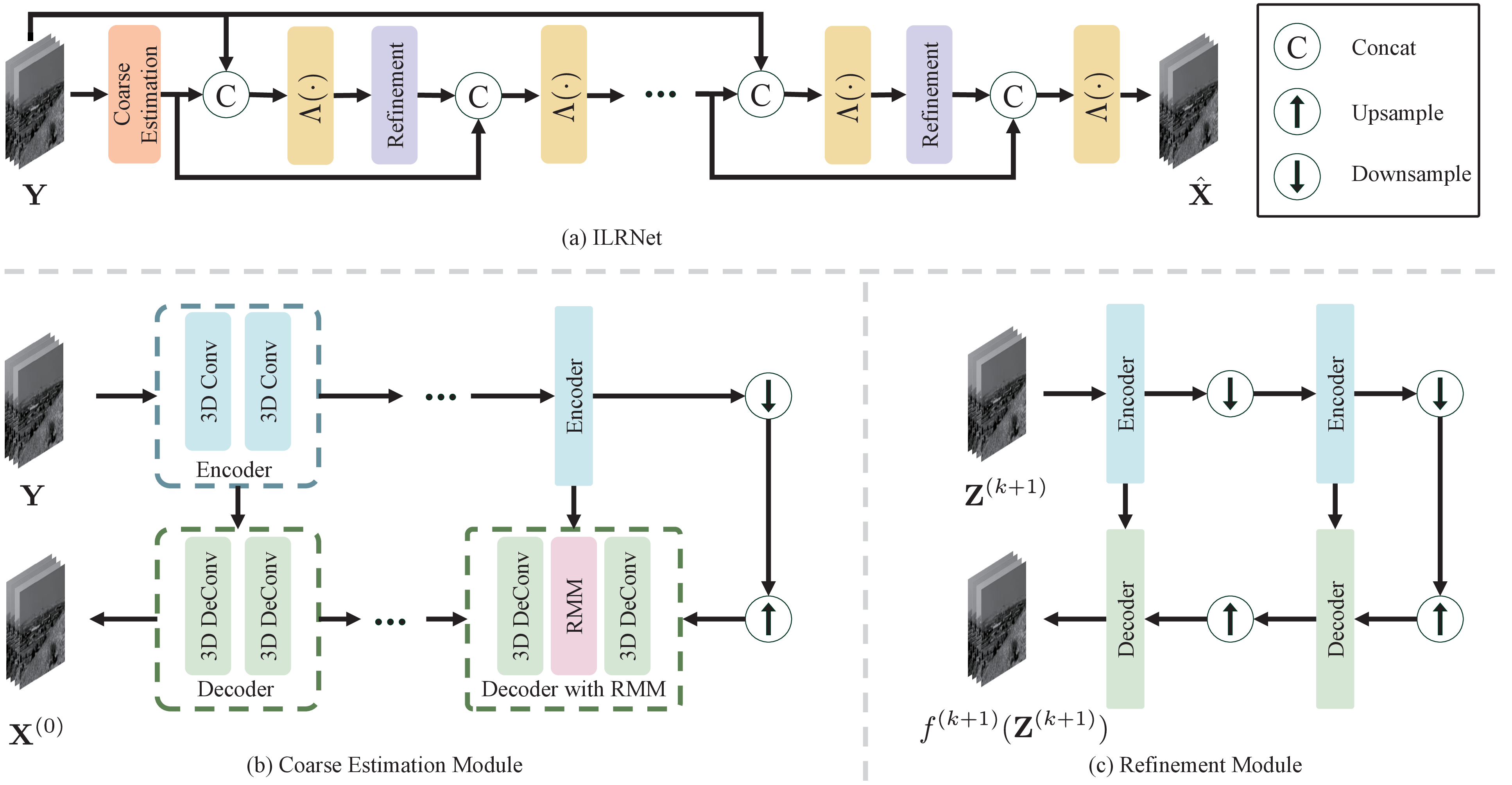}}
        \caption{The overall architecture of ILRNet, featuring a progressive refinement design. It encompasses: (1) a coarse estimation module to initialize the denoising process with a low-rank constraint, (2) a refinement module iteratively improving the estimation of the clean HSI, and (3) the $\Lambda(\cdot)$ module, which adaptively generates weights for adaptive linear combinations.}\label{fig:ilrnet}
\end{figure*}

\section{Method}\label{method}

This section provides a detailed  description of the ILRNet and  the design motivations behind its core modules: the coarse estimation module, multi-stage refinement module, and the ${\Lambda}({\cdot})$ module.
\subsection{Problem Formulation}

Let $\Y \in \mathbb{R}^{N \times B} $ be a HSI with $N$ pixels and $B$ bands. Typically, the measured noisy HSI $\Y$ can be modeled by

\begin{equation}
        \Y=\X+\E,    \label{eq:noise}
\end{equation} where $\X \in \mathbb{R}^{N \times B}$ is the clean HSI to be recovered and $\E \in \mathbb{R}^{N \times B}$ is the noise term. Physical priors in HSIs are beneficial for denoising; however, they are often overlooked by most data-driven deep learning methods.  On the other hand,   it is generally challenging to build denoisers that strike a balance between effectively removing noise and preserving image details, especially when faced with severe noise levels that obscure fine structures and image nuances.

To address this challenge, we leverage the low-rank characteristics of HSIs as constraints within a deep neural network and approach HSI denoising as an iterative refinement process. Specifically, we employ an initial denoiser $g(\cdot)$ to obtain a coarse estimation of the clean HSI from the noisy observation:

\begin{equation}
        \X^{(0)}=g(\Y).    \label{eq:coarse}
\end{equation}
The initial denoiser $g(\cdot)$ serves as a fundamental step in extracting the primary structure from the noisy HSIs. Subsequently, we iteratively refine the previous estimation through a process that consists of two main steps:

\begin{equation}
        \begin{alignedat}{2}
           &\Z^{(k+1)}=(1-\lambda^{(k)}_{1})\Y+\lambda^{(k)}_{1}\X^{(k)},\\
           &\X^{(k+1)}=(1-\lambda^{(k)}_{2})f^{(k+1)}(\Z^{(k+1)})+\lambda_{2}^{(k)}\X^{(k)},\label{eq:refinement}\\
        \end{alignedat}
\end{equation} where $\lambda_{1} \in {\mathbb{R}^{B\times 1}}$ and $\lambda_{2} \in {\mathbb{R}^{B\times 1}}$ control the refinement level. The $\Z$ denotes the intermediate denoising result. The superscript $k$ indexes the iteration.

In each iteration, the intermediate image $\Z^{(k + 1)}$ is created by linearly combining the previously denoised HSI $\X^{(k)}$ with the original noisy HSI $\Y$ and contains more image details. The intermediate image is then refined by the  denoiser $f^{(k + 1)}(\cdot)$ to produce a new estimation $\X^{(k + 1)}$. This iterative process leverages the benefits of the lower noise level in $\Z^{(k + 1)}$ and the improved quality of $\X^{(k)}$ over time, making the refinement denoiser easiser to select and  more effective in capturing image details. As the iterative process progresses, the quality of $\X^{(k)}$ improves while the noise intensity decreases. Introducing $\Y$ in later iterations could compromise the quality of the output from $f^{(k + 1)}(\Z^{(k + 1)})$ due to potential reintroduction of too much noise. Combining $\X^{(k)}$ with the refined output from $f^{(k + 1)}(\Z^{(k + 1)})$ further mitigates this risk. To adjust the combined weights dynamically during the iteration process, $\lambda_1$ and $\lambda_2$ are generated adaptively by the $\Lambda(\cdot)$ module. In the following sections, we will describe  $g(\cdot)$, $f(\cdot)$, and $\Lambda(\cdot)$ in more detail.


\subsection{Iterative Low-rank  Network}

The  ILRNet is based on an iterative refinement process and is composed of three key components: the initial coarse estimation module, the multi-stage refinement module, and the adaptive weighting module $\Lambda(\cdot)$, as depicted in Fig.~\ref{fig:ilrnet}.

\subsubsection{Coarse Estimation Module} This module corresponds to  $g(\cdot)$ and combines the hybrid benefits of U-Net and low-rank representation to provide an initial estimation of clean HSI $\X^{(0)}$ from the noisy input $\Y$. $g({\cdot})$ has a 3D U-Net architecture with the residual connection constructed between the encoder and decoder.  In the encoder, 3D convolutions with kernel size of $3\times3\times3$ progressively extract spatial-spectral features, increasing their dimensions to \{16, 32, 64, 128\} while concurrently downsampling the feature maps by a factor of 2. Conversely, the decoder upsamples these features by the same factor, applying 3D deconvolutions to reconstruct the HSI, with feature dimensions incrementally restored to \{64, 32, 16, 1\}. ReLU activation follows each convolutional and deconvolutional layer to introduce non-linearity.

\begin{figure}
        \centering
        \includegraphics[width=1\linewidth]{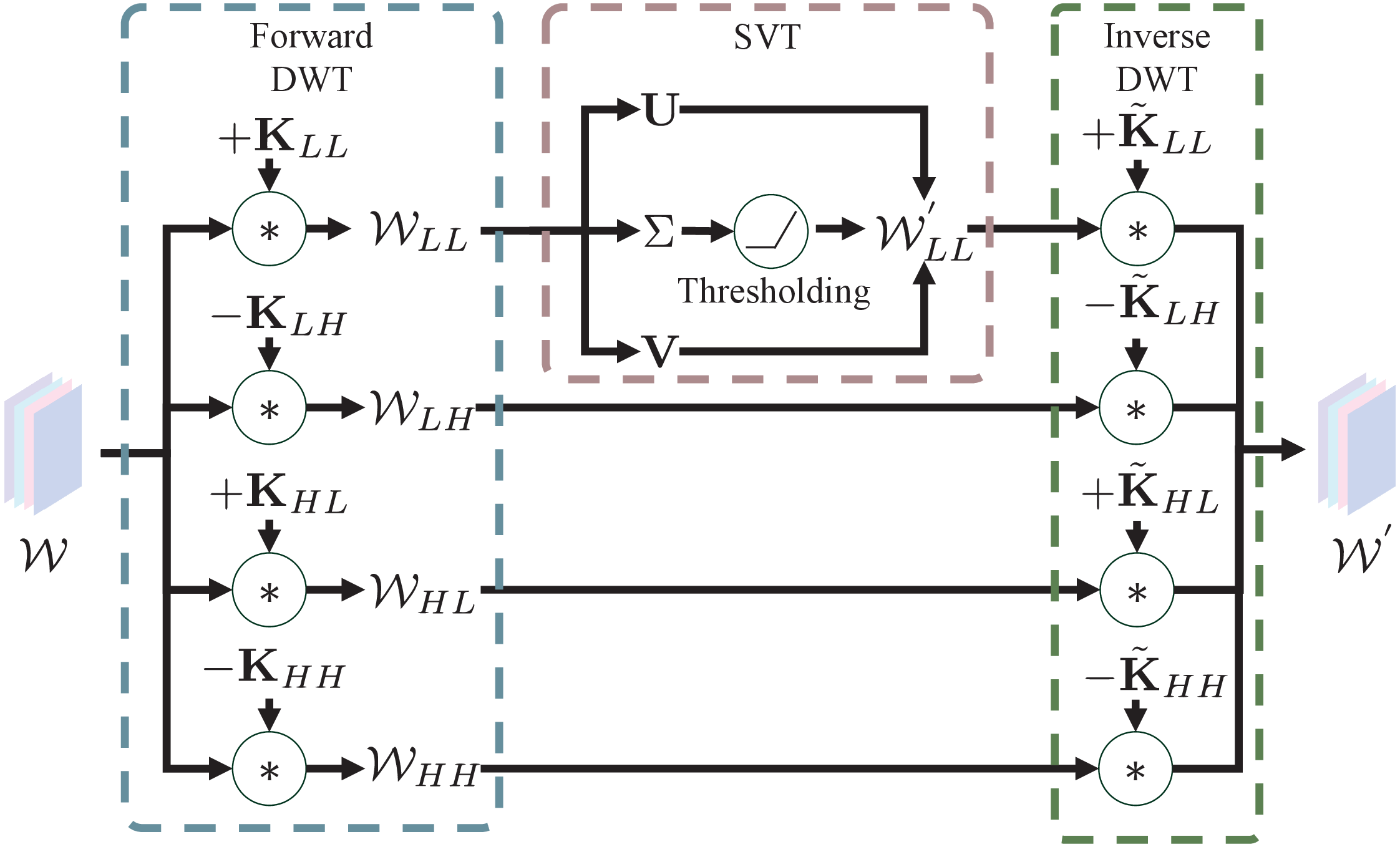}
        \caption{The implementation details of the RMM which  achieves low-rank constraints by applying SVT to low-frequency component of given feature map.}\label{fig:lr_module}
\end{figure}

As mentioned earlier, the low-rank representation in the transformed domain has more advantages than in the original domain.   Since the convolution operation in CNNs can be represented as the inner product of two vectors, showing a similarity to transforms in signal processing, which are also based on inner products~\cite{LiWaveletKernelNet}. Instead of being analytically fixed, the transforms are adaptively learned from the data with profound flexibility and map the input image to the feature space. On the other hand, just as frequency domain analysis can reveal the frequency characteristics of signals, the feature space generated by CNNs possesses unique attributes. Xu \emph{et al.}\cite{xutraining} and Lempitsky \emph{et al.}\cite{LempitskyDIP} consistently show that during the training of neural networks, models tend to first learn the low-frequency components of the image, which reflect the overall structure rather than local details. Therefore, to leverage the hybrid advantages of low-rank representation and CNNs, we integrate a RMM into the decoder of $g(\cdot)$. As illustrated in Fig.~\ref{fig:lr_module}, given feature maps $\Wcal$ with dimensions $C \times B \times W \times H$, RMM performs discrete wavelet transformation to obtain four distinct components: the low-frequency component $\Wcal_{LL}$; the vertical high-frequency component $\Wcal_{LH}$; the horizontal high-frequency component $\Wcal_{HL}$; and the diagonal high-frequency component $\Wcal_{HH}$. Using the 2D Haar DWT, these frequency maps are mathematically generated by
\begin{equation}
	\begin{split}
		&\Wcal_{LL}=\Wcal\otimes \left[\begin{array}{cc}
						1&1\\
					    1&1\\	
					 \end{array}\right]\downarrow	\\
	  &\Wcal_{LH}=\Wcal\otimes \left[\begin{array}{cc}
						-1&-1\\
					    1&1\\	
					 \end{array}\right]\downarrow	\\
      &\Wcal_{HL}=\Wcal\otimes \left[\begin{array}{cc}
						-1&1\\
					    -1&1\\	
					 \end{array}\right]\downarrow	\\
	&\Wcal_{HH}=\Wcal\otimes \left[\begin{array}{cc}
						1&-1\\
					    -1&1\\	
					 \end{array}\right]\downarrow	\\
	\end{split}
\end{equation}
where  $\downarrow$ denotes the downsampling operator.

The high-frequency components typically contain noise but also include intricate image details, which typically have no low-rank property. The low-frequency component is known for preserving the overall geometry structure of an image, which tends to exhibit low-rank characteristics. Applying SVT only to the low-frequency components helps avoid excessive smoothing of high-frequency components, especially fine details and edge information. Consequently, following the decomposition, we apply soft SVT to each channel of the low-frequency component $\Wcal_{LL}$, which has dimensions $B \times \frac{W}{2}\times \frac{H}{2}$, to achieve the spectral low-rankness characteristic of HSIs. Reshaping $\Wcal_{LL}$ into $\W_{LL}\in \mathbb{R}^{B\times (\frac{W}{2}\times \frac{H}{2}) }$, in the forward pass, SVD is performed as follows :
\begin{equation}
       \W_{LL}=\U\Sigma\V.    \label{eq:SVD}
\end{equation}
Here, $\U$ represents the left singular matrix, $\V$ represents the right singular matrix, and $\Sigma$ is the diagonal matrix containing singular values, typically arranged in descending order.  Traditionally, the SVT is performed  according to
\begin{equation}
        \W_{LL}^{'}=\sum_{i} \text{Relu}(\sigma_{i}- \lambda)\u_{i}\v_{i}^{T},    \label{eq:threshold}
\end{equation}
where $\u_i$ and $\v_i$ denote the left and right singular vectors corresponding to $\U$ and $\V$, respectively, and $\sigma_i$ is the $i$-th entry in $\Sigma$. The thresholding parameter $\lambda$ is challenging to determine but has a significant impact on performance. A larger $\lambda$ results in an over-smoothed image, whereas a smaller $\lambda$ leads to ineffective noise removal.

%

 Therefore,  we instead adaptively learn the threshold parameter from data and perform SVT  as follows:
\begin{equation}
       \W_{LL}^{'}=\sum_{i} \text{Relu}(\sigma_{i}- \text{Sigmoid}(d)\ast\sigma_{1})\u_{i}\v_{i}^{\T},    \label{eq:threshold}
\end{equation}
Here, $\sigma_{1}$ is the largest singular value, and $\text{Sigmoid}(d) \ast\sigma_{1}$ serves as  threshold in the range of $[0, \sigma_{1}]$, with $d$ learned from the data. In such a manner, the threshold parameter is data-adaptive, proving more flexibility for different scenarios. Finally, $\W_{LL}^{'}$ is reshaped into $\Wcal_{LL}^{'} \in \mathbb{R}^{B\times \frac{W}{2}\times \frac{H}{2}}$   for reconstructing feature maps $\Wcal^{'}$ .

Based on the derivation by Ionescu \emph{et al.}~\cite{Ionescu2015Matrix}, the formula for backpropagation of SVD in the RMM is given as follows:
\begin{equation}
\begin{split}
        \frac{\partial \mathcal{L}_\theta}{\partial \W_{LL}^{'}}=&\U \bigg\lbrace2\Sigma {\left(\K^{T}\cdot \left(\V^{T} \frac{\partial \mathcal{L}_\theta}{\partial \V}\right)\right)}_{sym} \\
        &+ {\left(\frac{\partial \mathcal{L}_\theta}{\partial \Sigma}\right)}_{diag} \bigg\rbrace \V^{\T}    \label{eq:SVD_backpropagation}
\end{split}
\end{equation}with
\begin{equation}
        \K_{i,j}=
        \begin{cases}
                \frac{1}{{\sigma}^2_i-{\sigma}^2_j},\quad &i \neq j\\
                0,\quad &i = j
        \end{cases},
\end{equation} where $\mathcal{L}_\theta$ denotes the loss function, $\sigma_i$ and $\sigma_j$ denote the $i$-th and $j$-th largest singular values, respectively. For any square matrix $\M$, $\M_{sym}$ denotes $(\M^\T + \M )/2$, and $\M_{diag}$ denotes setting all non-diagonal elements of $\M$ to 0. To mitigate gradient explosion risks associated with closely spaced singular values, we adopt the strategy of Wang \emph{et al.}~\cite{Wang2022Robust} that introduces a Taylor expansion truncated at the 9th order for $\K_{i,j}$, ensuring numerical stability without compromising the effectiveness of the low-rank constraint:
\begin{equation}
        \K_{i,j}\approx
        \begin{cases}
                 \frac{1}{\sigma_i+\sigma_j}\frac{1}{\sigma_i}\sum_{k=0}^{9}{\left( \frac{\sigma_j}{\sigma_i} \right)}^{k}, &\sigma_i > \sigma_j\\
                -\frac{1}{\sigma_i+\sigma_j}\frac{1}{\sigma_j}\sum_{k=0}^{9}{\left( \frac{\sigma_i}{\sigma_j} \right)}^{k}, &\sigma_i < \sigma_j\\
        \end{cases}.
\end{equation}

\subsubsection{Multi-stage Refinement Module} Since performing SVD on large matrices is extremely time-consuming, and executing SVD in every iteration would make the inference time unaffordable.  Additionally, the purpose of the refinement process is to supplement the intermediate results with detailed textures to further enhance the denoising effect. The intermediate results have already achieved a certain level of denoising, and we believe this task can be effectively accomplished with a straightforward network. Therefore, the RMM is excluded in this module. Analogous to the coarse estimation stage, this U-net also employs 3D convolutions with kernel size of $3\times3\times3$ and a downsampling factor of 2 in its encoder, extracting spatial-spectral features of dimensions of \{16,32\} to handle the cleaner input. The decoder employs 3D deconvolutions and upsampling, restoring the HSI while reducing dimensions back to \{16,1\}. As iterations progress and noise diminishes, the evolving demands necessitate distinct $f(\cdot)$ configurations for each refinement iteration, leading to non-shared parameters across these iterations.

\subsubsection{$ {\Lambda}({\cdot}) $ Module}


The $\Lambda(\cdot)$ module is specifically designed to dynamically generate weights $\lambda_1$ and $\lambda_2$ for each refinement iteration.  These weights play crucial roles: $\lambda_1$ controls the integration of $\X^{(k)}$ with $\Y$ to enrich the level of detail, while $\lambda_2$ combines $\X^{(k)}$ with the output of the refinement step $f^{(k+1)}(\Z^{(k+1)})$ to counterbalance potential degradation from noise introduced by $\Y$. As the refinement process continues, $\X^{(k)}$ improves in quality and reduces in noise, necessitating adaptive adjustments in $\lambda_1$ and $\lambda_2$ rather than relying on  the fixed  values. Moreover, given that noise intensity varies across different spectral bands due to differences in sensor sensitivities, $\lambda_1$ and $\lambda_2$ must be band-dependent to effectively accommodate these spectral variations. To this end, our ${\lambda}({\cdot})$ module is tasked with inferring bandwise $\lambda_1$ and $\lambda_2$ for every iteration:
\begin{equation}
        \begin{alignedat}{2}
           &\lambda^{(k)}_{1}=\Lambda(\text{concat}(\X^{(k)},\Y)),\\
           &\lambda^{(k)}_{2}=\Lambda(\text{concat}(\X^{(k)},\Z^{(k+1)}))\label{eq:generate_lambda}\\
        \end{alignedat}
\end{equation}

To promote band-specific variation in the inferred $\lambda_1$ and $\lambda_2$, we employ 2D convolutions tailored for each band. The process starts by extracting $56 \times 56$ pixel patches from the central region of the input, followed by a sequence of three 2D convolutional layers with downsampling, ultimately culminating in the application of a sigmoid function to generate $\lambda$. Recognizing the different roles of $\lambda_1$ and $\lambda_2$, two separate $\Lambda(\cdot)$ modules are employed to ensure tailored predictions for each weight.

\subsection{Loss Function}
The loss function of our ILRNet measures the Euclidean distance between the estimated HSI and the ground truth HSI:
\begin{equation}
        \mathcal{L}_{\theta}=\frac{1}{2N}\sum_{i}^{N} {\parallel \text{ILRNet}(\Y_i,\theta)-\X_i\parallel}^2_F
\end{equation}
where $\theta$ represents the network parameters, $N$ is the total number of training samples, and $\parallel \cdot \parallel^2_F$ denotes the Frobenius norm.

\section{Experiment}\label{experiment}

We compared  ILRNet against alternative methods across both synthetic and real-world noise removal scenarios to evaluate its effectiveness. The denoising capabilities of all methods were quantified using standard metrics, including Peak Signal-to-Noise Ratio (PSNR), Structural Similarity Index (SSIM), and Spectral Angle Mapper (SAM). In general, higher PSNR and SSIM values, along with lower SAM values, denote superior denoising performance and fidelity to the original data.
\subsection{Experiment Settings and Implementation Details}

\subsubsection{Benchmarked Models}
The comparison involves 12 methods in the literature, encompassing 7 model-driven methods and 5 data-driven methods. The model-driven methods include BM4D~\cite{Maggioni2013BM4D}, MTSNMF~\cite{Ye2015MTSNMF}, LLRT~\cite{Chang2017LLRT}, NGMeet~\cite{He2022NGMeet}, LRMR~\cite{Zhang2014LRMR}, E-3DTV~\cite{Peng2020E-3DTV}, and 3DlogTNN~\cite{Zheng20203DlogTNN}, while the data-driven methods include T3SC~\cite{bodrito2021T3SC}, MAC-Net~\cite{Xiong2022MAC-Net}, TRQ3D~\cite{Pang2022TRQ3DNet}, SST~\cite{li2022spatialspectral} and DPNet-S~\cite{Xiongdpnet}.

\subsubsection{Noise Patterns}
In this paper, we consider non-independent and identically distributed (non-i.i.d.) Gaussian noise and mixture noise, as these two noise patterns are the most common noise patterns observed in real-world scenarios. Additionally, following the setup by Bodrito~\emph{et al.}~\cite{bodrito2021T3SC}, we adopt the noise pattern of "noise with spectrally correlated variance". The configurations are as follows:
\begin{itemize}
        \item \textbf{Non-i.i.d. Gaussian Noise:} Each band is contaminated by Gaussian noise with a standard deviation of $\sigma$. The standard deviations are uniformly drawn in a fixed interval, i.e., [0, 15], [0, 55], and [0, 95].
        \item \textbf{Mixture Noise:} On top of being contaminated by the non-i.i.d. Gaussian noise with $\sigma\in$[0, 95], the mixture noise additionally includes the following noises: impulse noise with intensities between 0.1 and 0.7, strip noise with columns of 5\%-15\%, and deadline noise, each contaminating one-third of the bands.
        \item \textbf{Noise with Spectrally Correlated Variance:} Gaussian noise with standard deviation $\sigma$ varying continuously across bands, following a Gaussian curve. Specifically, for each band $i\in [0,B-1]$, the standard deviation of the Gaussian noise is defined as
        \begin{equation}
                \sigma_i=\beta exp[-\frac{1}{4\eta^2}{(\frac{i}{c}-\frac{1}{2})}^2],    \label{eq:scv}
        \end{equation} where $\beta = 23.08$ and $ \eta = 0.157$ as in~\cite{bodrito2021T3SC}.
\end{itemize}
\subsubsection{Implementation Details}
Our method was implemented in PyTorch and trained with a batchsize of 8 on a single NVIDIA GeForce RTX 3090 GPU for 50 epochs. The Adam optimizer minimizes the loss function with an initial learning rate of 0.0001 and cuts to half every twenty epochs. The iteration count for ILRNet was set to 9. For data-driven methods, we trained different models for different noise patterns. We selected 100 HSIs from the ICVL dataset and randomly crop patches of size $64\times64\times31$ to construct the training dataset. We tested the model trained on the ICVL dataset on different testing datasets to verify the robustness of the denoising method. Considering that T3SC, TRQ3D, and SST can only operate on testing datasets with the same number of bands as the training dataset, for testing datasets with a different number of bands than ICVL, we divide them into multiple sub-images with the same number of bands as ICVL for testing.

\begin{table*}[!t]
        \caption{Comparison of Different Methods on 50 Testing HSIs from ICVL Dataset. The Top Three Values Are Marked as \1{Red}, \2{Blue}, And \3{Green}.}\label{tab:icvl}
        \centering
        \resizebox{\linewidth}{!}{
                \tablesize{
        \begin{tabular}{c|c|c|c|c|c|c|c|c|c|c|c|c|c|c|c}
                \Xhline{1.2pt}
        \multirow{3}*{$\sigma$}&\multirow{3}*{Index}&\multirow{3}*{\makebox[0.06\textwidth][c]{Noisy}}
        &\multicolumn{7}{c|}{\textbf{Model-driven methods}}&\multicolumn{6}{c}{\textbf{Data-driven methods}}\\
        \cline{4-16}
        &&&\multirow{1}*{\makebox[0.06\textwidth][c]{BM4D}}&\multirow{1}*{\makebox[0.06\textwidth][c]{MTSNMF}}&\multirow{1}*{\makebox[0.06\textwidth][c]{LLRT}}&\multirow{1}*{\makebox[0.06\textwidth][c]{NGMeet}}&\multirow{1}*{\makebox[0.06\textwidth][c]{LRMR}}&\multirow{1}*{\makebox[0.06\textwidth][c]{E-3DTV}}&\multirow{1}*{\makebox[0.06\textwidth][c]{3DlogTNN}}&\multirow{1}*{\makebox[0.06\textwidth][c]{T3SC}}&\multirow{1}*{\makebox[0.06\textwidth][c]{MAC-Net}}&\multirow{1}*{\makebox[0.06\textwidth][c]{TRQ3D}}&\multirow{1}*{\makebox[0.06\textwidth][c]{SST}}&\multirow{1}*{\makebox[0.06\textwidth][c]{DPNet-S}}&\makebox[0.06\textwidth][c]{\textbf{ILRNet}}\\
        &&&\multirow{1}*{\makebox[0.06\textwidth][c]{\cite{Maggioni2013BM4D}}}&\multirow{1}*{\makebox[0.06\textwidth][c]{\cite{Ye2015MTSNMF}}}&\multirow{1}*{\makebox[0.06\textwidth][c]{\cite{Chang2017LLRT}}}&\multirow{1}*{\makebox[0.06\textwidth][c]{\cite{He2022NGMeet}}}&\multirow{1}*{\makebox[0.06\textwidth][c]{\cite{Zhang2014LRMR}}}&\multirow{1}*{\makebox[0.06\textwidth][c]{\cite{Peng2020E-3DTV}}}&\multirow{1}*{\makebox[0.06\textwidth][c]{\cite{Zheng20203DlogTNN}}}&\multirow{1}*{\makebox[0.06\textwidth][c]{\cite{bodrito2021T3SC}}}&\multirow{1}*{\makebox[0.06\textwidth][c]{\cite{Xiong2022MAC-Net}}}&\multirow{1}*{\makebox[0.06\textwidth][c]{\cite{Pang2022TRQ3DNet}}}&\multirow{1}*{\makebox[0.06\textwidth][c]{\cite{li2022spatialspectral}}}&\multirow{1}*{\makebox[0.06\textwidth][c]{\cite{Xiongdpnet}}}&\makebox[0.06\textwidth][c]{\textbf{(Ours)}}\\
        \Xhline{1.2pt}   		
        \multirow{3}*{\textbf{[0,15]}}
        & PSNR$\uparrow$  & 33.18 & 44.39 & 45.39 & 45.74 & 39.63 & 41.50 & 46.05 & 43.89 & 49.68 & 48.21 & 46.43 & \2{50.87} & \3{50.24} & \1{51.22} \\
        & SSIM$\uparrow$  & .6168 & .9683 & .9592 & .9657 & .8612 & .9356 & .9811 & .9902 & .9912 & .9915 & .9878 & \2{.9938} & \3{.9934} & \1{.9945} \\
        & SAM$\downarrow$ & .3368 & .0692 & .0845 & .0832 & .2144 & .1289 & .0560 & \1{.0150} & .0486 & .0387 & .0437 & \3{.0298} & .0299 & \2{.0255} \\
        \hline 		
        \multirow{3}*{\textbf{[0,55]}}
        & PSNR$\uparrow$  & 21.72 & 37.63 & 38.02 & 36.80 & 31.53 & 31.50 & 40.20 & 33.37 & \3{45.15} & 43.74 & 44.64 & \2{46.39} & 44.82 & \1{46.88} \\
        & SSIM$\uparrow$  & .2339 & .9008 & .8586 & .8285 & .6785 & .6233 & .9505 & .6892 & .9810 & .9768 & \3{.9840} & \2{.9872} & .9832 & \1{.9886} \\
        & SAM$\downarrow$ & .7012 & .1397 & .2340 & .2316 & .4787 & .3583 & .0993 & .2766 & .0652 & .0582 & .0487 & \2{.0457} & \3{.0480} & \1{.0350} \\
        \hline\
        \multirow{3}*{\textbf{[0,95]}}\
        & PSNR$\uparrow$  & 17.43 & 34.71 & 34.81 & 31.89 & 27.62 & 27.00 & 37.80 & 24.53 & 43.10 & 41.24 & \3{43.54} & \2{44.83} & 42.95 & \1{45.46} \\
        & SSIM$\uparrow$  & .1540 & .8402 & .7997 & .6885 & .5363 & .4208 & .9279 & .4251 & .9734 & .9577 & \3{.9806} & \2{.9838} & .9772 & \1{.9853} \\
        & SAM$\downarrow$ & .8893 & .1906 & .3266 & .3444 & .6420 & .5142 & .1317 & .6087 & .0747 & .0841 & \3{.0523} & \2{.0513} & .0574 & \1{.0404} \\
        \Xhline{1.2pt}  		  \
        \multirow{3}*{\textbf{Mixture}}  \
        & PSNR$\uparrow$  & 13.21 & 23.36 & 27.55 & 18.23 & 23.61 & 23.10 & 34.90 & 17.52 & 34.09 & 28.44 & \1{39.73} & \2{39.22} & 38.61 & \3{39.06} \\
        & SSIM$\uparrow$  & .0841 & .4275 & .6743 & .1731 & .4448 & .3463 & .9041 & .2389 & .9052 & .7393 & \2{.9491} & \1{.9626} & \3{.9312} & .9227 \\
        & SAM$\downarrow$ & .9124 & .5476 & .5326 & .6873 & .6252 & .5144 & .1468 & .6905 & .2340 & .4154 & \2{.0869} & \1{.0743} & \3{.1178} & .1423 \\
        \Xhline{1.2pt}  		\
        \multirow{3}*{\textbf{Corr}}\
        & PSNR$\uparrow$  & 28.22 & 41.15 & 42.44 & 41.92 & 35.82 & 39.32 & 43.58 & 41.49 & \3{47.33} & 45.60 & 46.26 & \2{47.59} & 45.51 & \1{47.80} \\
        & SSIM$\uparrow$  & .4640 & .8963 & .9221 & .9080 & .7891 & .9081 & .9733 & .9709 & .9858 & .9853 & \3{.9870} & \1{.9904} & .9813 & \2{.9890} \\
        & SAM$\downarrow$ & .4601 & .1582 & .1121 & .1547 & .3113 & .1212 & .0601 & .0574 & .0524 & .0456 & \3{.0403} & \1{.0258} & .0481 & \2{.0323} \\
        \Xhline{1.2pt} 		
        \end{tabular}}}
\end{table*}

\begin{table*}[!t]
        \caption{Comparisons of the number of parameters and running time of different methods.}\label{tab:params_time}
        \centering
        \resizebox{\linewidth}{!}{
                \tablesize{
        \begin{tabular}{c|c|c|c|c|c|c|c|c|c|c|c|c|c}
                \Xhline{1.2pt}
        \multirow{3}*{Metric}
        &\multicolumn{7}{c|}{\textbf{Model-driven methods}}&\multicolumn{6}{c}{\textbf{Data-driven methods}}\\
        \cline{2-14}
        &\multirow{1}*{\makebox[0.06\textwidth][c]{BM4D}}&\multirow{1}*{\makebox[0.06\textwidth][c]{MTSNMF}}&\multirow{1}*{\makebox[0.06\textwidth][c]{LLRT}}&\multirow{1}*{\makebox[0.06\textwidth][c]{NGMeet}}&\multirow{1}*{\makebox[0.06\textwidth][c]{LRMR}}&\multirow{1}*{\makebox[0.06\textwidth][c]{E-3DTV}}&\multirow{1}*{\makebox[0.06\textwidth][c]{3DlogTNN}}&\multirow{1}*{\makebox[0.06\textwidth][c]{T3SC}}&\multirow{1}*{\makebox[0.06\textwidth][c]{MAC-Net}}&\multirow{1}*{\makebox[0.06\textwidth][c]{TRQ3D}}&\multirow{1}*{\makebox[0.06\textwidth][c]{SST}}&\multirow{1}*{\makebox[0.06\textwidth][c]{DPNet-S}}&\makebox[0.06\textwidth][c]{\textbf{ILRNet}}\\
        &\multirow{1}*{\makebox[0.06\textwidth][c]{\cite{Maggioni2013BM4D}}}&\multirow{1}*{\makebox[0.06\textwidth][c]{\cite{Ye2015MTSNMF}}}&\multirow{1}*{\makebox[0.06\textwidth][c]{\cite{Chang2017LLRT}}}&\multirow{1}*{\makebox[0.06\textwidth][c]{\cite{He2022NGMeet}}}&\multirow{1}*{\makebox[0.06\textwidth][c]{\cite{Zhang2014LRMR}}}&\multirow{1}*{\makebox[0.06\textwidth][c]{\cite{Peng2020E-3DTV}}}&\multirow{1}*{\makebox[0.06\textwidth][c]{\cite{Zheng20203DlogTNN}}}&\multirow{1}*{\makebox[0.06\textwidth][c]{\cite{bodrito2021T3SC}}}&\multirow{1}*{\makebox[0.06\textwidth][c]{\cite{Xiong2022MAC-Net}}}&\multirow{1}*{\makebox[0.06\textwidth][c]{\cite{Pang2022TRQ3DNet}}}&\multirow{1}*{\makebox[0.06\textwidth][c]{\cite{li2022spatialspectral}}}&\multirow{1}*{\makebox[0.06\textwidth][c]{\cite{Xiongdpnet}}}&\makebox[0.06\textwidth][c]{\textbf{(Ours)}}\\
        \Xhline{1.2pt}
        PSNR & 34.71 & 34.81 & 31.89 & 27.62 & 27.00 & 37.80 & 24.53 & 43.10 & 41.24 & 43.54 & 44.83 & 42.95 & 45.46 \\
        \hline    		
        \#Params(M) & - & - & - & - & - & - & - & 0.83 & 0.43 & 0.67 & 4.14 & 0.20 & 3.67 \\
        \hline 		
        Times(s) & 198.05 & 40.00 & 1673.80 & 512.13 & 274.92 & 55.55 & 188.51 & 0.57 & 3.41 & 1.44 & 1.83 & 0.98 & 2.12 \\
	\Xhline{1.2pt}
        \end{tabular}}}
\end{table*}

\begin{figure*}[!t]
        \centering
        \subfloat[]{\label{fig:Master5000K_clean_}\includegraphics[width=0.1240\linewidth]{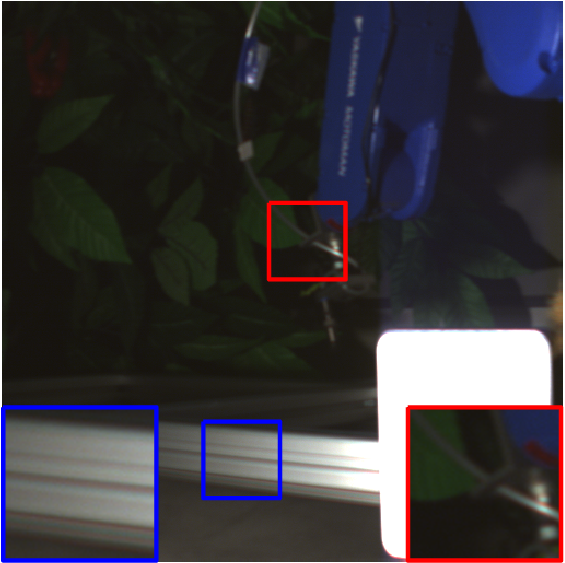}}
        \hspace{-1.1mm}
        \subfloat[]{\label{fig:Master5000K_noise_}\includegraphics[width=0.1240\linewidth]{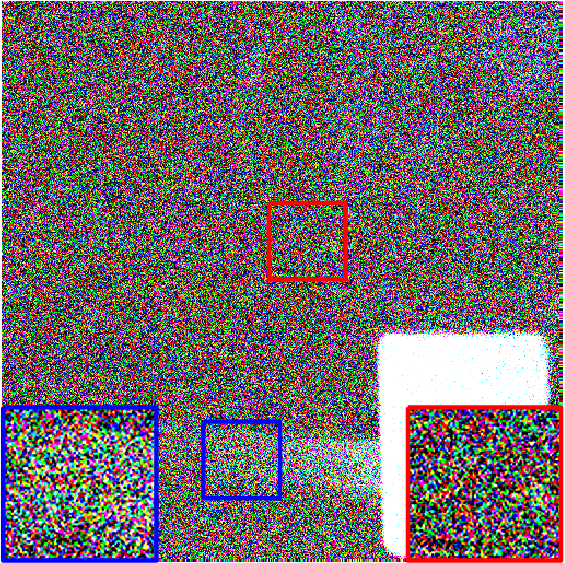}}
        \hspace{-1.1mm}
        \subfloat[]{\label{fig:Master5000K_BM4D}\includegraphics[width=0.1240\linewidth]{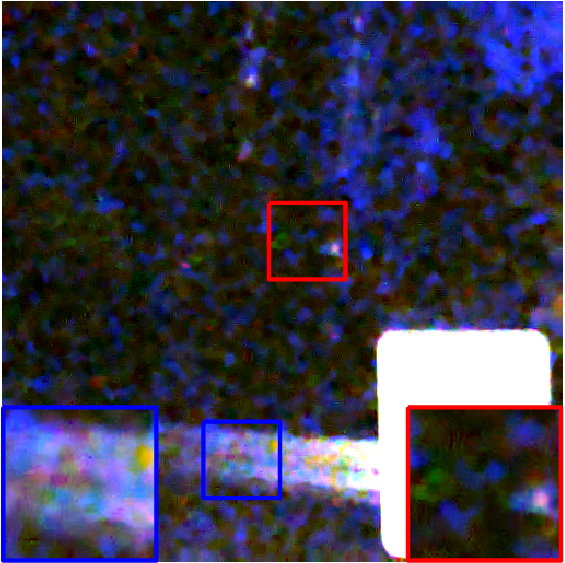}}
        \hspace{-1.1mm}
        \subfloat[]{\label{fig:Master5000K_MTSNMF}\includegraphics[width=0.1240\linewidth]{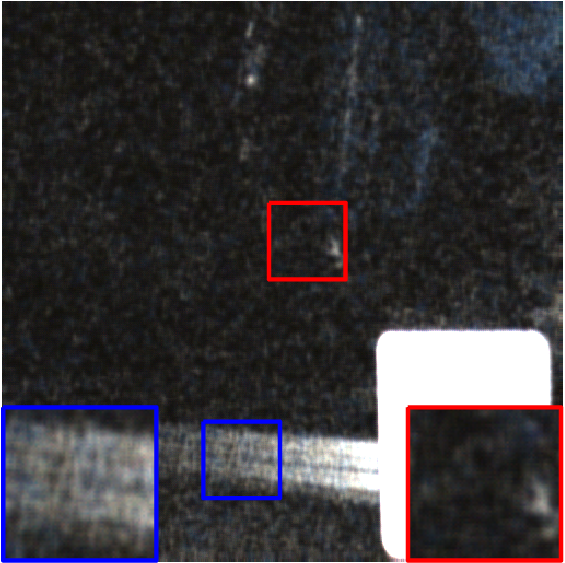}}
        \hspace{-1.1mm}
        \subfloat[]{\label{fig:Master5000K_LLRT}\includegraphics[width=0.1240\linewidth]{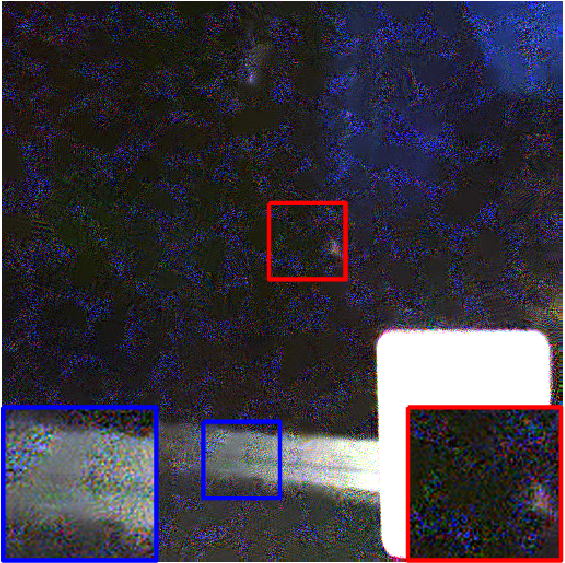}}
        \hspace{-1.1mm}
        \subfloat[]{\label{fig:Master5000K_NGMeet}\includegraphics[width=0.1240\linewidth]{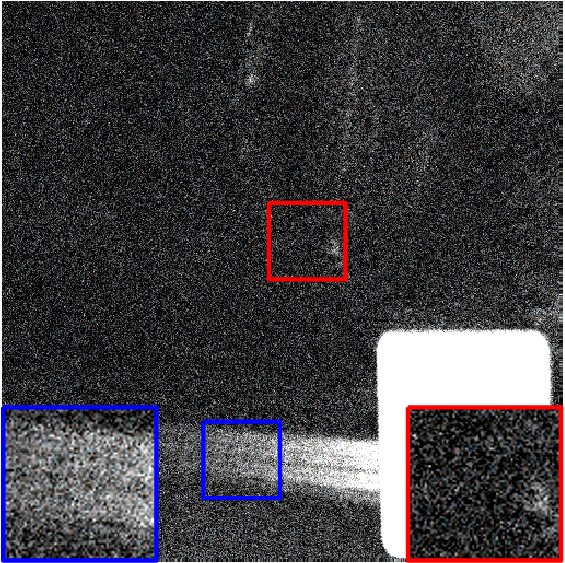}}
        \hspace{-1.1mm}
        \subfloat[]{\label{fig:Master5000K_LRMR}\includegraphics[width=0.1240\linewidth]{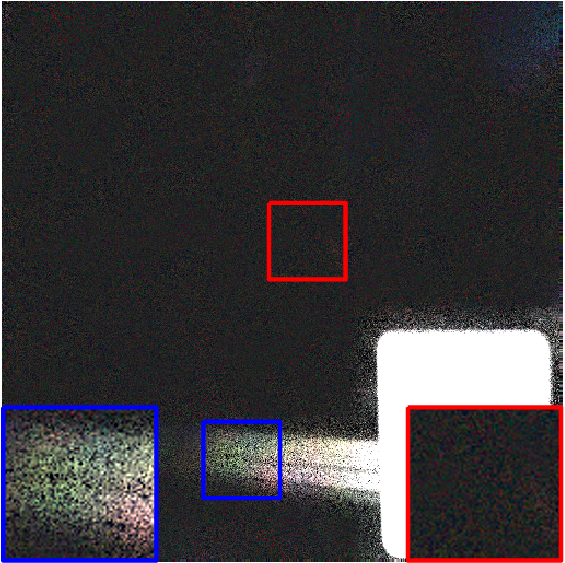}}
        \hspace{-1.1mm}
        \subfloat[]{\label{fig:Master5000K_E-3DTV}\includegraphics[width=0.1240\linewidth]{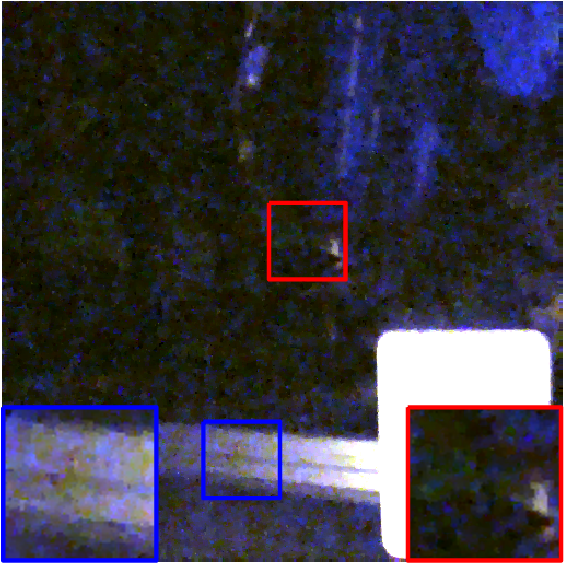}}
        \hspace{-1.1mm}
        \subfloat[]{\label{fig:Master5000K_3DlogTNN}\includegraphics[width=0.1240\linewidth]{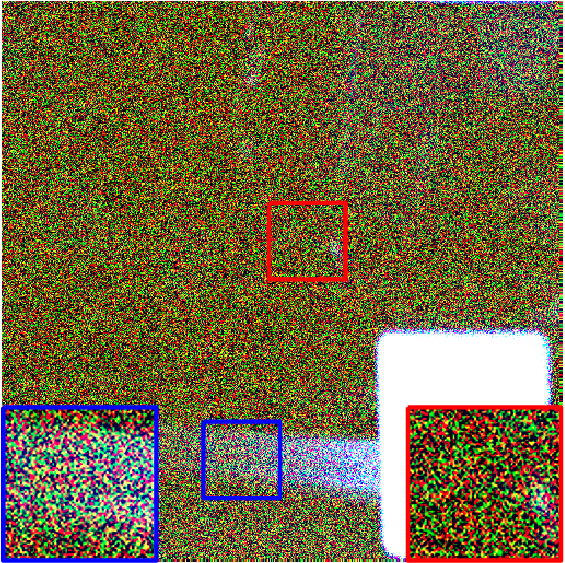}}
        \hspace{-1.1mm}
        \subfloat[]{\label{fig:Master5000K_T3SC}\includegraphics[width=0.1240\linewidth]{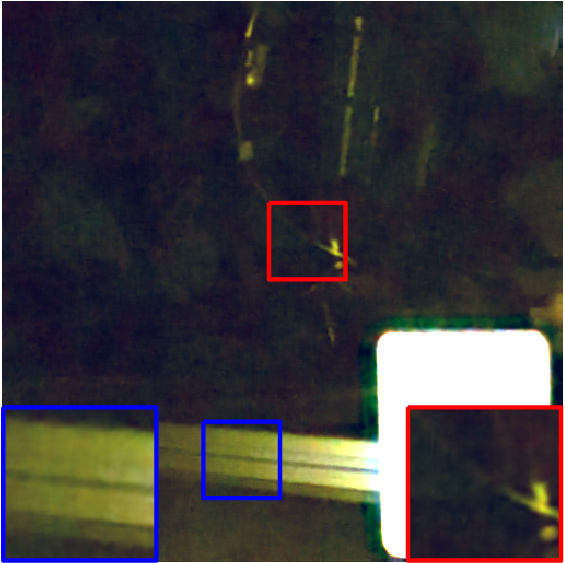}}
        \hspace{-1.1mm}
        \subfloat[]{\label{fig:Master5000K_MAC-Net}\includegraphics[width=0.1240\linewidth]{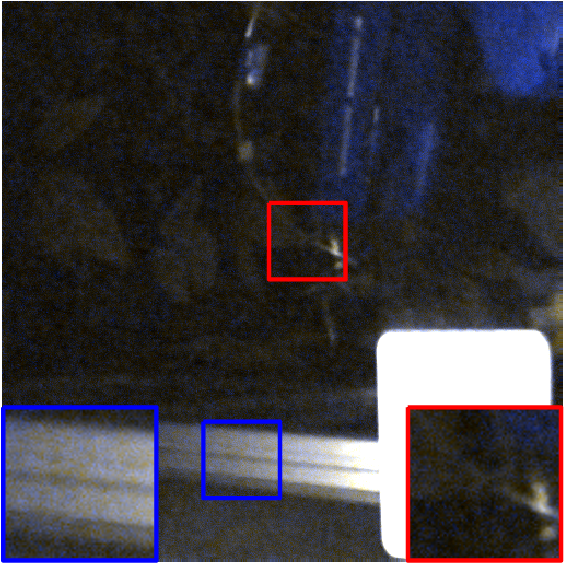}}
        \hspace{-1.1mm}
        \subfloat[]{\label{fig:Master5000K_TRQ3D}\includegraphics[width=0.1240\linewidth]{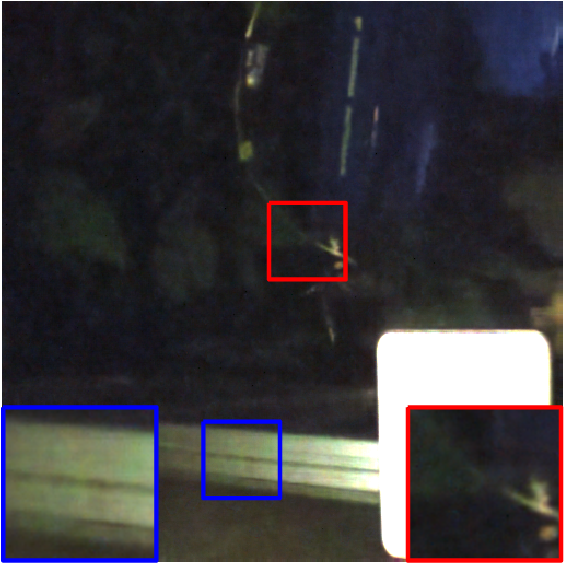}}
        \hspace{-1.1mm}
        \subfloat[]{\label{fig:Master5000K_SST}\includegraphics[width=0.1240\linewidth]{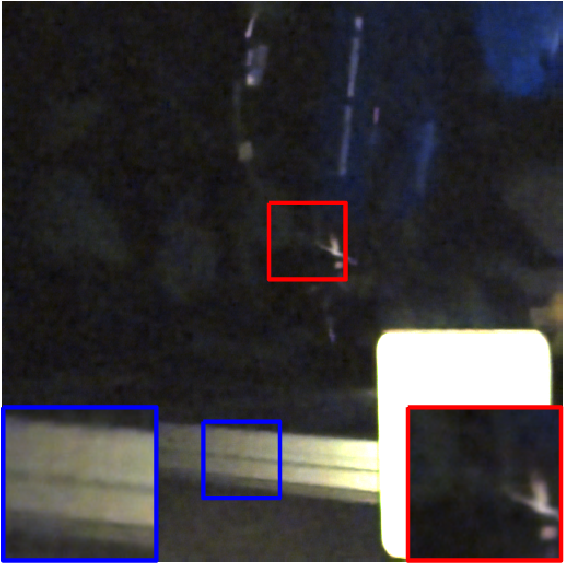}}
        \hspace{-1.1mm}
        \subfloat[]{\label{fig:Master5000K_DPNet}\includegraphics[width=0.1240\linewidth]{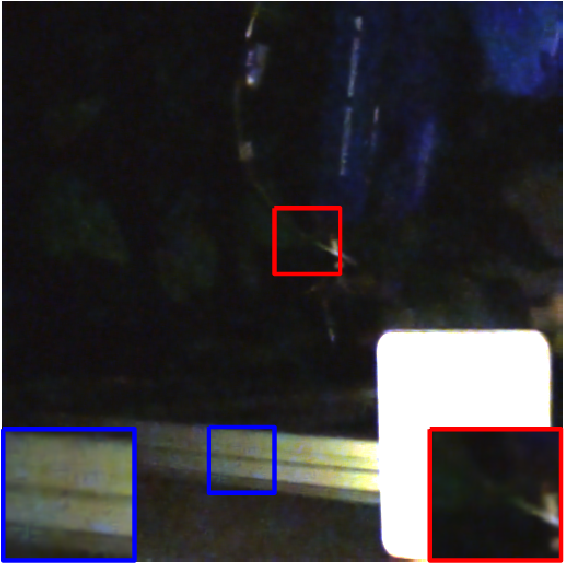}}
        \hspace{-1.1mm}
        \subfloat[]{\label{fig:Master5000K_ILRNet}\includegraphics[width=0.1240\linewidth]{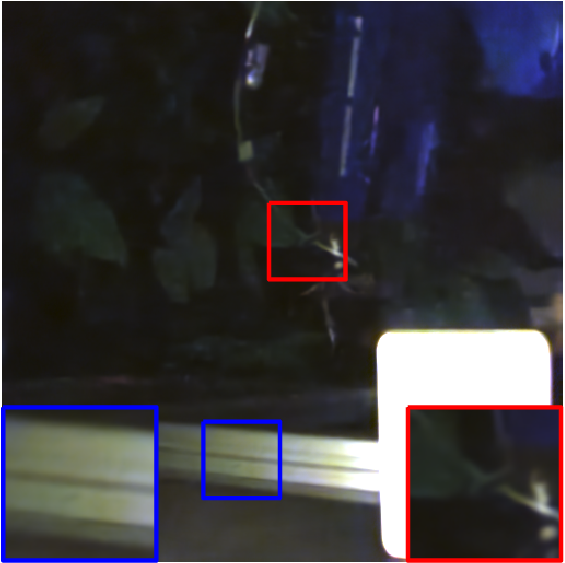}}
        \caption{Denoising results on the Master5000K HSI from the ICVL dataset with the non-i.i.d. Gaussian noise with $\sigma \in [0,95]$. The false-color images are generated by combining bands 23, 15, and 9. (a) Clean. (b) Noisy. (c) BM4D~\cite{Maggioni2013BM4D}. (d) MTSNMF~\cite{Ye2015MTSNMF}. (e) LLRT~\cite{Chang2017LLRT}. (f) NGMeet~\cite{He2022NGMeet}. (g) LRMR~\cite{Zhang2014LRMR}. (h) E-3DTV~\cite{Peng2020E-3DTV}. (i) 3DlogTNN~\cite{Zheng20203DlogTNN}. (j) T3SC~\cite{bodrito2021T3SC}. (k) MAC-Net~\cite{Xiong2022MAC-Net}. (l) TRQ3D~\cite{Pang2022TRQ3DNet}. (m) SST~\cite{li2022spatialspectral}. (n) DPNet-S~\cite{Xiongdpnet}. (o) \textbf{ILRNet}.} \label{fig:ICVL_visual}
\end{figure*}

\begin{figure*}[!t]
        \centering
        \subfloat[]{\label{fig:Master5000K_clean_reflectance}\includegraphics[width=0.1240\linewidth]{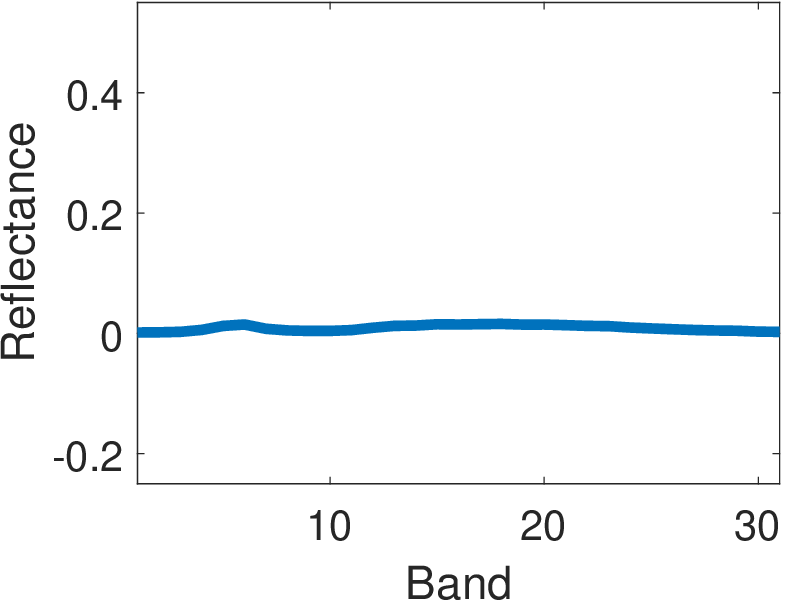}}
        \hspace{-1.1mm}
        \subfloat[]{\label{fig:Master5000K_noise_reflectance}\includegraphics[width=0.1240\linewidth]{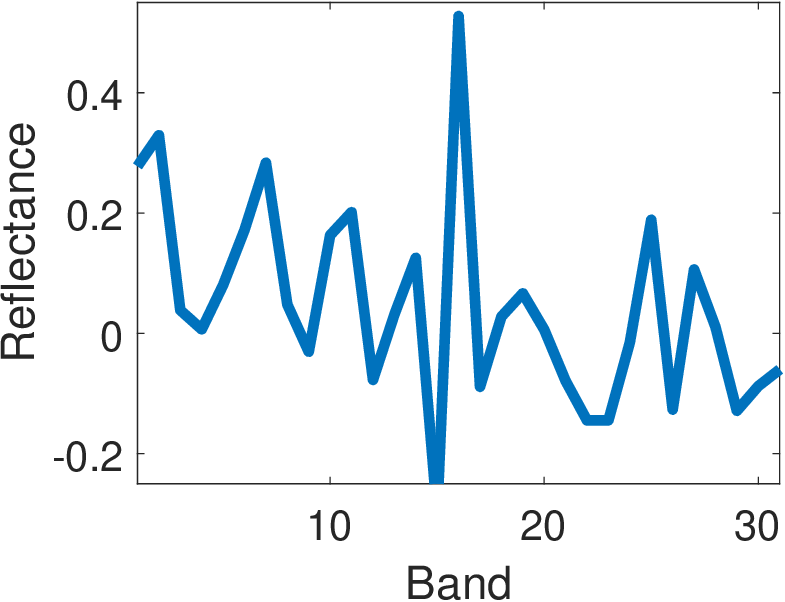}}
        \hspace{-1.1mm}
        \subfloat[]{\label{fig:Master5000K_BM4D_reflectance}\includegraphics[width=0.1240\linewidth]{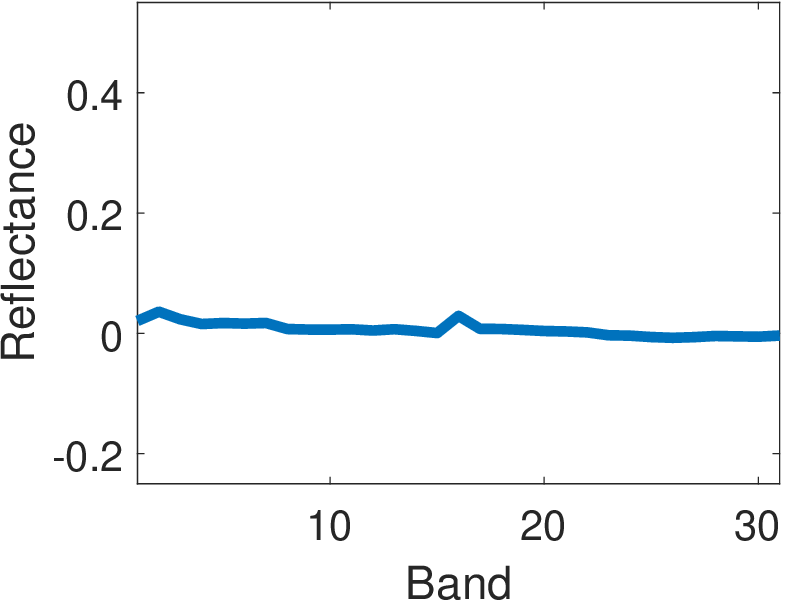}}
        \hspace{-1.1mm}
        \subfloat[]{\label{fig:Master5000K_MTSNMF_reflectance}\includegraphics[width=0.1240\linewidth]{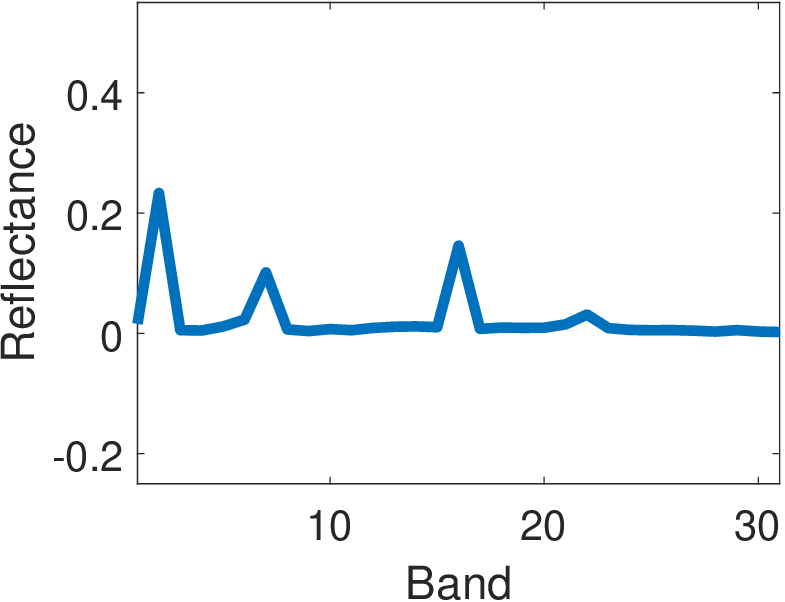}}
        \hspace{-1.1mm}
        \subfloat[]{\label{fig:Master5000K_LLRT_reflectance}\includegraphics[width=0.1240\linewidth]{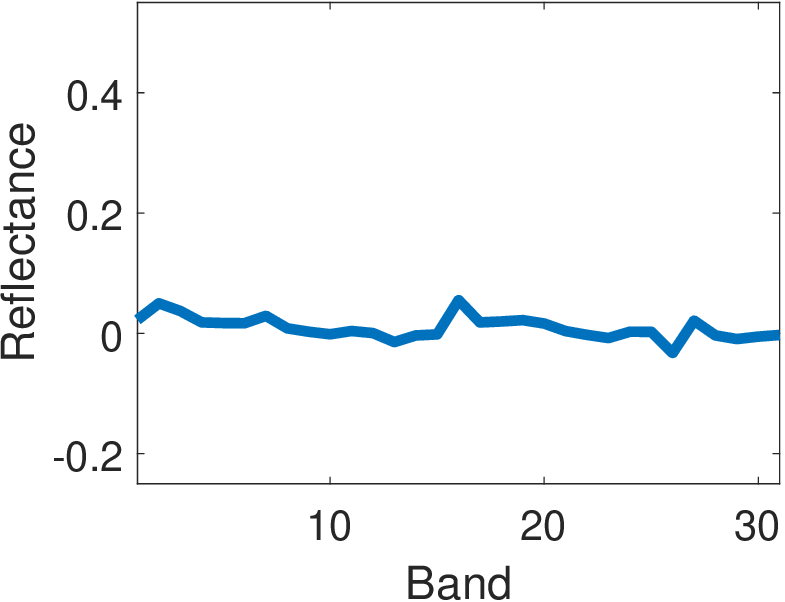}}
        \hspace{-1.1mm}
        \subfloat[]{\label{fig:Master5000K_NGMeet_reflectance}\includegraphics[width=0.1240\linewidth]{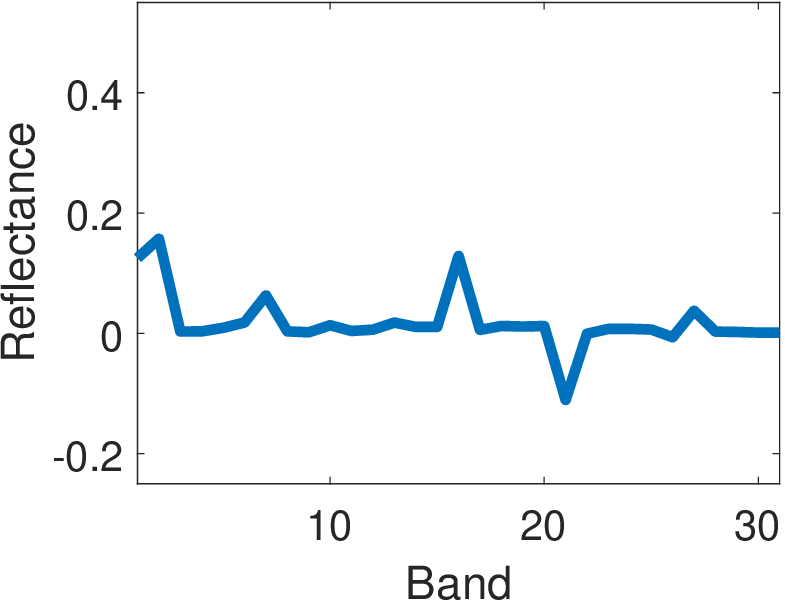}}
        \hspace{-1.1mm}
        \subfloat[]{\label{fig:Master5000K_LRMR_reflectance}\includegraphics[width=0.1240\linewidth]{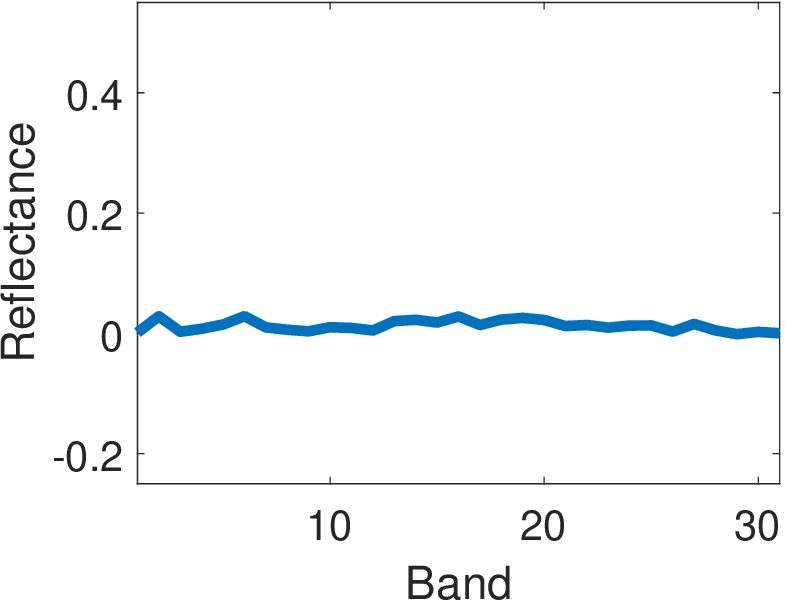}}
        \hspace{-1.1mm}
        \subfloat[]{\label{fig:Master5000K_E-3DTV_reflectance}\includegraphics[width=0.1240\linewidth]{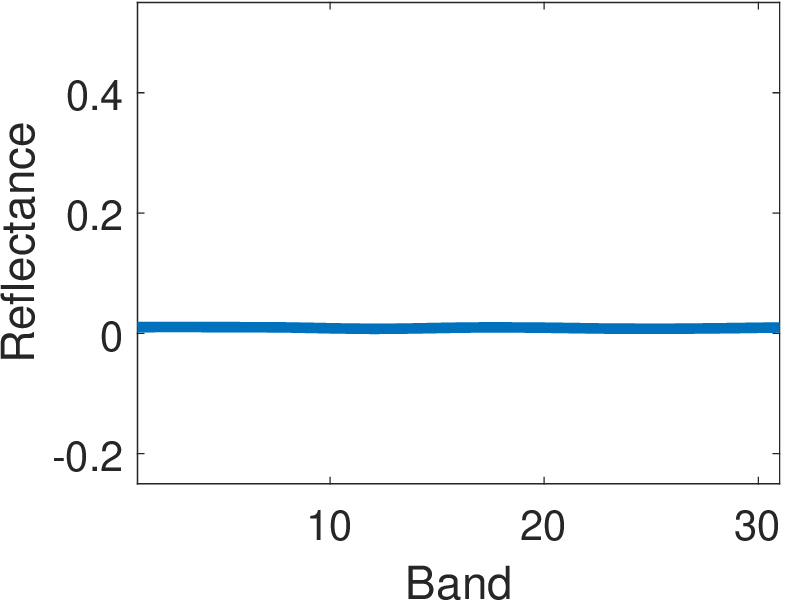}}
        \hspace{-1.1mm}
        \subfloat[]{\label{fig:Master5000K_3DlogTNN_reflectance}\includegraphics[width=0.1240\linewidth]{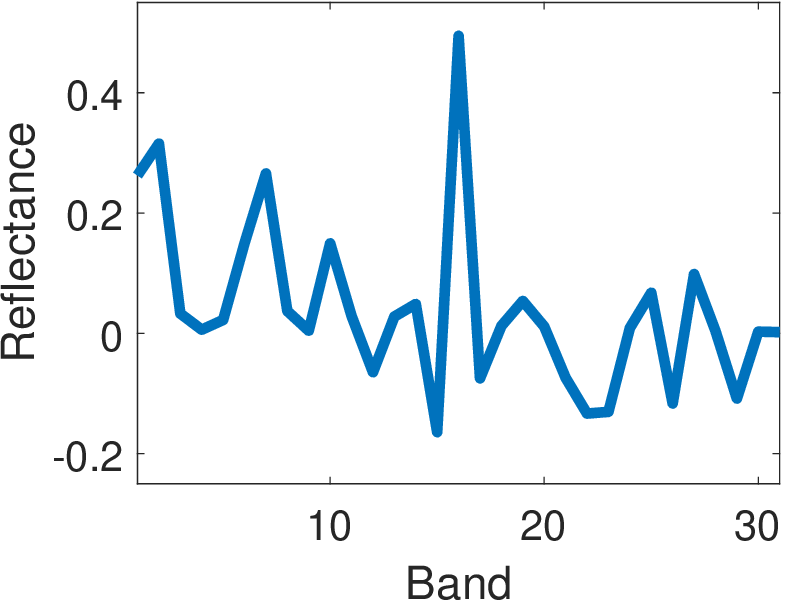}}
        \hspace{-1.1mm}
        \subfloat[]{\label{fig:Master5000K_T3SC_reflectance}\includegraphics[width=0.1240\linewidth]{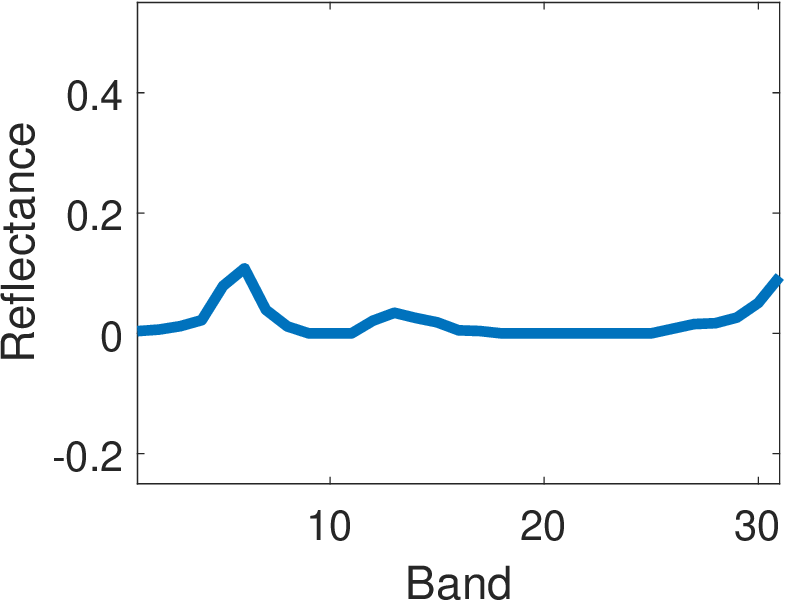}}
        \hspace{-1.1mm}
        \subfloat[]{\label{fig:Master5000K_MAC-Net_reflectance}\includegraphics[width=0.1240\linewidth]{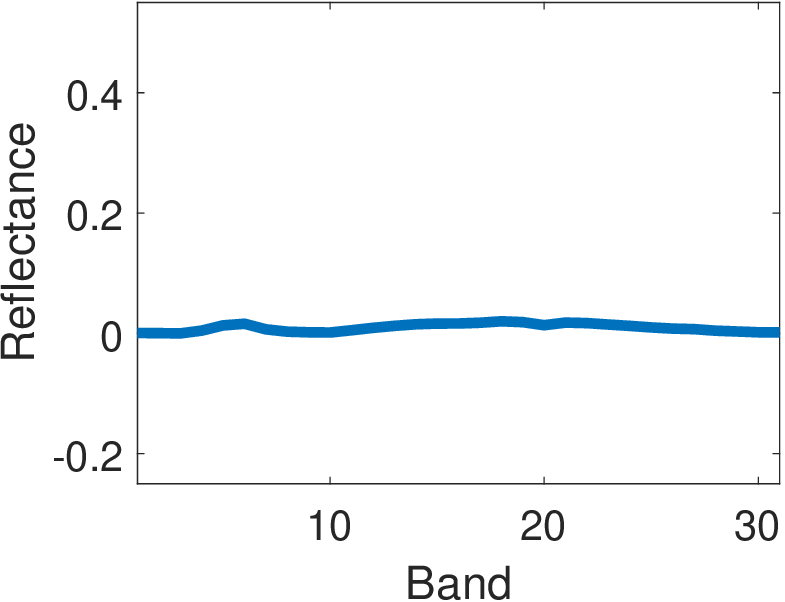}}
        \hspace{-1.1mm}
        \subfloat[]{\label{fig:Master5000K_TRQ3D_reflectance}\includegraphics[width=0.1240\linewidth]{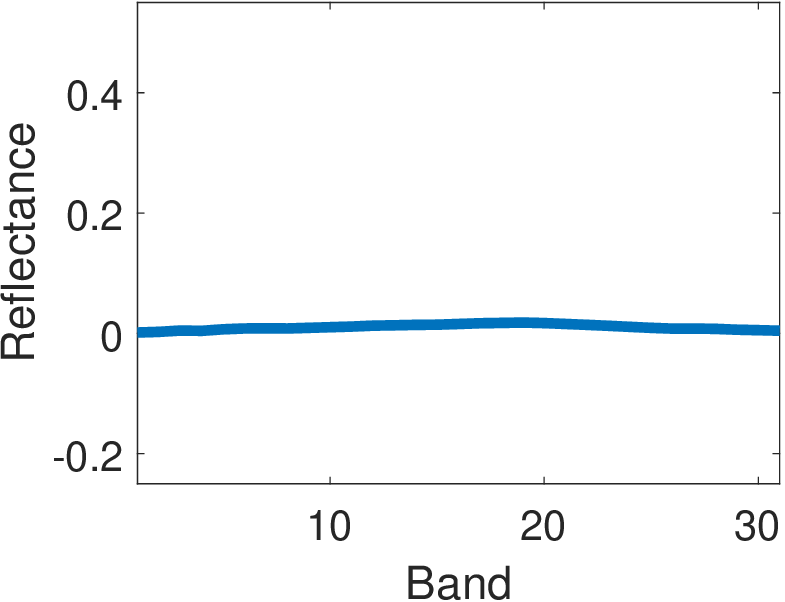}}
        \hspace{-1.1mm}
        \subfloat[]{\label{fig:Master5000K_SST_reflectance}\includegraphics[width=0.1240\linewidth]{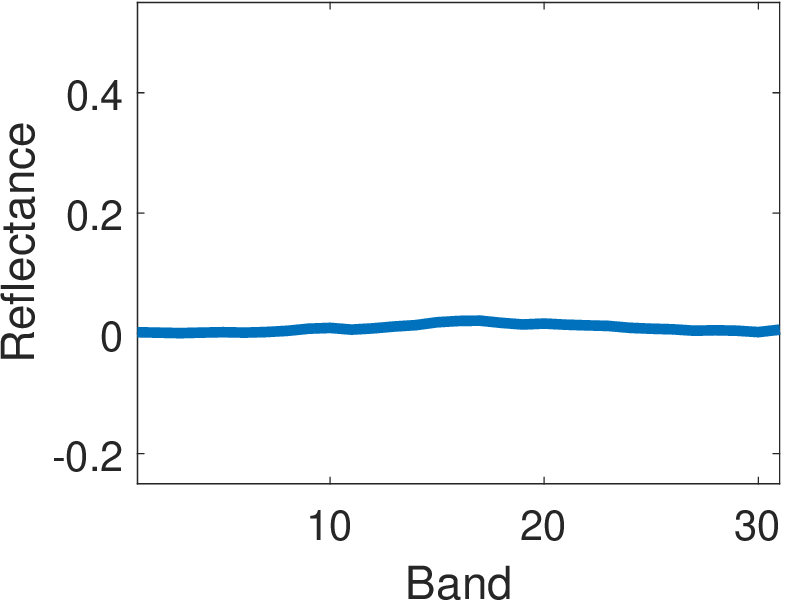}}
        \hspace{-1.1mm}
        \subfloat[]{\label{fig:Master5000K_DPNet_reflectance}\includegraphics[width=0.1240\linewidth]{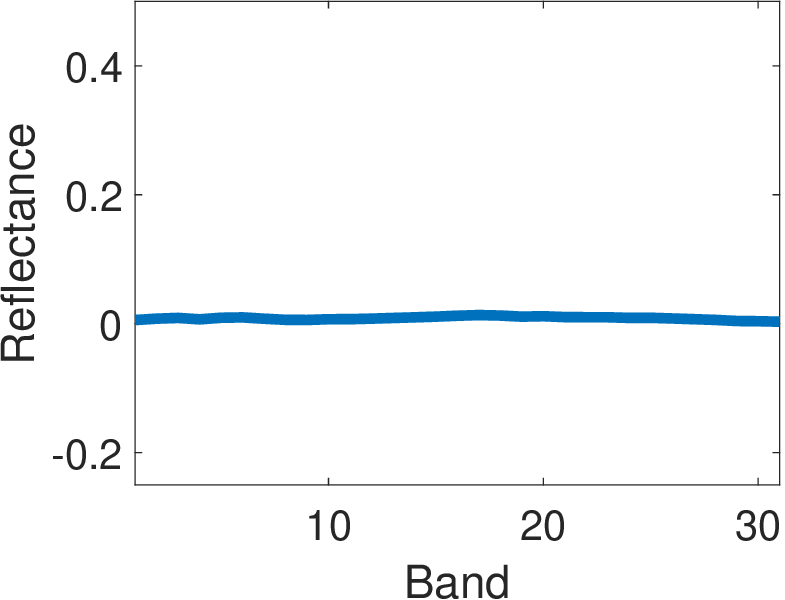}}
        \hspace{-1.1mm}
        \subfloat[]{\label{fig:Master5000K_ILRNet_reflectance}\includegraphics[width=0.1240\linewidth]{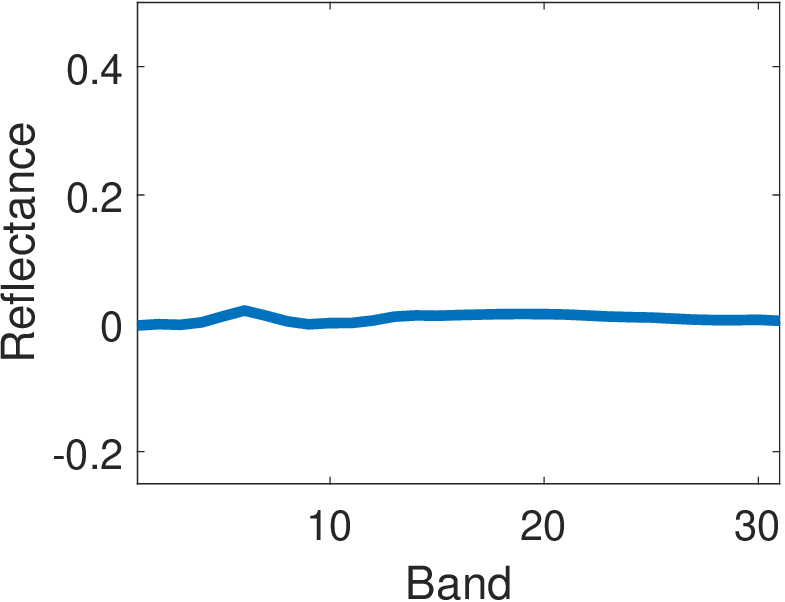}}
        \caption{Reflectance of pixel (340, 390) in the Master5000K HSI from the ICVL dataset with the non-i.i.d. Gaussian noise with $\sigma \in [0,95]$. (a) Clean. (b) Noisy. (c) BM4D~\cite{Maggioni2013BM4D}. (d) MTSNMF~\cite{Ye2015MTSNMF}. (e) LLRT~\cite{Chang2017LLRT}. (f) NGMeet~\cite{He2022NGMeet}. (g) LRMR~\cite{Zhang2014LRMR}. (h) E-3DTV~\cite{Peng2020E-3DTV}. (i) 3DlogTNN~\cite{Zheng20203DlogTNN}. (j) T3SC~\cite{bodrito2021T3SC}. (k) MAC-Net~\cite{Xiong2022MAC-Net}. (l) TRQ3D~\cite{Pang2022TRQ3DNet}. (m) SST~\cite{li2022spatialspectral}. (n) DPNet-S~\cite{Xiongdpnet}. (o) \textbf{ILRNet}.} \label{fig:ICVL_reflectance}
\end{figure*}

\subsection{Synthetic Noise Removal}

We conducted quantitative and qualitative experiments on denoising synthetic noise in both close-range HSIs, i.e., the ICVL dataset, and remote-range HSIs, i.e., the Pavia City Center HSI.
\subsubsection{ICVL Dataset}

The ICVL dataset was acquired using the Specim PS Kappa DX4 hyperspectral camera and a rotating platform for spatial scanning. Each image has a resolution of $1392 \times 1300$ pixels and contains 31 spectral channels from 400nm to 700nm at intervals of 10nm. The ICVL testing dataset consists of 50 HSIs from the ICVL dataset. To reduce computational burden, all these HSIs were cropped to a size of $512 \times 512 \times 31$.

Table~\ref{tab:icvl} quantitatively compares the denoising performance of different methods on the ICVL dataset. The top three results are highlighted with bold \1{red}, bold \2{blue}, and bold \3{green}, respectively. Benefiting from the powerful representation capability of DNNs, data-driven methods generally outperform model-driven methods. Since NGMeet is designed for i.i.d. Gaussian noise, it struggles to estimate noise intensity to compute the spectral subspace and adjust the parameters under non-i.i.d. Gaussian noise cases, leading to poor performance. E-3DTV demonstrates significantly better performance compared to other model-driven methods by naturally encoding the correlations and differences among spectral bands. It even approaches data-driven methods closely in experiments aimed at removing mixture noise. T3SC, MAC-Net and DPNet-S acquire strong noise removal capabilities through a combination of physical modeling and data-driven learning. Meanwhile, TRQ3D and SST achieve further noise removal performance by capturing non-local spatial self-similarity through attention mechanisms. Unlike the modeling approach to physical priors in methods such as MAC-Net and DPNet-S or the computationally intensive attention mechanisms in methods such as SST, ILRNet proposes an iterative refinement process. This mechanism models HSI denoising as an iterative process of detail supplementation and refinement. Furthermore, ILRNet embeds a low-rank prior for HSI into the deep neural network, enabling to benefit from both model-driven and data-driven methods. These approaches effectively enhance the model's  denoising capability. As a result, ILRNet achieved optimal results under most noise patterns, while maintaining only a very small gap from the optimal results under other noise patterns. We also observe that our ILRNet lags behind in scenarios involving mixture noise. This could be attributed to the fact that TRQ3D and SST utilize transformers to encode nonlocal spatial correlations, enhancing their denoising capabilities, particularly in cases of severe noise.

To comprehensively compare all the methods, Table~\ref{tab:params_time} shows the number of parameters and the average running time on the ICVL testing dataset. The model-driven methods were evaluated on an Intel(R) Xeon(R) Silver 4310 CPU@2.10GHz, while the data-driven methods were evaluated on an NVIDIA GeForce RTX 3090 GPU. Thanks to the speedup provided by GPUs, the data-driven methods consume considerably less time than the model-driven methods. Among the data-driven methods, the difference in running time is slight. Since the performing of SVD, our method consumes slightly more time, but this difference is acceptable considering the performance advantage of ILRNet. Regarding the number of parameters, although T3SC, MAC-Net, TRQ3D, and DPNet-S have fewer parameters than ILRNet, they exhibit a considerable gap in PSNR. While SST has more parameters than ILRNet to achieve higher PSNR, its performance is still inferior to ILRNet. Overall, ILRNet achieves a better balance between parameter count and performance.

To more intuitively and comprehensively compare the denoising effects of all methods, we extracted the bands 23, 15, and 9 from the clean HSI, the noisy HSI, and the denoising results of each method to generate false-color images, which are shown in Fig.~\ref{fig:ICVL_visual}. As shown in Fig.~\ref{fig:ICVL_visual}\subref{fig:Master5000K_clean_} and Fig.~\ref{fig:ICVL_visual}\subref{fig:Master5000K_noise_}, the noise severely degrades the clean HSI, posing a significant challenge to denoising. Consistent with the results in Table~\ref{tab:icvl}, data-driven methods show a clear advantage over model-driven methods. The denoising results of most model-driven methods have poor visual quality, characterized by residual noise, blurring, or color distortion. Due to the modeling of physical priors such as spectral low-rankness and sparsity of HSI, E-3DTV, T3SC, MAC-Net and DPNet-S achieve strong denoising capabilities, and their denoising results showed significant improvement visually. However, color distortion and blurring issues constrain further improvement in visual quality. These issues still exist in methods based on attention mechanisms such as TRQ3D and SST. Due to ILRNet's modeling of the low-rank prior of HSI and the effective supplementation of details using an iterative refinement process, it not only effectively removes noise but also significantly improves detail recovery. Based on observations of the enlarged regions of the images, ILRNet successfully recover details that were difficult for the competing methods to recover. In conclusion, ILRNet demonstrates significantly superior restoration quality compared to the compared methods in visual comparison.

Moreover, in Fig.~\ref{fig:ICVL_reflectance}, we compare the spectral reflectance curves at pixel (340, 390) of the HSIs shown in Fig.~\ref{fig:ICVL_visual}. Similarly, model-driven methods except E-3DTV fail to recover clean results from severely corrupted spectral reflectance curve. Although data-driven methods such as T3SC, TRQ3D, and SST effectively remove noise, the results recovered by T3SC were notably inconsistent with the clean spectral reflectance curve, and those recovered by E-3DTV, TRQ3D, SST and DPNet-S were excessively smoothed. In contrast, MAC-Net and ILRNet successfully recover results consistent with the clean spectral reflectance curve. In summary, ILRNet also demonstrates superiority in spectral reflectance  restoration.

\subsubsection{Pavia City Center HSI}

The Pavia City Center HSI was acquired using a Reflective Optics System Imaging Spectrometer (ROSIS-3) over the city of Pavia, Italy. It covers a spectral range from 430 to 860 nm. Following the experimental setup in \cite{Zhang2014LRMR}, we selected a clean sub-image of size $200 \times 200 \times 80$ as the ground truth and added noise for testing. We used the model trained on the ICVL dataset for evaluation.

\begin{table*}[!t]
        \caption{Comparison of Different Methods on Pavia City Center HSI. The Top Three Values Are Marked as \1{Red}, \2{Blue}, And \3{Green}.}\label{tab:pavia}
        \centering
        \resizebox{\linewidth}{!}{
                \tablesize{
        \begin{tabular}{c|c|c|c|c|c|c|c|c|c|c|c|c|c|c|c}
                \Xhline{1.2pt}
        \multirow{3}*{$\sigma$}&\multirow{3}*{Index}&\multirow{3}*{\makebox[0.06\textwidth][c]{Noisy}}
        &\multicolumn{7}{c|}{\textbf{Model-driven methods}}&\multicolumn{6}{c}{\textbf{Data-driven methods}}\\
        \cline{4-16}
        &&&\multirow{1}*{\makebox[0.06\textwidth][c]{BM4D}}&\multirow{1}*{\makebox[0.06\textwidth][c]{MTSNMF}}&\multirow{1}*{\makebox[0.06\textwidth][c]{LLRT}}&\multirow{1}*{\makebox[0.06\textwidth][c]{NGMeet}}&\multirow{1}*{\makebox[0.06\textwidth][c]{LRMR}}&\multirow{1}*{\makebox[0.06\textwidth][c]{E-3DTV}}&\multirow{1}*{\makebox[0.06\textwidth][c]{3DlogTNN}}&\multirow{1}*{\makebox[0.06\textwidth][c]{T3SC}}&\multirow{1}*{\makebox[0.06\textwidth][c]{MAC-Net}}&\multirow{1}*{\makebox[0.06\textwidth][c]{TRQ3D}}&\multirow{1}*{\makebox[0.06\textwidth][c]{SST}}&\multirow{1}*{\makebox[0.06\textwidth][c]{DPNet-S}}&\makebox[0.06\textwidth][c]{\textbf{ILRNet}}\\
        &&&\multirow{1}*{\makebox[0.06\textwidth][c]{\cite{Maggioni2013BM4D}}}&\multirow{1}*{\makebox[0.06\textwidth][c]{\cite{Ye2015MTSNMF}}}&\multirow{1}*{\makebox[0.06\textwidth][c]{\cite{Chang2017LLRT}}}&\multirow{1}*{\makebox[0.06\textwidth][c]{\cite{He2022NGMeet}}}&\multirow{1}*{\makebox[0.06\textwidth][c]{\cite{Zhang2014LRMR}}}&\multirow{1}*{\makebox[0.06\textwidth][c]{\cite{Peng2020E-3DTV}}}&\multirow{1}*{\makebox[0.06\textwidth][c]{\cite{Zheng20203DlogTNN}}}&\multirow{1}*{\makebox[0.06\textwidth][c]{\cite{bodrito2021T3SC}}}&\multirow{1}*{\makebox[0.06\textwidth][c]{\cite{Xiong2022MAC-Net}}}&\multirow{1}*{\makebox[0.06\textwidth][c]{\cite{Pang2022TRQ3DNet}}}&\multirow{1}*{\makebox[0.06\textwidth][c]{\cite{li2022spatialspectral}}}&\multirow{1}*{\makebox[0.06\textwidth][c]{\cite{Xiongdpnet}}}&\makebox[0.06\textwidth][c]{\textbf{(Ours)}}\\
        \Xhline{1.2pt}     		
        \multirow{3}*{\textbf{[0,15]}}
        & PSNR$\uparrow$  & 32.27 & 38.90 & 40.78 & 34.82 & 39.73 & 37.34 & 40.12 & 41.64 & 41.15 & \2{47.25} & 34.64 & 44.30 & \3{45.76} & \1{47.45} \\
        & SSIM$\uparrow$  & .8400 & .9739 & .9879 & .8973 & .9598 & .9761 & .9845 & .9907 & .9741 & \1{.9966} & .9567 & .9940 & \3{.9954} & \2{.9965} \\
        & SAM$\downarrow$ & .2620 & .0950 & .0475 & .2006 & .1282 & .0800 & .0575 & .0456 & .1146 & \2{.0352} & .0707 & .0519 & \3{.0406} & \1{.0351} \\
        \hline 		
        \multirow{3}*{\textbf{[0,55]}}
        & PSNR$\uparrow$  & 22.05 & 32.39 & 34.82 & 24.74 & 31.75 & 32.77 & 35.09 & 35.76 & 36.05 & \2{40.88} & 33.36 & 37.09 & \3{39.37} & \1{41.52} \\
        & SSIM$\uparrow$  & .4920 & .9090 & .9512 & .6085 & .8786 & .9382 & .9552 & .9662 & .9304 & \2{.9864} & .9414 & .9563 & \3{.9818} & \1{.9880} \\
        & SAM$\downarrow$ & .6048 & .1628 & .0841 & .4734 & .3192 & .1203 & .0870 & .0769 & .1682 & \2{.0588} & .1030 & .1467 & \3{.0694} & \1{.0554} \\
        \hline
        \multirow{3}*{\textbf{[0,95]}}
        & PSNR$\uparrow$  & 18.92 & 29.16 & 31.49 & 20.50 & 28.60 & 31.21 & 33.16 & 31.62 & 34.67 & \3{36.63} & 32.62 & 36.25 & \2{39.28} & \1{41.80} \\
        & SSIM$\uparrow$  & .3725 & .8186 & .8964 & .4564 & .8219 & .9169 & .9330 & .8656 & .9193 & \3{.9747} & .9302 & .9391 & \2{.9830} & \1{.9895} \\
        & SAM$\downarrow$ & .8195 & .2049 & .1258 & .6749 & .3906 & .1377 & .0989 & .2818 & .1813 & \3{.0830} & .1157 & .1781 & \2{.0714} & \1{.0558} \\
        \Xhline{1.2pt}  		
        \multirow{3}*{\textbf{Mixture}}
        & PSNR$\uparrow$  & 13.45 & 23.47 & 25.16 & 16.83 & 23.70 & 28.90 & 30.43 & 23.96 & 28.66 & 27.73 & 28.23 & \3{31.86} & \2{34.60} & \1{36.17} \\
        & SSIM$\uparrow$  & .2180 & .6010 & .7409 & .3193 & .7176 & .8808 & .8940 & .6531 & .8471 & .8724 & .8665 & \3{.9234} & \2{.9519} & \1{.9653} \\
        & SAM$\downarrow$ & .8892 & .3980 & .4086 & .7164 & .4670 & .1649 & \3{.1135} & .5674 & .2280 & .3222 & .1961 & .1676 & \2{.1013} & \1{.0927} \\
        \Xhline{1.2pt}  		
        \multirow{3}*{\textbf{Corr}}
        & PSNR$\uparrow$  & 28.21 & 35.02 & 38.91 & 30.78 & 38.73 & 35.85 & 36.96 & \2{39.21} & 38.75 & 37.41 & 34.33 & 36.84 & \3{39.14} & \1{41.41} \\
        & SSIM$\uparrow$  & .7408 & .9016 & \3{.9814} & .8080 & .9583 & .9694 & .9719 & \2{.9839} & .8506 & .9742 & .9504 & .9754 & .9770 & \1{.9875} \\
        & SAM$\downarrow$ & .3695 & .2183 & \2{.0542} & .3037 & .1508 & .0856 & .0650 & \3{.0556} & .1709 & .0793 & .0886 & .1270 & .0761 & \1{.0525} \\
        \Xhline{1.2pt} 		
        \end{tabular}}}
\end{table*}

\begin{figure*}[!t]
        \centering
        \subfloat[]{\label{fig:pavia_clean_visual}\includegraphics[width=0.12\linewidth]{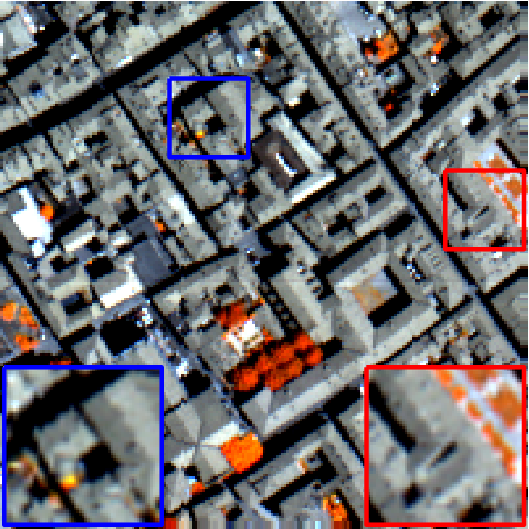}}
        \hspace{-0.8mm}
        \subfloat[]{\label{fig:pavia_noise_visual}\includegraphics[width=0.12\linewidth]{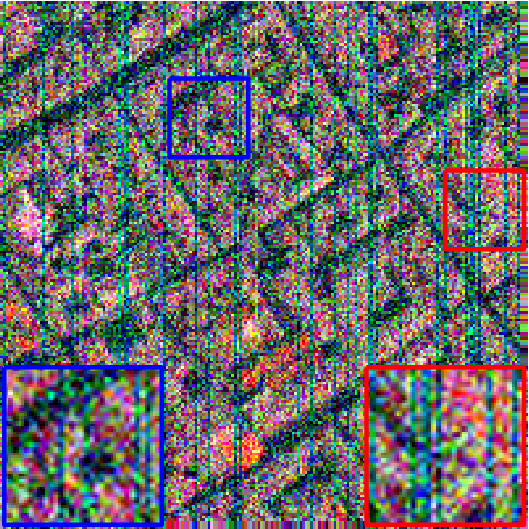}}
        \hspace{-0.8mm}
        \subfloat[]{\label{fig:pavia_BM4D_visual}\includegraphics[width=0.12\linewidth]{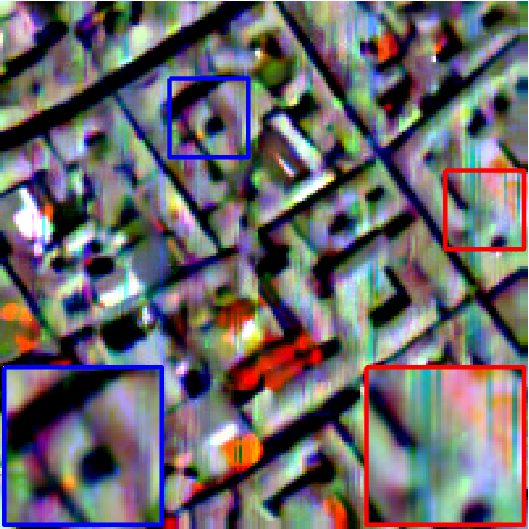}}
        \hspace{-0.8mm}
        \subfloat[]{\label{fig:pavia_MTSNMF_visual}\includegraphics[width=0.12\linewidth]{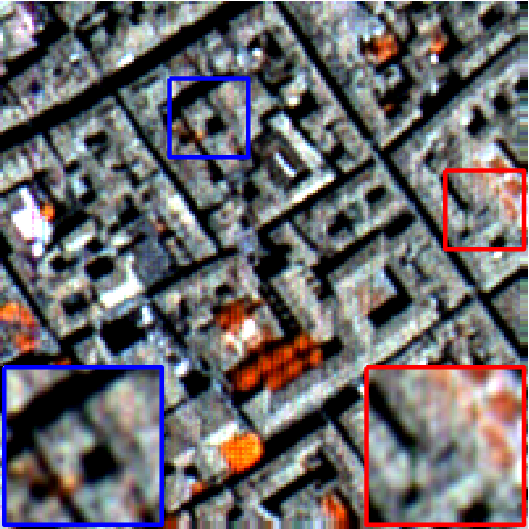}}
        \hspace{-0.8mm}
        \subfloat[]{\label{fig:pavia_LLRT_visual}\includegraphics[width=0.12\linewidth]{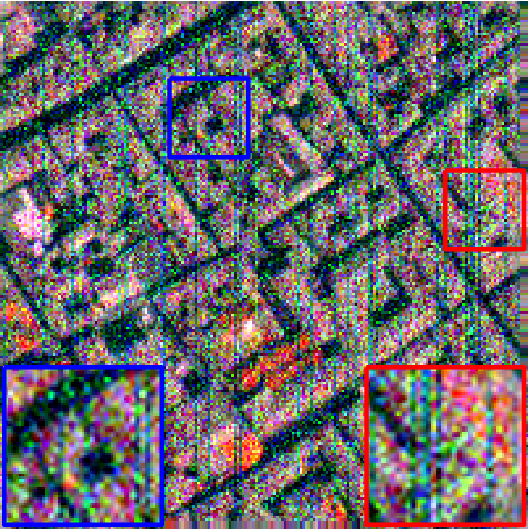}}
        \hspace{-0.8mm}
        \subfloat[]{\label{fig:pavia_NGMeet_visual}\includegraphics[width=0.12\linewidth]{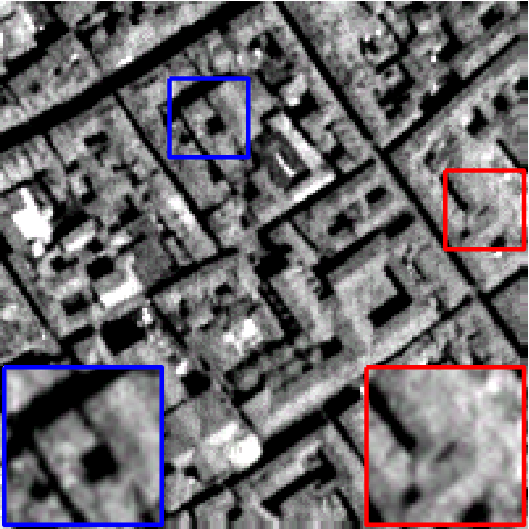}}
        \hspace{-0.8mm}
        \subfloat[]{\label{fig:pavia_LRMR_visual}\includegraphics[width=0.12\linewidth]{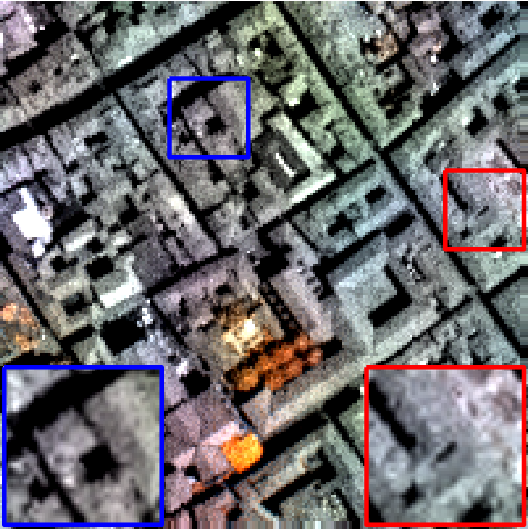}}
        \hspace{-0.8mm}
        \subfloat[]{\label{fig:pavia_E-3DTV_visual}\includegraphics[width=0.12\linewidth]{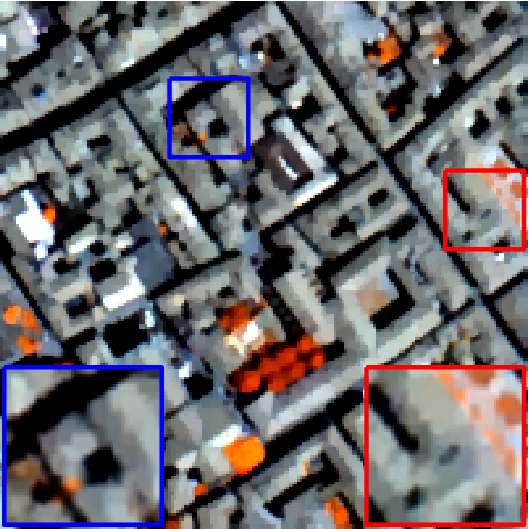}}
        \hspace{-0.8mm}
        \subfloat[]{\label{fig:pavia_3DlogTNN_visual}\includegraphics[width=0.12\linewidth]{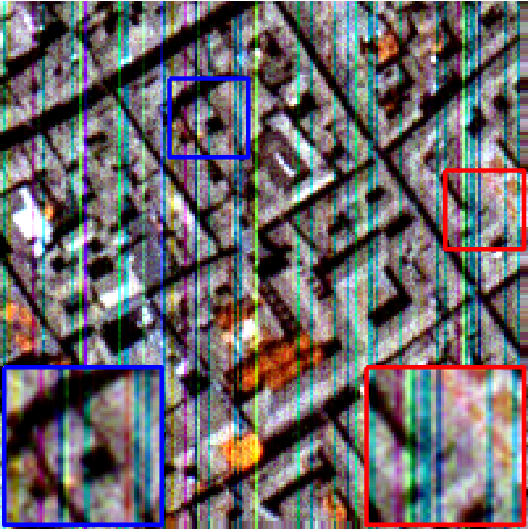}}
        \hspace{-0.8mm}
        \subfloat[]{\label{fig:pavia_T3SC_visual}\includegraphics[width=0.12\linewidth]{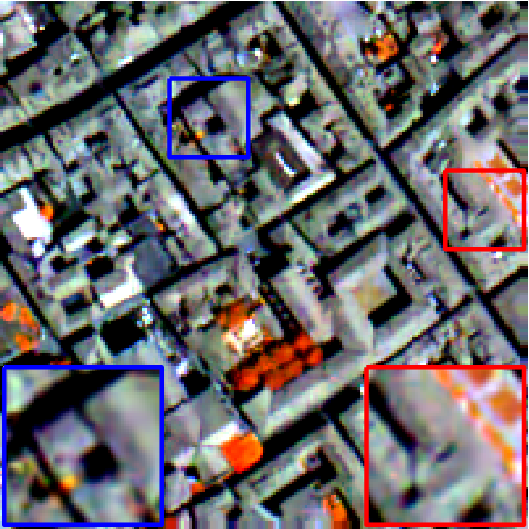}}
        \hspace{-0.8mm}
        \subfloat[]{\label{fig:pavia_MAC-Net_visual}\includegraphics[width=0.12\linewidth]{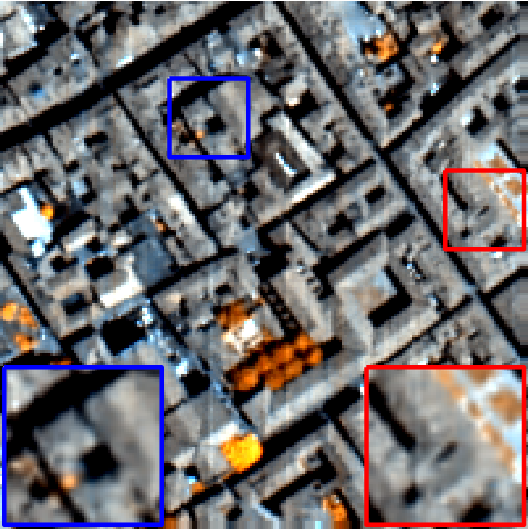}}
        \hspace{-0.8mm}
        \subfloat[]{\label{fig:pavia_TRQ3D_visual}\includegraphics[width=0.12\linewidth]{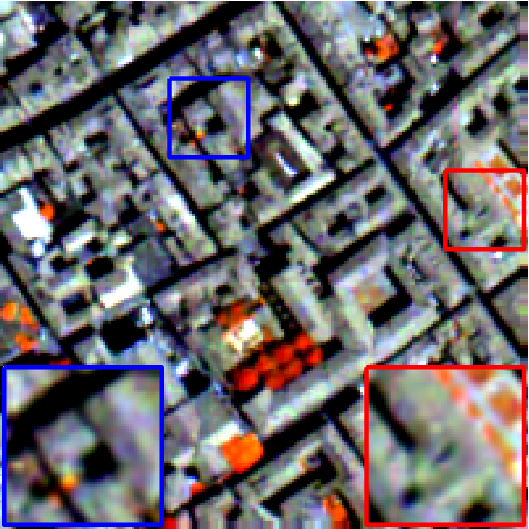}}
        \hspace{-0.8mm}
        \subfloat[]{\label{fig:pavia_SST_visual}\includegraphics[width=0.12\linewidth]{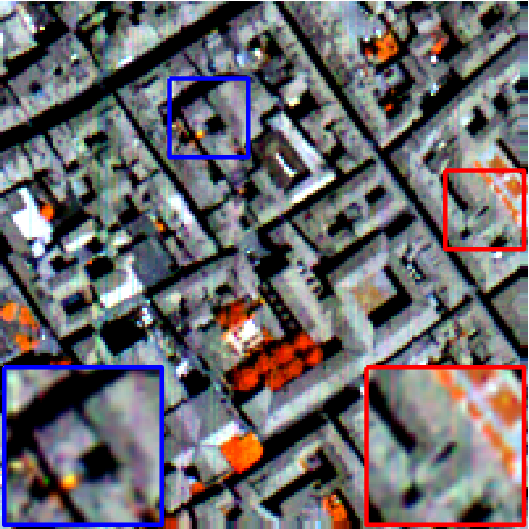}}
        \hspace{-0.8mm}
        \subfloat[]{\label{fig:pavia_DPNet_visual}\includegraphics[width=0.12\linewidth]{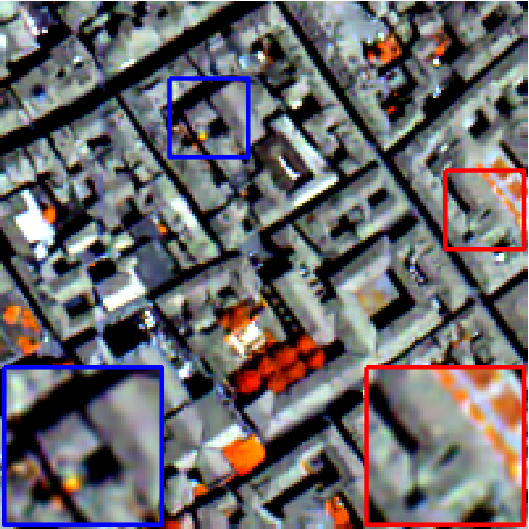}}
        \hspace{-0.8mm}
        \subfloat[]{\label{fig:pavia_ILRNet_visual}\includegraphics[width=0.12\linewidth]{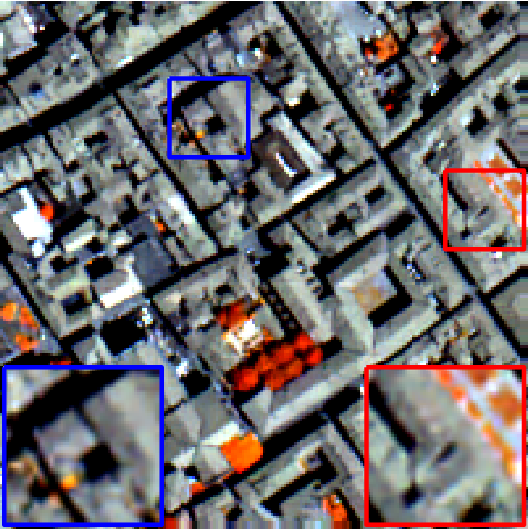}}
        \caption{Denoising results on the the Pavia City Center HSI under the mixture noise. The false-color images are generated by combining bands 70, 50, and 30. (a) Clean. (b) Noisy. (c) BM4D~\cite{Maggioni2013BM4D}. (d) MTSNMF~\cite{Ye2015MTSNMF}. (e) LLRT~\cite{Chang2017LLRT}. (f) NGMeet~\cite{He2022NGMeet}. (g) LRMR~\cite{Zhang2014LRMR}. (h) E-3DTV~\cite{Peng2020E-3DTV}. (i) 3DlogTNN~\cite{Zheng20203DlogTNN}. (j) T3SC~\cite{bodrito2021T3SC}. (k) MAC-Net~\cite{Xiong2022MAC-Net}. (l) TRQ3D~\cite{Pang2022TRQ3DNet}. (m) SST~\cite{li2022spatialspectral}. (n) DPNet-S~\cite{Xiongdpnet}. (o) \textbf{ILRNet}.} \label{fig:pavia_visual}
\end{figure*}

\begin{figure*}[!t]
        \centering
        \subfloat[]{\label{fig:pavia_clean_reflectance}\includegraphics[width=0.1240\linewidth]{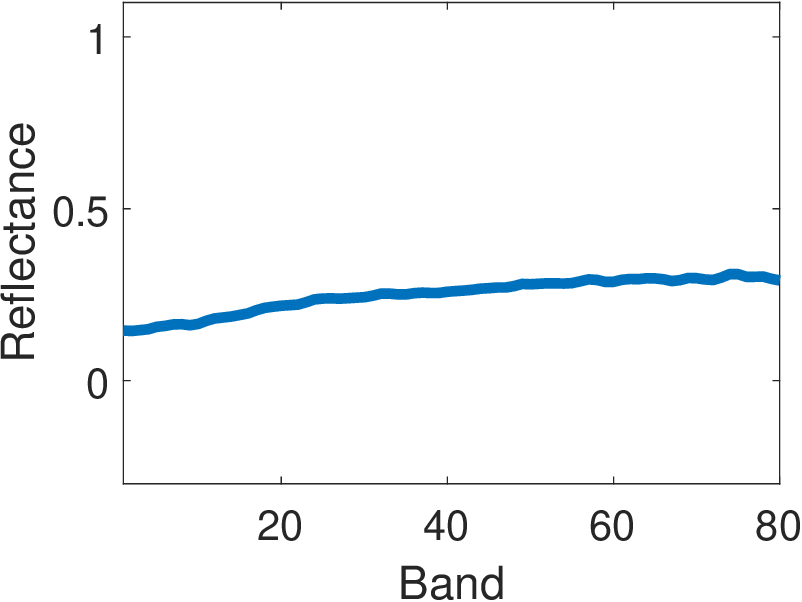}}
        \hspace{-1.1mm}
        \subfloat[]{\label{fig:pavia_noise_reflectance}\includegraphics[width=0.1240\linewidth]{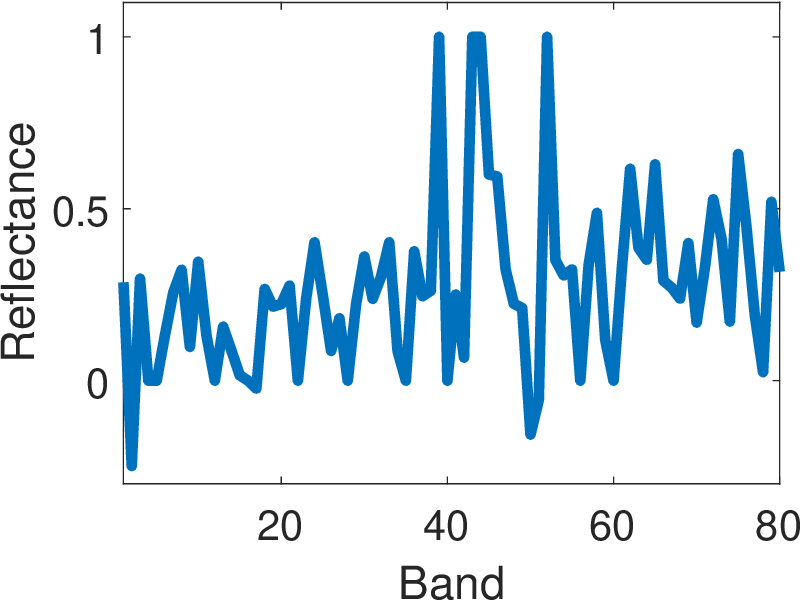}}
        \hspace{-1.1mm}
        \subfloat[]{\label{fig:pavia_BM4D_reflectance}\includegraphics[width=0.1240\linewidth]{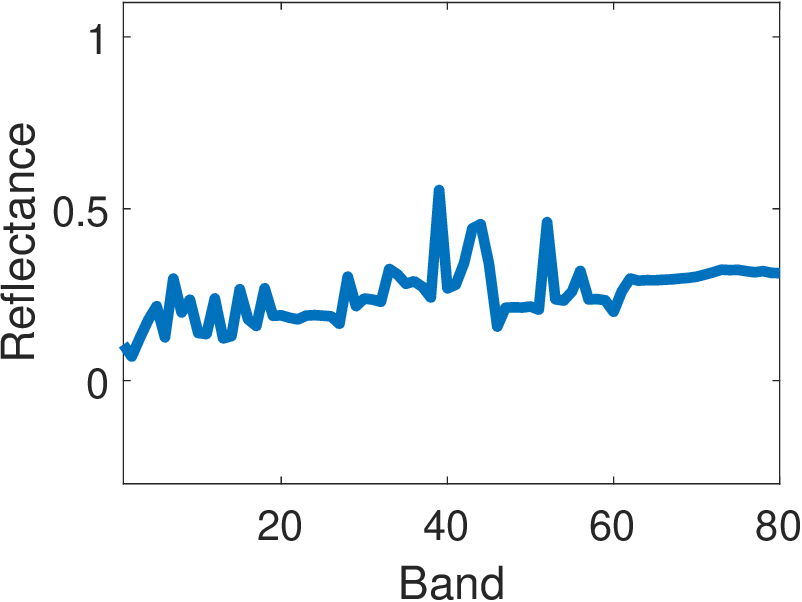}}
        \hspace{-1.1mm}
        \subfloat[]{\label{fig:pavia_MTSNMF_reflectance}\includegraphics[width=0.1240\linewidth]{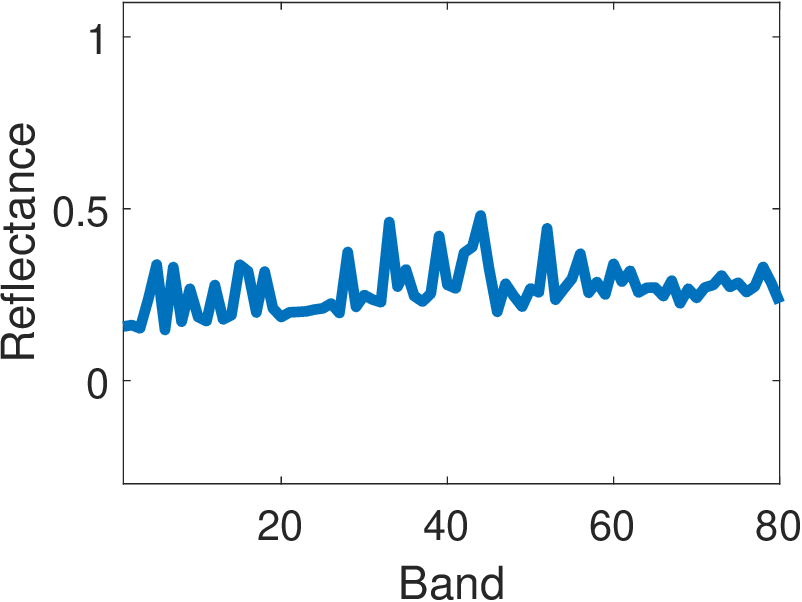}}
        \hspace{-1.1mm}
        \subfloat[]{\label{fig:pavia_LLRT_reflectance}\includegraphics[width=0.1240\linewidth]{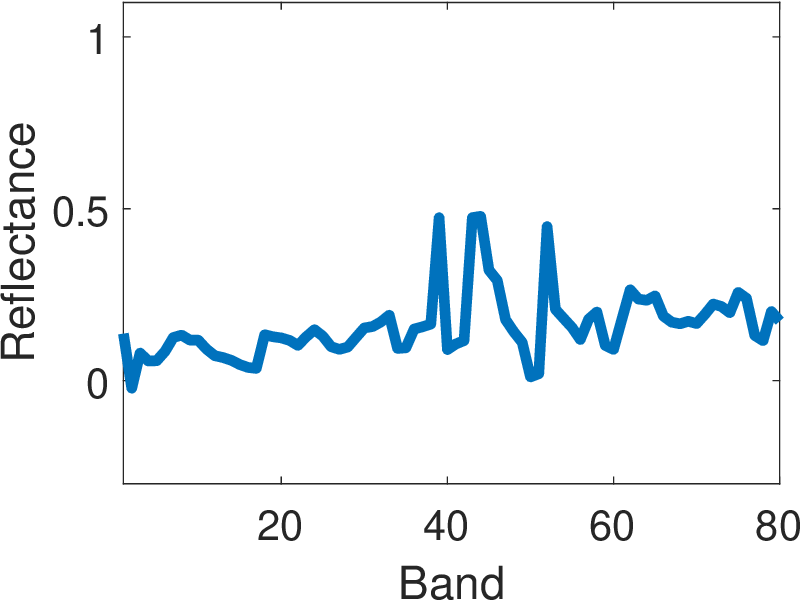}}
        \hspace{-1.1mm}
        \subfloat[]{\label{fig:pavia_NGMeet_reflectance}\includegraphics[width=0.1240\linewidth]{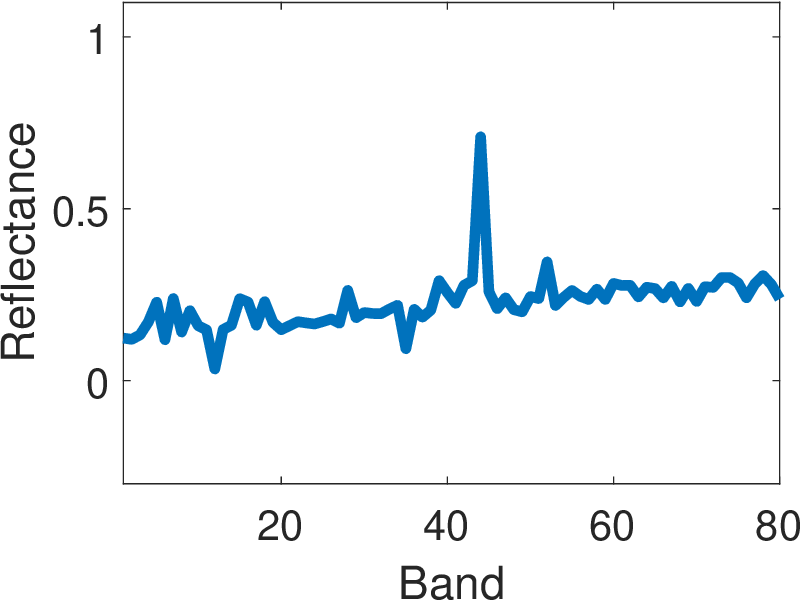}}
        \hspace{-1.1mm}
        \subfloat[]{\label{fig:pavia_LRMR_reflectance}\includegraphics[width=0.1240\linewidth]{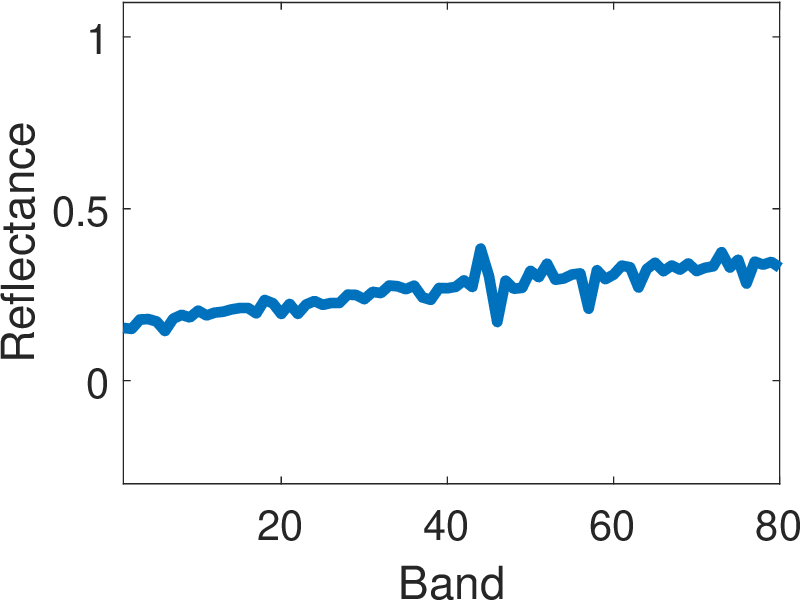}}
        \hspace{-1.1mm}
        \subfloat[]{\label{fig:pavia_E-3DTV_reflectance}\includegraphics[width=0.1240\linewidth]{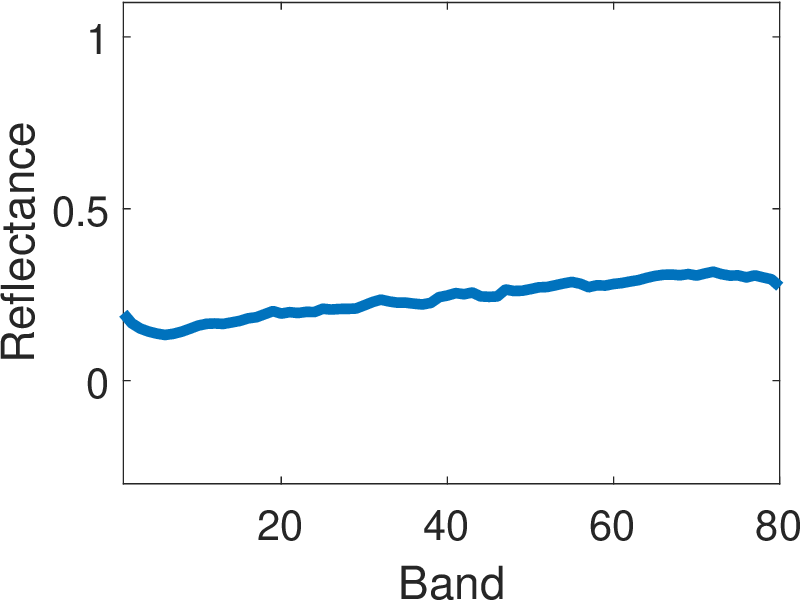}}
        \hspace{-1.1mm}
        \subfloat[]{\label{fig:pavia_3DlogTNN_reflectance}\includegraphics[width=0.1240\linewidth]{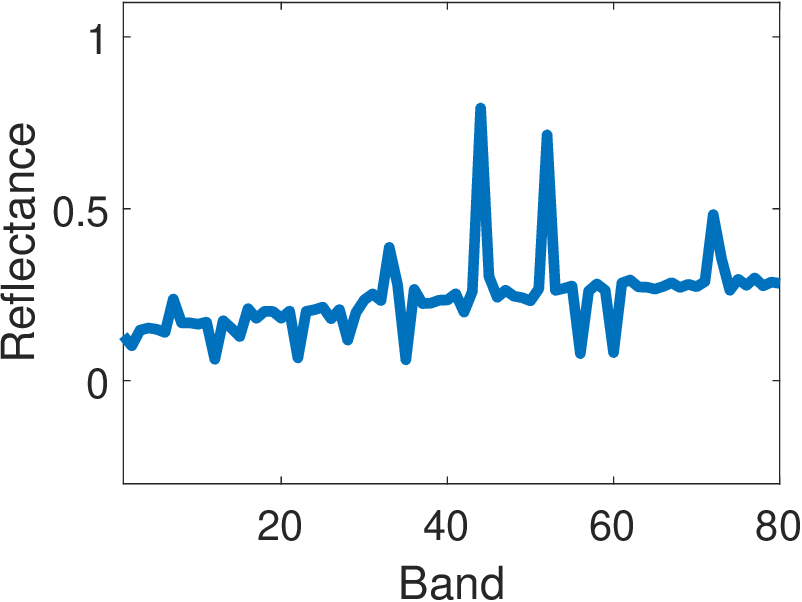}}
        \hspace{-1.1mm}
        \subfloat[]{\label{fig:pavia_T3SC_reflectance}\includegraphics[width=0.1240\linewidth]{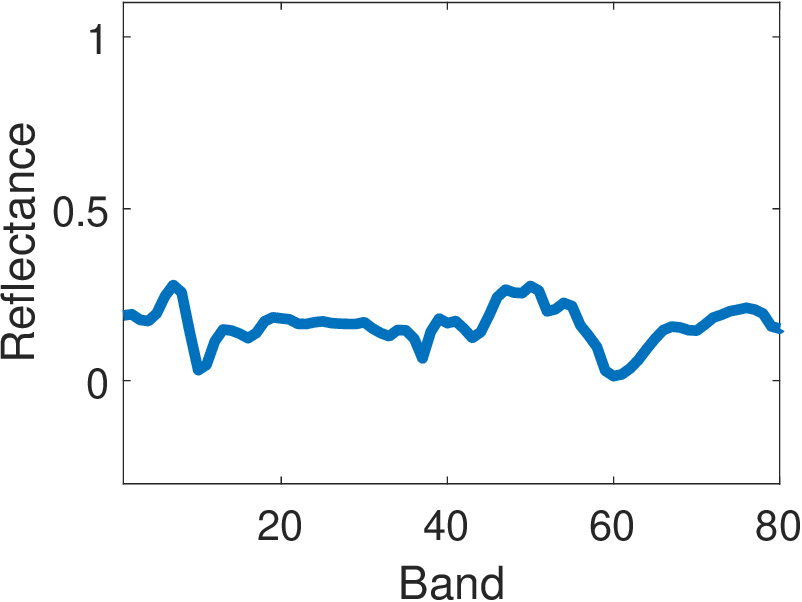}}
        \hspace{-1.1mm}
        \subfloat[]{\label{fig:pavia_MAC-Net_reflectance}\includegraphics[width=0.1240\linewidth]{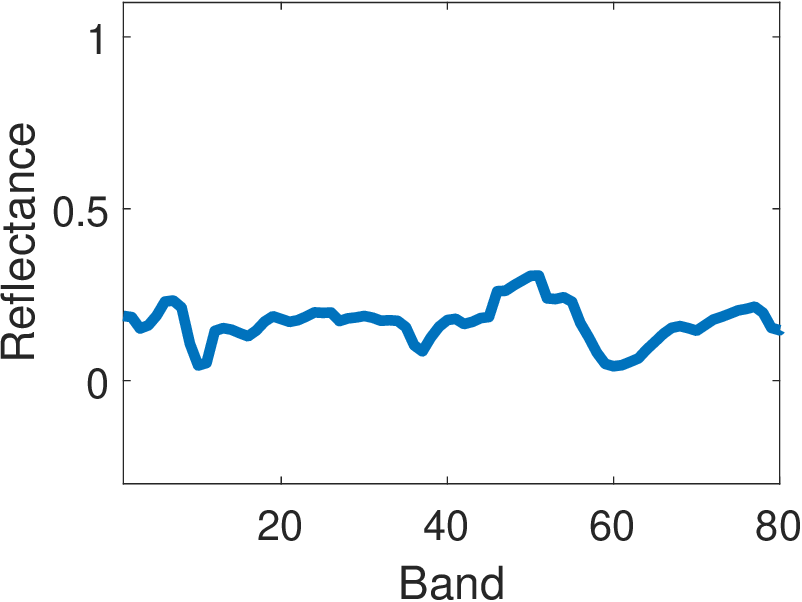}}
        \hspace{-1.1mm}
        \subfloat[]{\label{fig:pavia_TRQ3D_reflectance}\includegraphics[width=0.1240\linewidth]{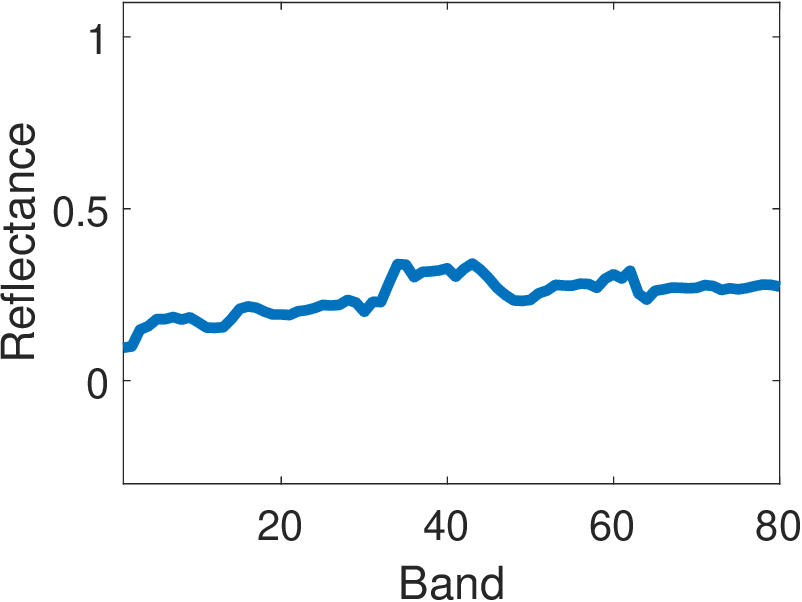}}
        \hspace{-1.1mm}
        \subfloat[]{\label{fig:pavia_SST_reflectance}\includegraphics[width=0.1240\linewidth]{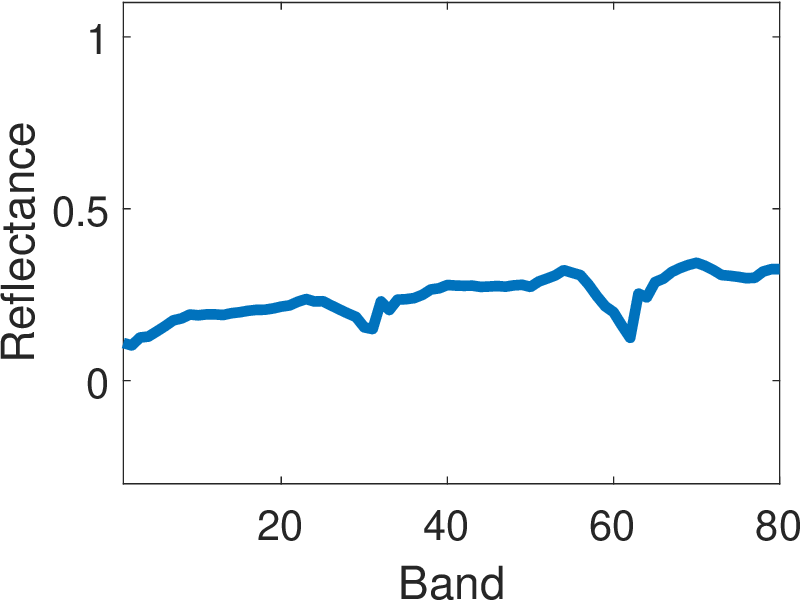}}
        \hspace{-1.1mm}
        \subfloat[]{\label{fig:pavia_DPNet_reflectance}\includegraphics[width=0.1240\linewidth]{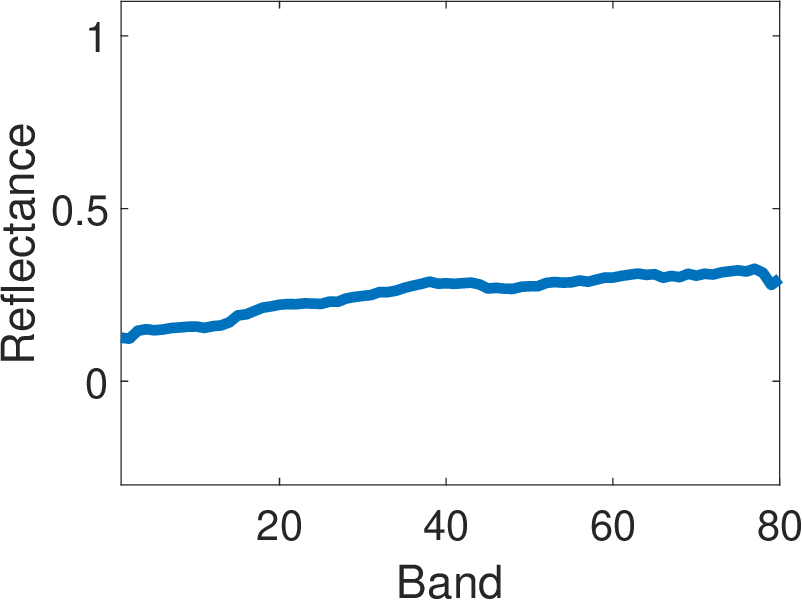}}
        \hspace{-1.1mm}
        \subfloat[]{\label{fig:pavia_ILRNet_reflectance}\includegraphics[width=0.1240\linewidth]{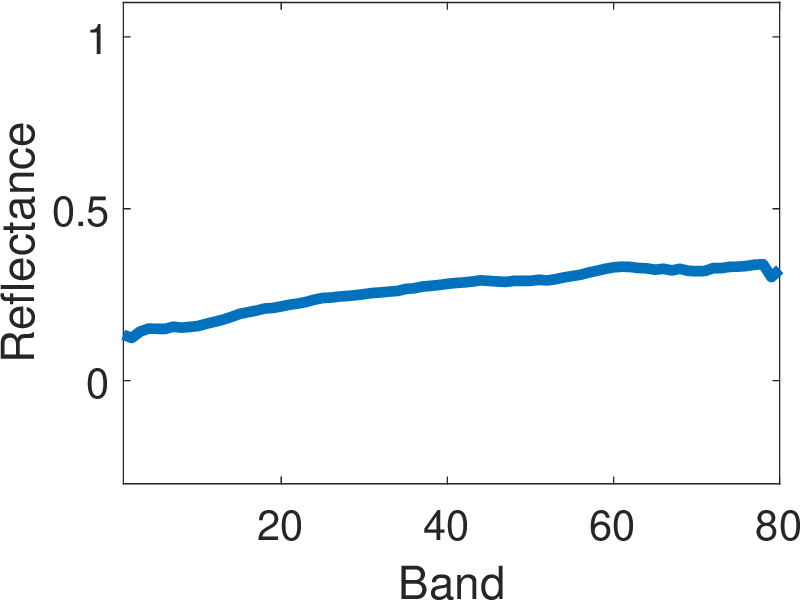}}
        \caption{Reflectance of pixel (111, 11) in the Pavia City Center HSI under the mixture noise. (a) Clean. (b) Noisy. (c) BM4D~\cite{Maggioni2013BM4D}. (d) MTSNMF~\cite{Ye2015MTSNMF}. (e) LLRT~\cite{Chang2017LLRT}. (f) NGMeet~\cite{He2022NGMeet}. (g) LRMR~\cite{Zhang2014LRMR}. (h) E-3DTV~\cite{Peng2020E-3DTV}. (i) 3DlogTNN~\cite{Zheng20203DlogTNN}. (j) T3SC~\cite{bodrito2021T3SC}. (k) MAC-Net~\cite{Xiong2022MAC-Net}. (l) TRQ3D~\cite{Pang2022TRQ3DNet}. (m) SST~\cite{li2022spatialspectral}. (n) DPNet-S~\cite{Xiongdpnet}. (o) \textbf{ILRNet}.} \label{fig:pavia_reflectance}
\end{figure*}

Table~\ref{tab:pavia} quantitatively compares the denoising results of different methods. Since all data-driven methods are trained on the ICVL dataset, differences in imaging environments and the number of bands pose challenges for data-driven methods. This is manifested in varying degrees of degradation in indexes for data-driven methods. Conversely, because model-driven methods directly model the physical priors of HSI without relying on data, some model-driven methods show performance comparable to data-driven methods. Among them, E-3DTV demonstrates significant advantages in mixture noise removal experiments, with its indexes even surpassing those of some data-driven methods. Thanks to its modeling of spectral low-rank and sparse prior of HSI, MAC-Net and DPNet-S exhibit a certain degree of robustness and outperforms methods based on attention mechanisms like TRQ3D and SST under non-i.i.d. Gaussian noise patterns. Benefiting from the modeling of low-rank prior of HSI by the RMM and the iterative refinement process, ILRNet also demonstrates strong robustness, exhibiting significant advantages or even leading in performance under all noise patterns.
\begin{figure*}[!t]
        \centering
        \subfloat[]{\label{fig:eo1_clean_visual}\includegraphics[width=0.1420\linewidth]{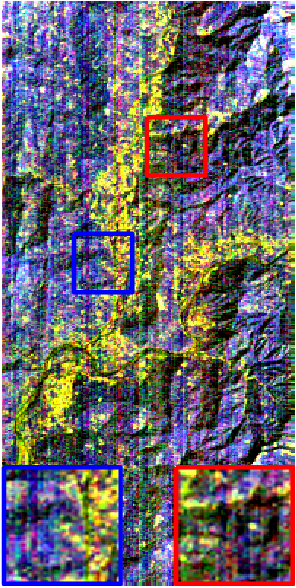}}
        \hspace{-1.1mm}
        \subfloat[]{\label{fig:eo1_BM4D_visual}\includegraphics[width=0.1420\linewidth]{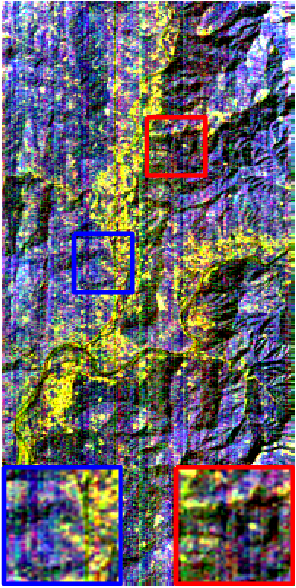}}
        \hspace{-1.1mm}
        \subfloat[]{\label{fig:eo1_MTSNMF_visual}\includegraphics[width=0.1420\linewidth]{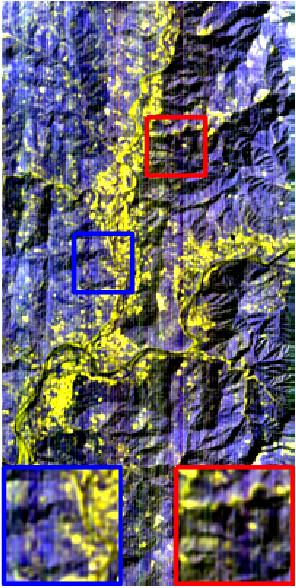}}
        \hspace{-1.1mm}
        \subfloat[]{\label{fig:eo1_LLRT_visual}\includegraphics[width=0.1420\linewidth]{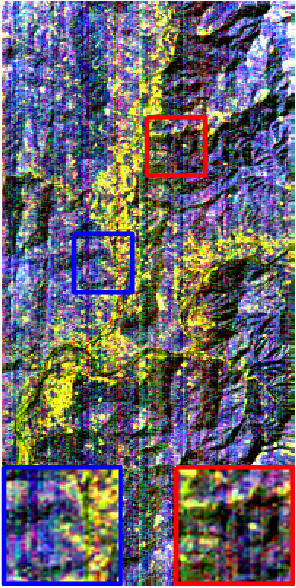}}
        \hspace{-1.1mm}
        \subfloat[]{\label{fig:eo1_NGMeet_visual}\includegraphics[width=0.1420\linewidth]{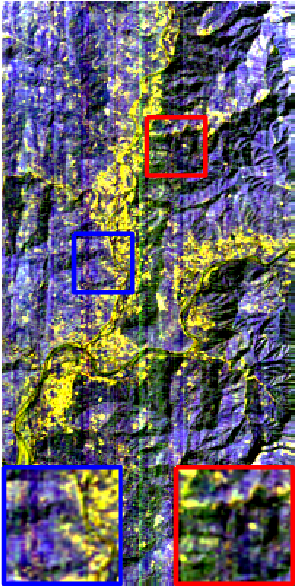}}
        \hspace{-1.1mm}
        \subfloat[]{\label{fig:eo1_LRMR_visual}\includegraphics[width=0.1420\linewidth]{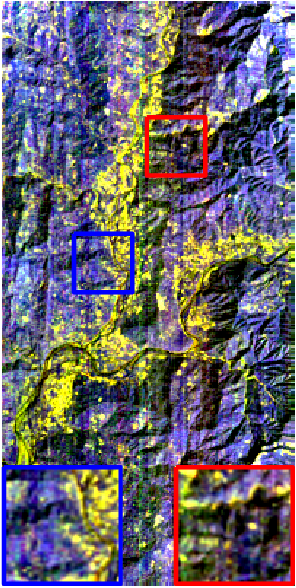}}
        \hspace{-1.1mm}
        \subfloat[]{\label{fig:eo1_E-3DTV_visual}\includegraphics[width=0.1420\linewidth]{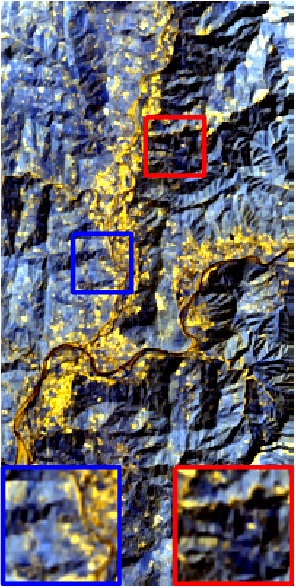}}
        \hspace{-1.1mm}
        \subfloat[]{\label{fig:eo1_3DlogTNN_visual}\includegraphics[width=0.1420\linewidth]{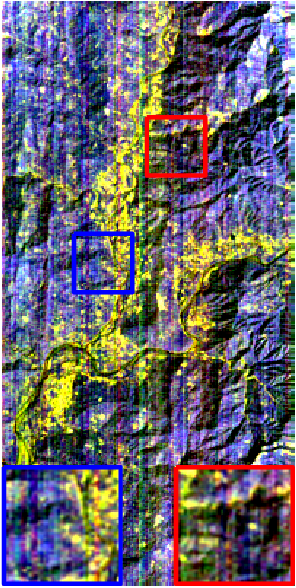}}
        \hspace{-1.1mm}
        \subfloat[]{\label{fig:eo1_T3SC_visual}\includegraphics[width=0.1420\linewidth]{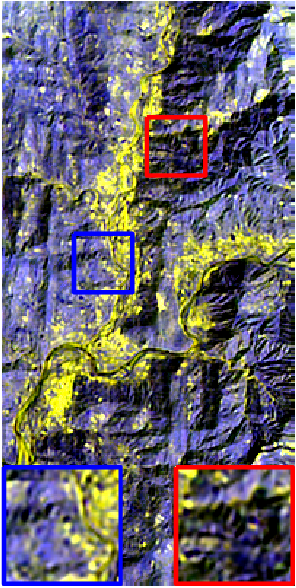}}
        \hspace{-1.1mm}
        \subfloat[]{\label{fig:eo1_MAC-Net_visual}\includegraphics[width=0.1420\linewidth]{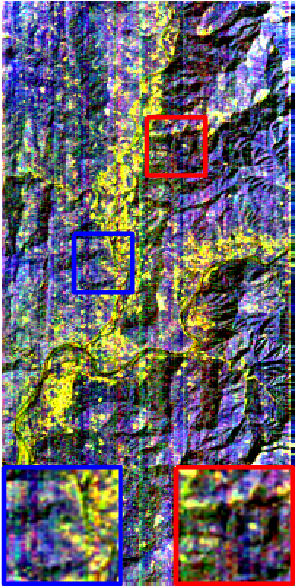}}
        \hspace{-1.1mm}
        \subfloat[]{\label{fig:eo1_TRQ3D_visual}\includegraphics[width=0.1420\linewidth]{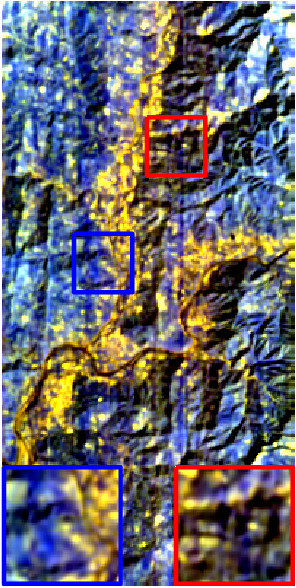}}
        \hspace{-1.1mm}
        \subfloat[]{\label{fig:eo1_SST_visual}\includegraphics[width=0.1420\linewidth]{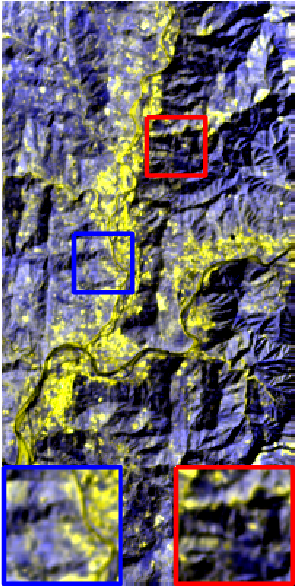}}
        \hspace{-1.1mm}
        \subfloat[]{\label{fig:eo1_DPNet_visual}\includegraphics[width=0.1420\linewidth]{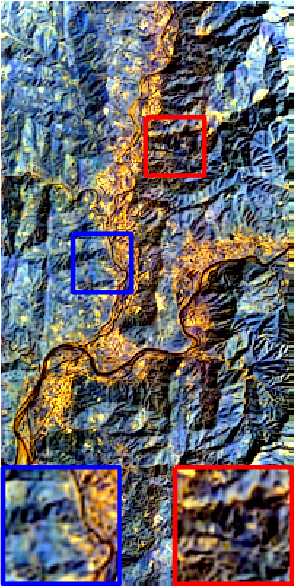}}
        \hspace{-1.1mm}
        \subfloat[]{\label{fig:eo1_ILRNet_visual}\includegraphics[width=0.1420\linewidth]{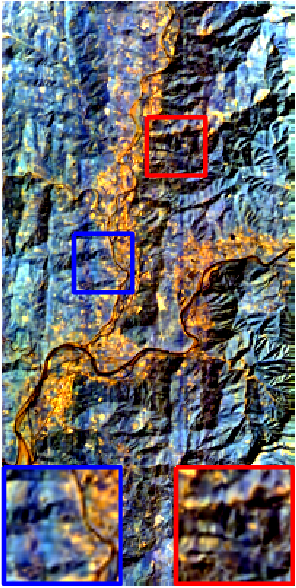}}
        \caption{Denoising results on the EO-1 HSI. The false-color images are generated by combining bands 97, 95, and 73. (a) Noisy. (b) BM4D~\cite{Maggioni2013BM4D}. (c) MTSNMF~\cite{Ye2015MTSNMF}. (d) LLRT~\cite{Chang2017LLRT}. (e) NGMeet~\cite{He2022NGMeet}. (f) LRMR~\cite{Zhang2014LRMR}. (g) E-3DTV~\cite{Peng2020E-3DTV}. (h) 3DlogTNN~\cite{Zheng20203DlogTNN}. (i) T3SC~\cite{bodrito2021T3SC}. (j) MAC-Net~\cite{Xiong2022MAC-Net}. (k) TRQ3D~\cite{Pang2022TRQ3DNet}. (l) SST~\cite{li2022spatialspectral}. (m) DPNet-S~\cite{Xiongdpnet}. (n) \textbf{ILRNet}.} \label{fig:eo1_visual}
\end{figure*}

Fig.~\ref{fig:pavia_visual} extracts bands 70, 50, and 30 from the Pavia City Center HSI to generate false-color images for visual comparison where the mixture noise is considered. Most model-driven methods are designed to deal with Gaussian noise and therefore cannot effectively remove strip noise and deadline noise. In contrast, the denoising results of data-driven methods have less noise residue. However, color distortion and blurring issues are prevalent in both model-driven and data-driven methods, leading to poor visual quality of their restoration results. Due to the iterative refinement process's supplementation of details, ILRNet's restoration result exhibits higher visual quality.

Fig.~\ref{fig:pavia_reflectance} presents the spectral reflectance curves at pixel (111, 11) of the HSIs shown in Fig.~\ref{fig:pavia_visual}. The spectral reflectance curves recovered by model-driven methods as well as most data-driven methods such as T3SC and MAC-Net exhibit noticeable residual noise. Due to the significant difference in the number of bands between the training and testing HSIs, the spectral reflectance curves recovered by TRQ3D and SST exhibit significant mismatches with the clean HSI. In contrast, DPNet-S and ILRNet recovers spectral curve consistent with the clean HSI, demonstrating strong robustness.

\subsection{Real-world Noise Removal}

We select two real-world remote sensing HSIs to comprehensively evaluate the denoising capabilities of all methods. Due to the lack of clean counterparts for real-world remote sensing HSIs, we chose to qualitatively compare the denoising results.

\subsubsection{Earth Observing-1 (EO-1) HSI}

\begin{figure*}[!t]
        \centering
        \subfloat[]{\label{fig:eo1_noise_reflectance}\includegraphics[width=0.1420\linewidth]{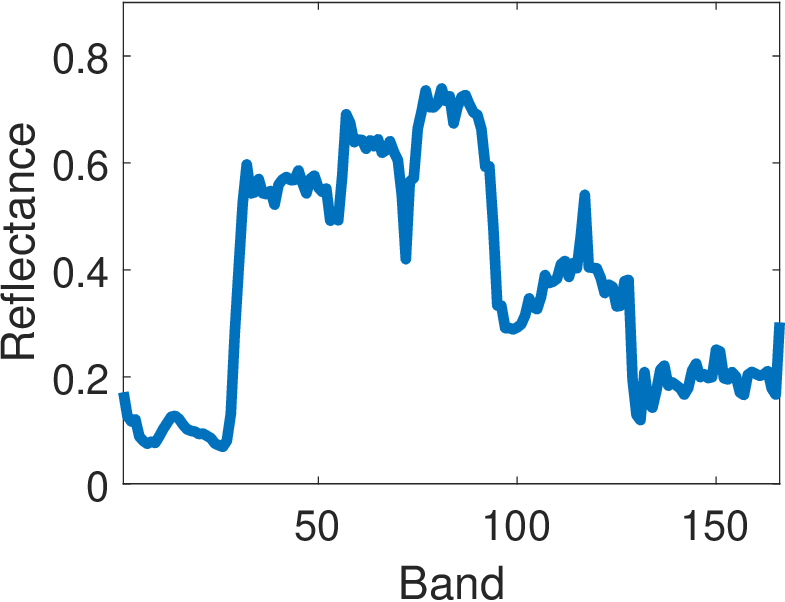}}
        \hspace{-1.1mm}
        \subfloat[]{\label{fig:eo1_BM4D_reflectance}\includegraphics[width=0.1420\linewidth]{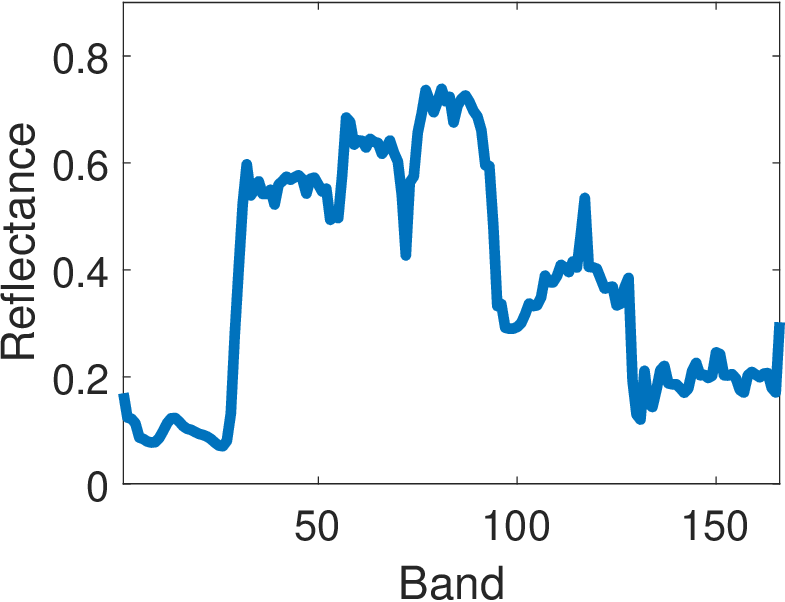}}
        \hspace{-1.1mm}
        \subfloat[]{\label{fig:eo1_MTSNMF_reflectance}\includegraphics[width=0.1420\linewidth]{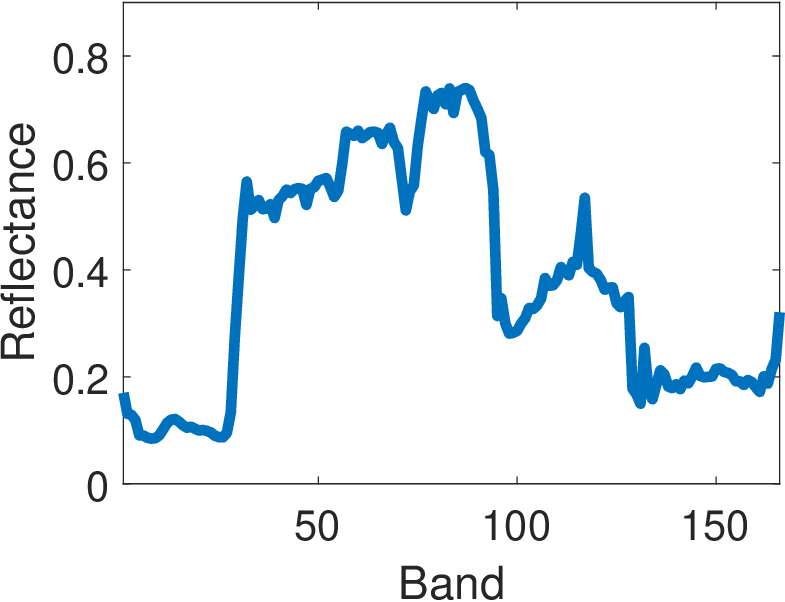}}
        \hspace{-1.1mm}
        \subfloat[]{\label{fig:eo1_LLRT_reflectance}\includegraphics[width=0.1420\linewidth]{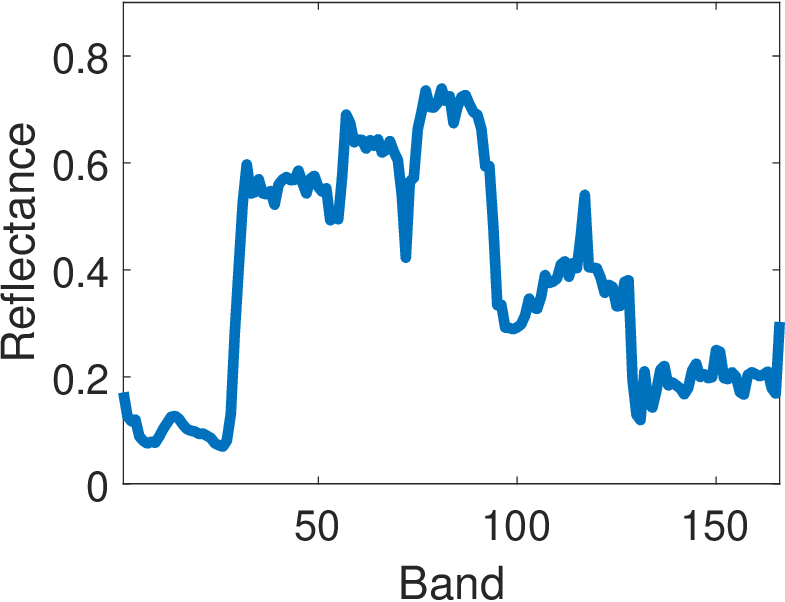}}
        \hspace{-1.1mm}
        \subfloat[]{\label{fig:eo1_NGMeet_reflectance}\includegraphics[width=0.1420\linewidth]{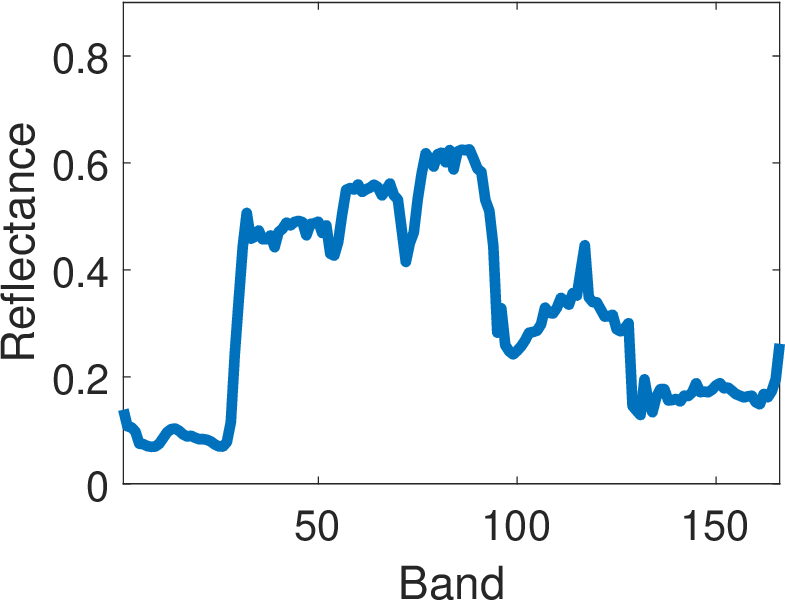}}
        \hspace{-1.1mm}
        \subfloat[]{\label{fig:eo1_LRMR_reflectance}\includegraphics[width=0.1420\linewidth]{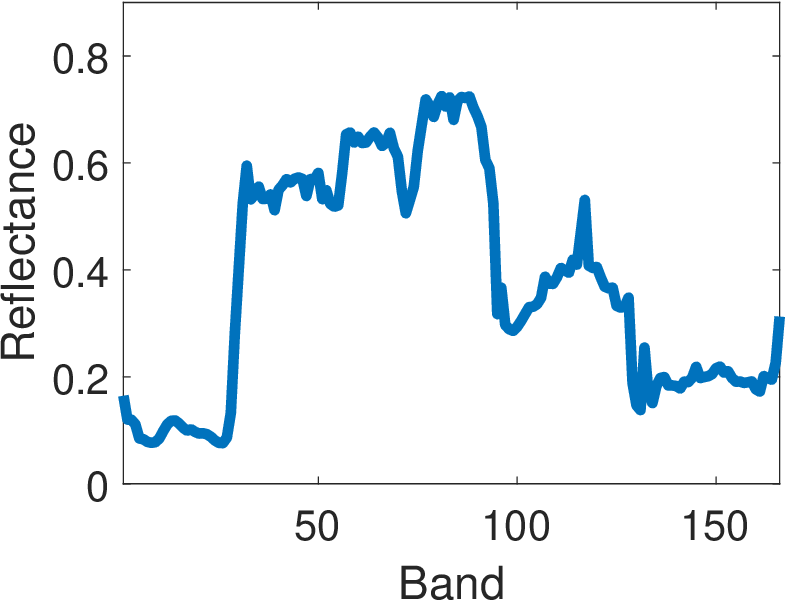}}
        \hspace{-1.1mm}
        \subfloat[]{\label{fig:eo1_E-3DTV_reflectance}\includegraphics[width=0.1420\linewidth]{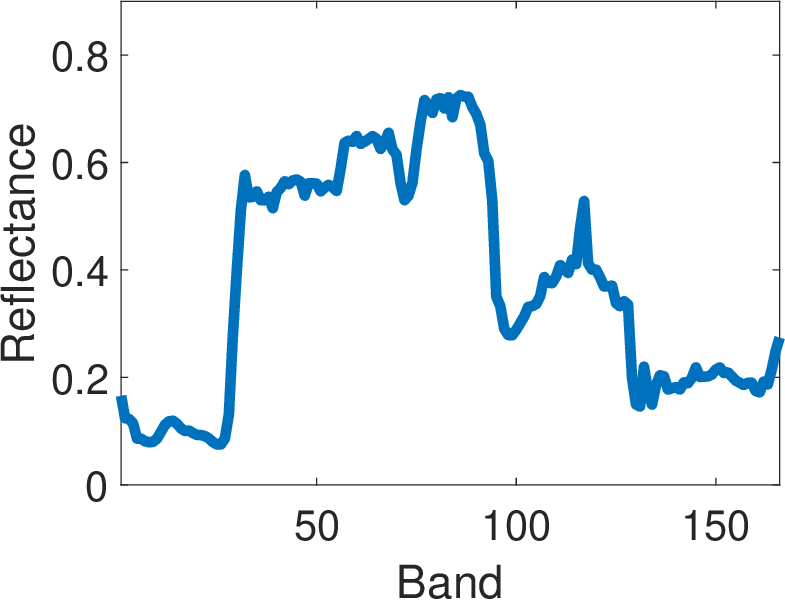}}
        \hspace{-1.1mm}
        \subfloat[]{\label{fig:eo1_3DlogTNN_reflectance}\includegraphics[width=0.1420\linewidth]{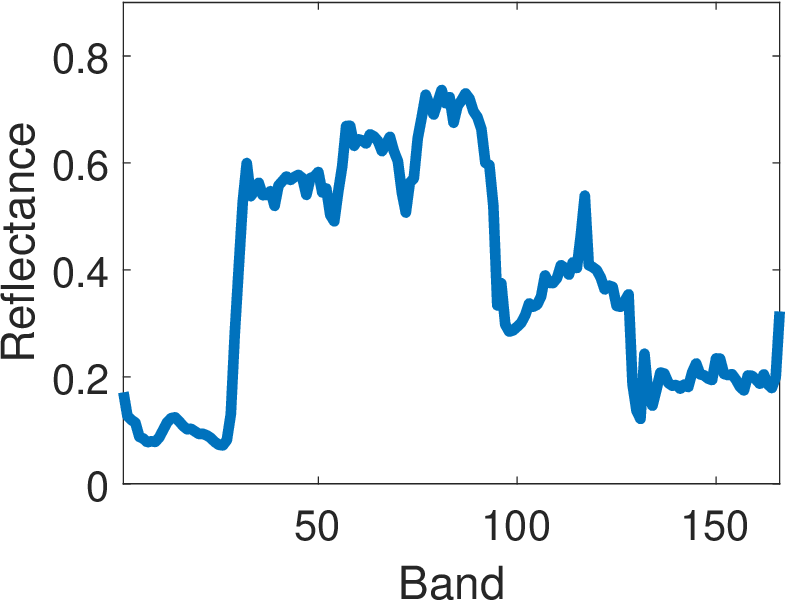}}
        \hspace{-1.1mm}
        \subfloat[]{\label{fig:eo1_T3SC_reflectance}\includegraphics[width=0.1420\linewidth]{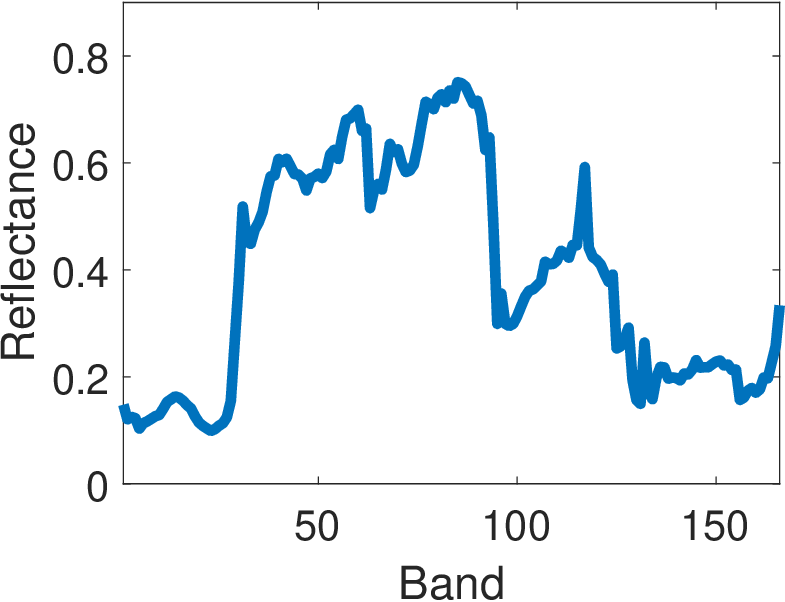}}
        \hspace{-1.1mm}
        \subfloat[]{\label{fig:eo1_MAC-Net_reflectance}\includegraphics[width=0.1420\linewidth]{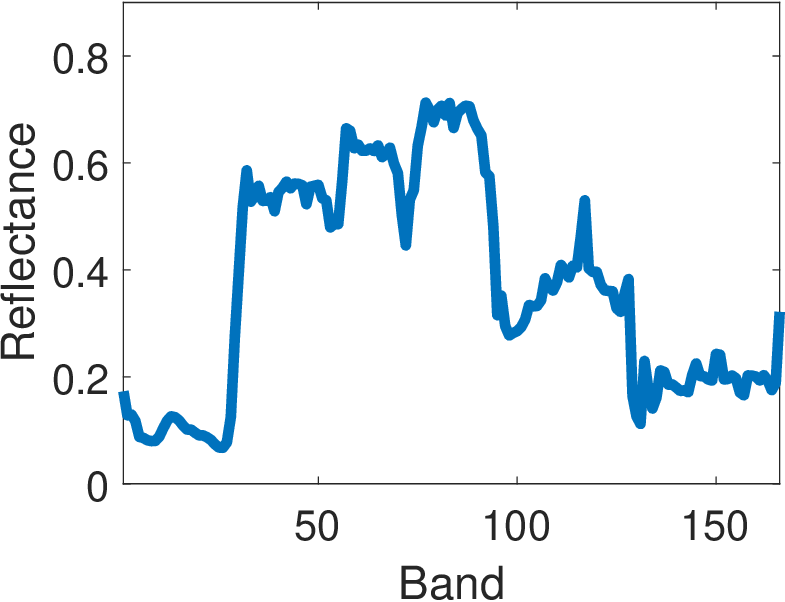}}
        \hspace{-1.1mm}
        \subfloat[]{\label{fig:eo1_TRQ3D_reflectance}\includegraphics[width=0.1420\linewidth]{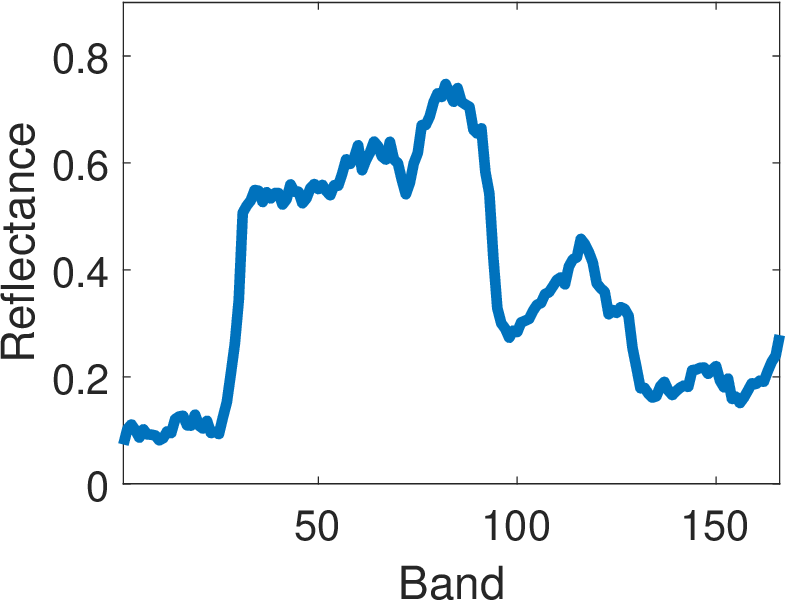}}
        \hspace{-1.1mm}
        \subfloat[]{\label{fig:eo1_SST_reflectance}\includegraphics[width=0.1420\linewidth]{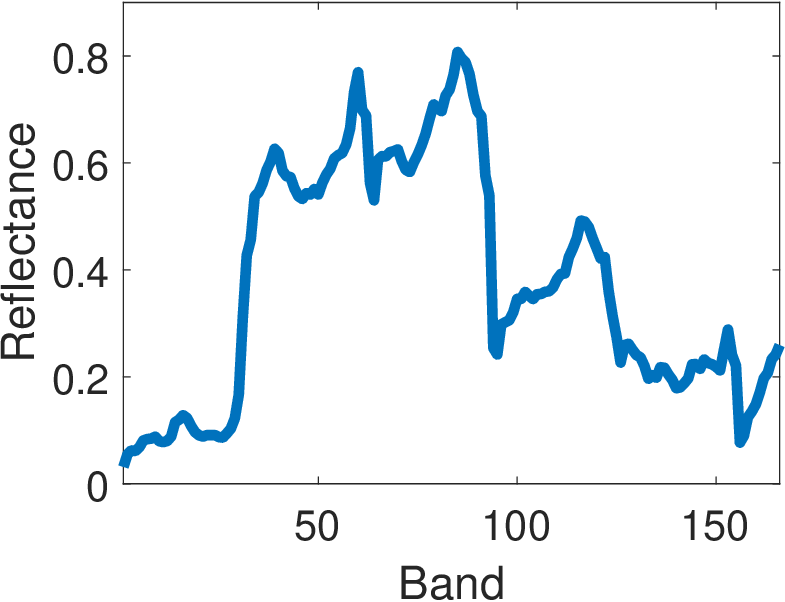}}
        \hspace{-1.1mm}
        \subfloat[]{\label{fig:eo1_DPNet_reflectance}\includegraphics[width=0.1420\linewidth]{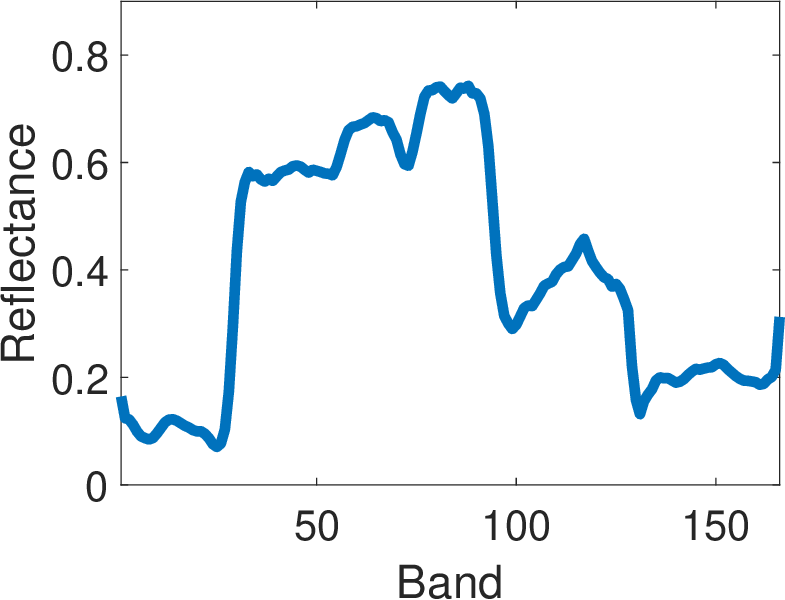}}
        \hspace{-1.1mm}
        \subfloat[]{\label{fig:eo1_ILRNet_reflectance}\includegraphics[width=0.1420\linewidth]{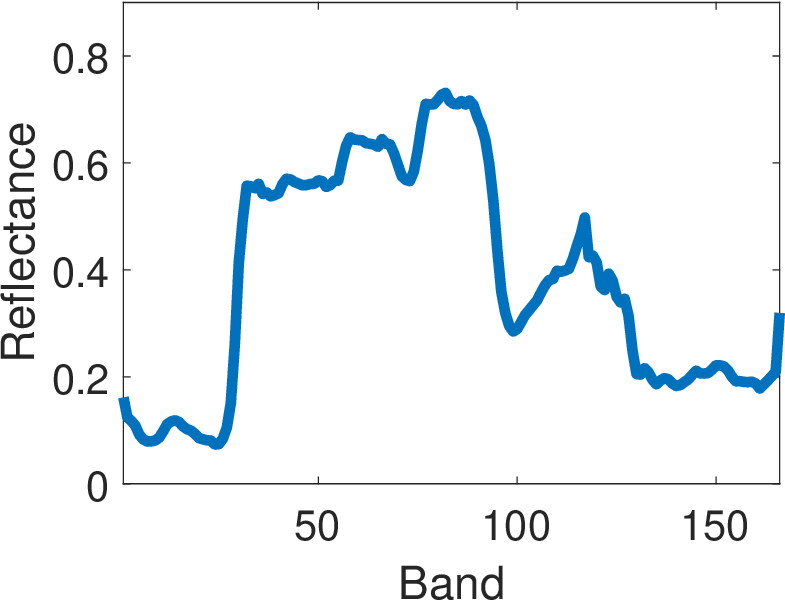}}
        \caption{Reflectance of pixel (92, 64) in the EO-1 HSI. (a) Noisy. (b) BM4D~\cite{Maggioni2013BM4D}. (c) MTSNMF~\cite{Ye2015MTSNMF}. (d) LLRT~\cite{Chang2017LLRT}. (e) NGMeet~\cite{He2022NGMeet}. (f) LRMR~\cite{Zhang2014LRMR}. (g) E-3DTV~\cite{Peng2020E-3DTV}. (h) 3DlogTNN~\cite{Zheng20203DlogTNN}. (i) T3SC~\cite{bodrito2021T3SC}. (j) MAC-Net~\cite{Xiong2022MAC-Net}. (k) TRQ3D~\cite{Pang2022TRQ3DNet}. (l) SST~\cite{li2022spatialspectral}. (m) DPNet-S~\cite{Xiongdpnet}. (n) \textbf{ILRNet}.} \label{fig:eo1_reflectance}
\end{figure*}

The first real-world noisy HSI was captured by the EO-1 satellite, with an original size of 400 $\times 1000 \times 242$ and a spectral range from 400 to 2500 nm. Following the experimental setup of LRMR~\cite{Zhang2014LRMR}, the sub-image of size $200 \times 400 \times 166$ was used for testing.

Fig.~\ref{fig:eo1_visual} provides a visual comparison of the denoising results obtained by different methods on real-world noisy HSI. As shown in Fig.~\ref{fig:eo1_visual}\subref{fig:eo1_clean_visual}, the original HSI was severely degraded by complex noise such as strip noise and deadline noise. Similar to the results observed in the synthetic noise removal experiments, most model-driven methods and MAC-Net fail to effectively remove strip noise as they are mainly designed to model Gaussian noise. Although T3SC, TRQ3D, and SST show no noticeable noise residue in their recovered results, the texture details are blurred. Due to its ability to effectively reflect the sparse characteristics of natural HSI structures, E-3DTV successfully remove strip noise, achieving a highly competitive visual quality. Thanks to  the effective combination of sparse representation and CNN, DPNet-S can effectively capture the complex structures of HSI and achieve high visual quality denoising. ILRNet, benefiting from the iterative refinement process that supplements important details while removing noise. Comparison of the enlarged portions shows that the results recovered by the most methods are relatively blurry, while ILRNet and DPNet-S can recover clearer texture details, demonstrating higher visual quality. Furthermore, an analysis of the spectral reflectance is conducted in Fig.~\ref{fig:eo1_reflectance}.  Due to the difference in the number of bands between the training and testing HSIs, T3SC, TRQ3D, and SST cannot effectively model global self-similarity, resulting in distortion of the spectral reflectance curves.  Compared to the comparative methods, ILRNet obtains  more accurate spectrum.

\subsubsection{Gaofen-5 (GF-5) CapitalAirport HSI}

\begin{figure*}[!t]
        \centering
        \subfloat[]{\label{fig:capitalairport_noise_visual}\includegraphics[width=0.1420\linewidth]{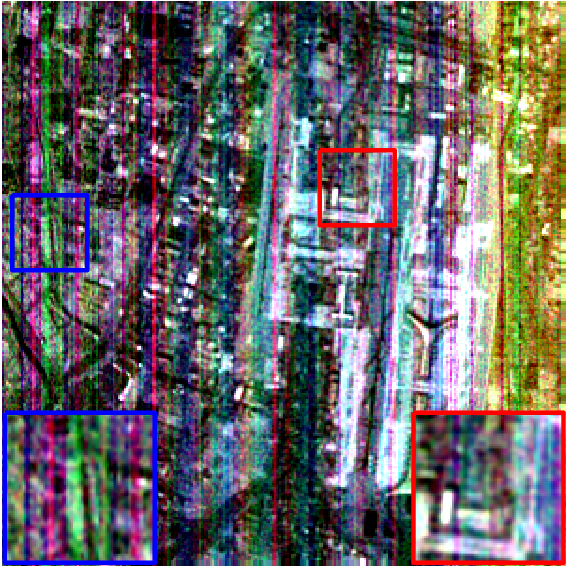}}
        \hspace{-1.1mm}
        \subfloat[]{\label{fig:capitalairport_BM4D_visual}\includegraphics[width=0.1420\linewidth]{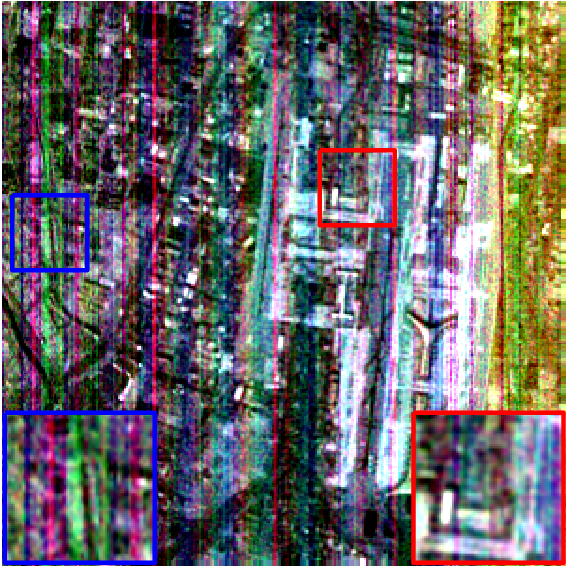}}
        \hspace{-1.1mm}
        \subfloat[]{\label{fig:capitalairport_MTSNMF_visual}\includegraphics[width=0.1420\linewidth]{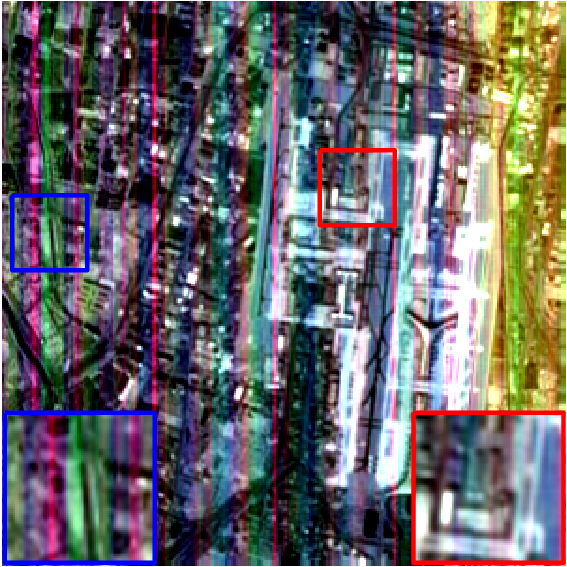}}
        \hspace{-1.1mm}
        \subfloat[]{\label{fig:capitalairport_LLRT_visual}\includegraphics[width=0.1420\linewidth]{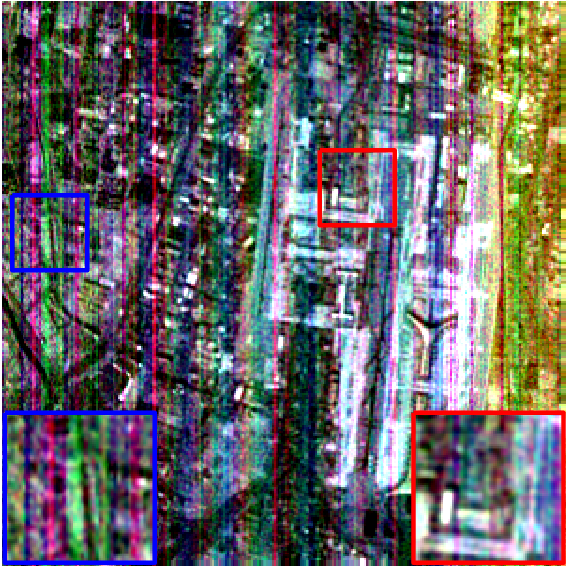}}
        \hspace{-1.1mm}
        \subfloat[]{\label{fig:capitalairport_NGMeet_visual}\includegraphics[width=0.1420\linewidth]{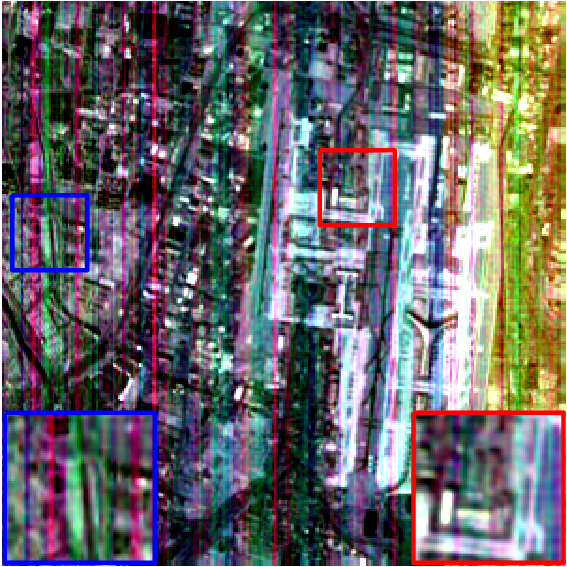}}
        \hspace{-1.1mm}
        \subfloat[]{\label{fig:capitalairport_LRMR_visual}\includegraphics[width=0.1420\linewidth]{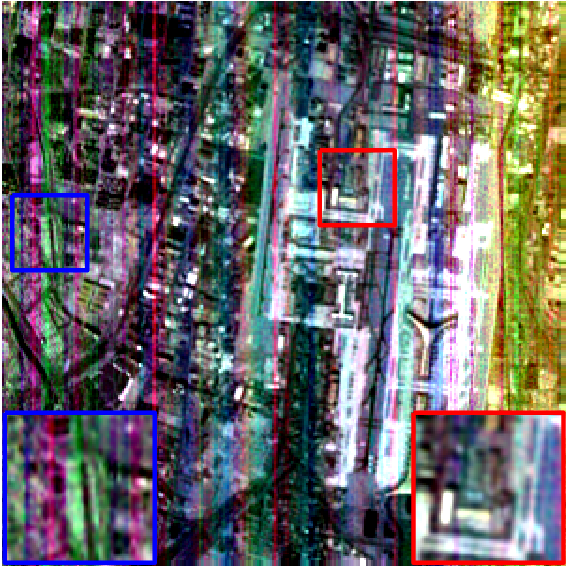}}
        \hspace{-1.1mm}
        \subfloat[]{\label{fig:capitalairport_E-3DTV_visual}\includegraphics[width=0.1420\linewidth]{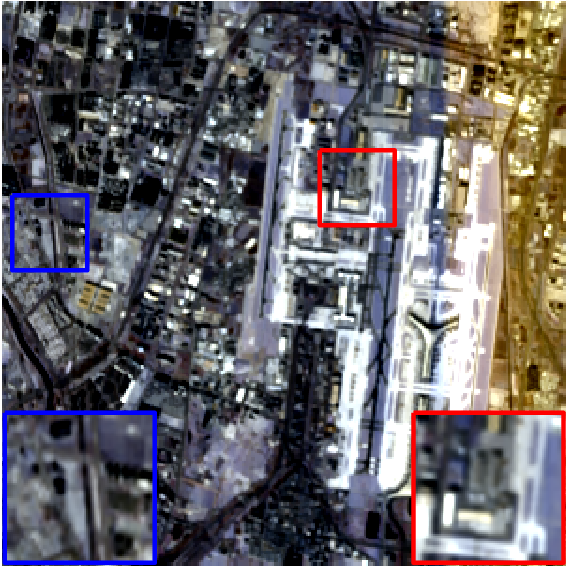}}
        \hspace{-1.1mm}
        \subfloat[]{\label{fig:capitalairport_3DlogTNN_visual}\includegraphics[width=0.1420\linewidth]{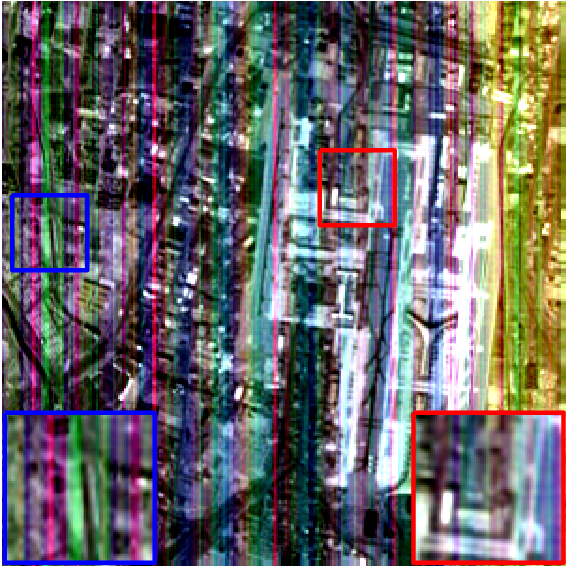}}
        \hspace{-1.1mm}
        \subfloat[]{\label{fig:capitalairport_T3SC_visual}\includegraphics[width=0.1420\linewidth]{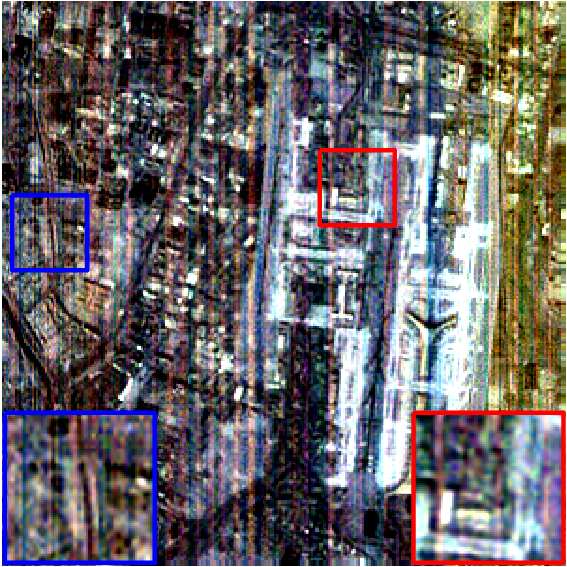}}
        \hspace{-1.1mm}
        \subfloat[]{\label{fig:capitalairport_MAC-Net_visual}\includegraphics[width=0.1420\linewidth]{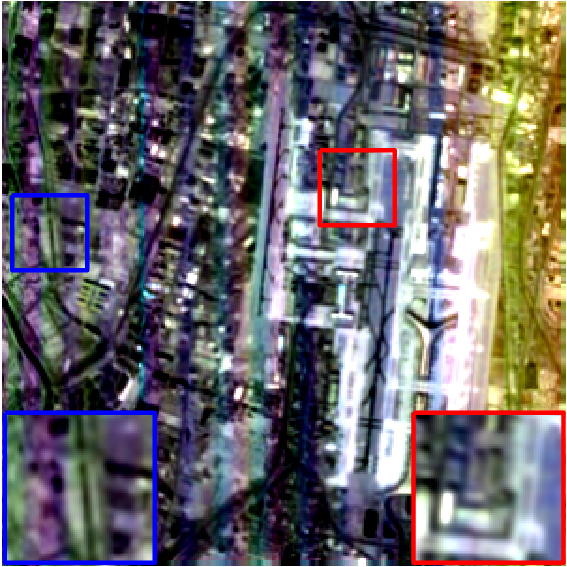}}
        \hspace{-1.1mm}
        \subfloat[]{\label{fig:capitalairport_TRQ3D_visual}\includegraphics[width=0.1420\linewidth]{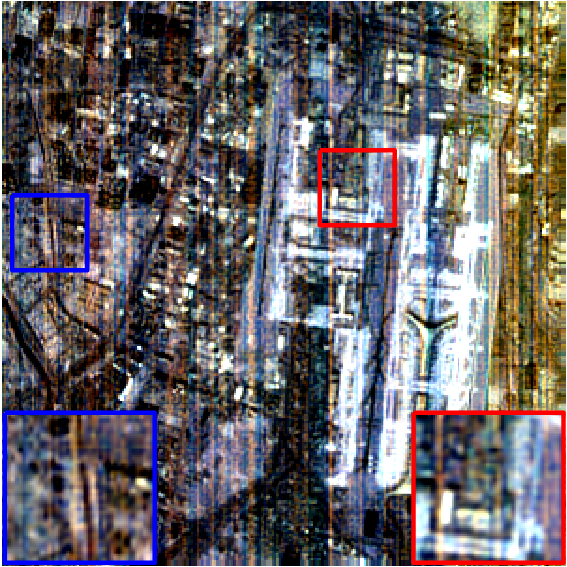}}
        \hspace{-1.1mm}
        \subfloat[]{\label{fig:capitalairport_SST_visual}\includegraphics[width=0.1420\linewidth]{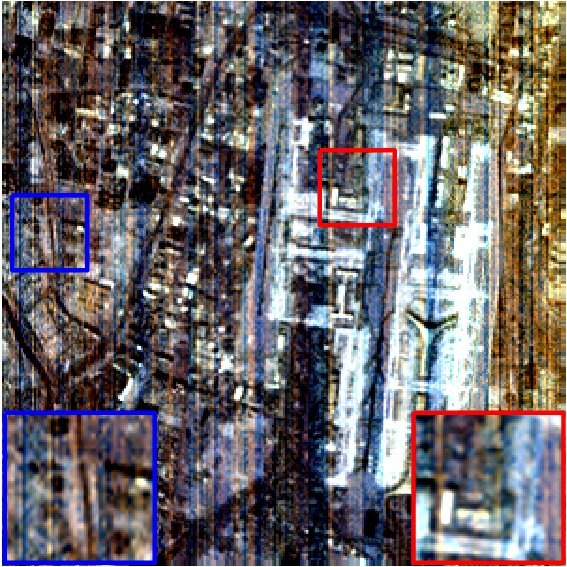}}
        \hspace{-1.1mm}
        \subfloat[]{\label{fig:capitalairport_DPNet_visual}\includegraphics[width=0.1420\linewidth]{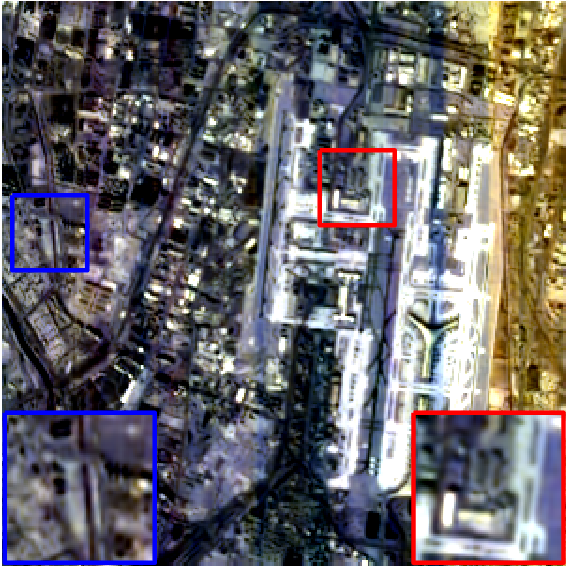}}
        \hspace{-1.1mm}
        \subfloat[]{\label{fig:capitalairport_ILRNet_visual}\includegraphics[width=0.1420\linewidth]{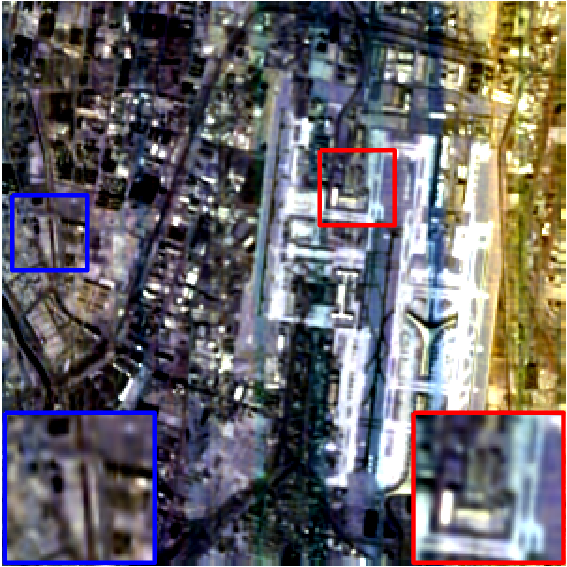}}
        \caption{Denoising results of CapitalAirport HSI collected from the GF-5 satellite. The false-color images are generated by combining bands 154, 152, and 139. (a) Noisy. (b) BM4D~\cite{Maggioni2013BM4D}. (c) MTSNMF~\cite{Ye2015MTSNMF}. (d) LLRT~\cite{Chang2017LLRT}. (e) NGMeet~\cite{He2022NGMeet}. (f) LRMR~\cite{Zhang2014LRMR}. (g) E-3DTV~\cite{Peng2020E-3DTV}. (h) 3DlogTNN~\cite{Zheng20203DlogTNN}. (i) T3SC~\cite{bodrito2021T3SC}. (j) MAC-Net~\cite{Xiong2022MAC-Net}. (k) TRQ3D~\cite{Pang2022TRQ3DNet}. (l) SST~\cite{li2022spatialspectral}. (m) DPNet-S~\cite{Xiongdpnet}. (n) \textbf{ILRNet}.} \label{fig:capitalairport_visual}
\end{figure*}

\begin{figure*}[!t]
        \centering
        \subfloat[]{\label{fig:capitalairport_noise_reflectance}\includegraphics[width=0.1420\linewidth]{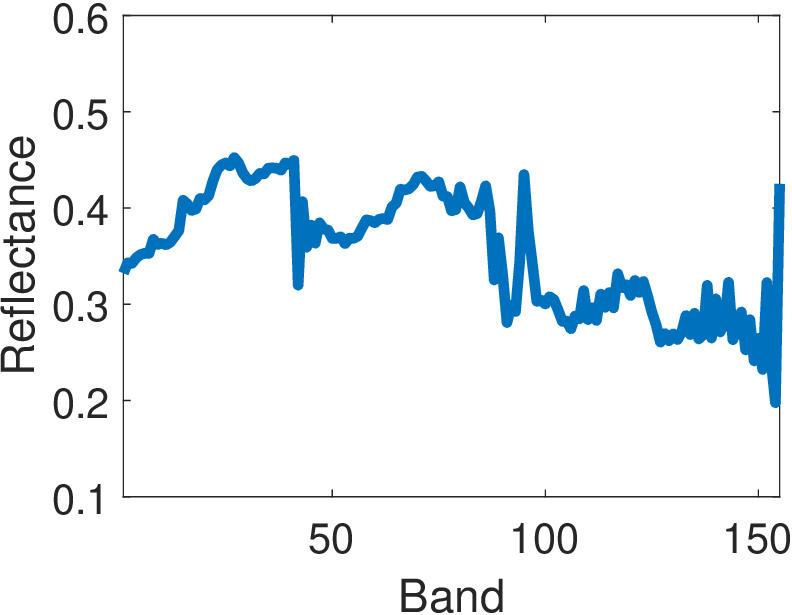}}
        \hspace{-1.1mm}
        \subfloat[]{\label{fig:capitalairport_BM4D_reflectance}\includegraphics[width=0.1420\linewidth]{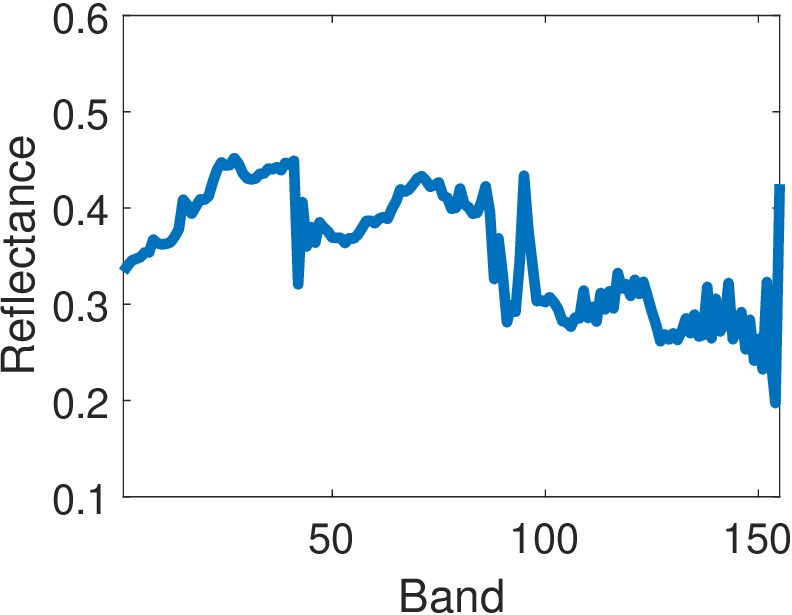}}
        \hspace{-1.1mm}
        \subfloat[]{\label{fig:capitalairport_MTSNMF_reflectance}\includegraphics[width=0.1420\linewidth]{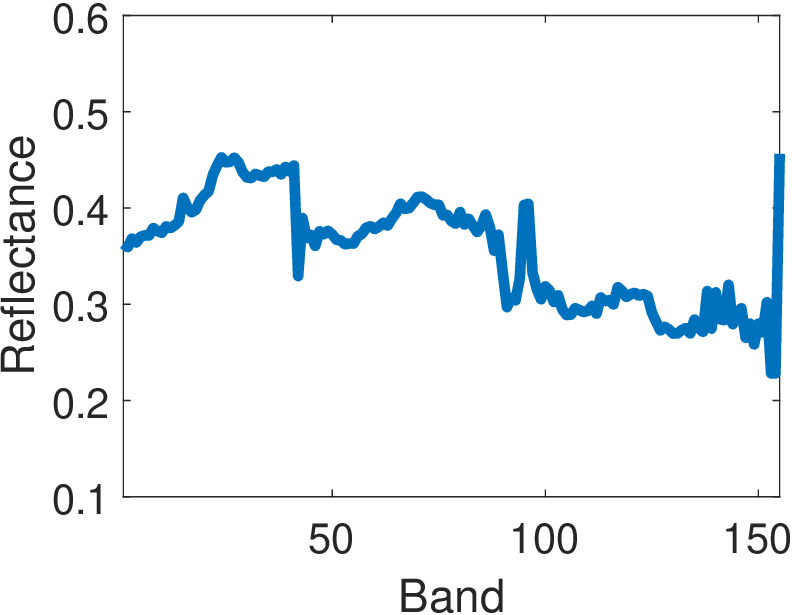}}
        \hspace{-1.1mm}
        \subfloat[]{\label{fig:capitalairport_LLRT_reflectance}\includegraphics[width=0.1420\linewidth]{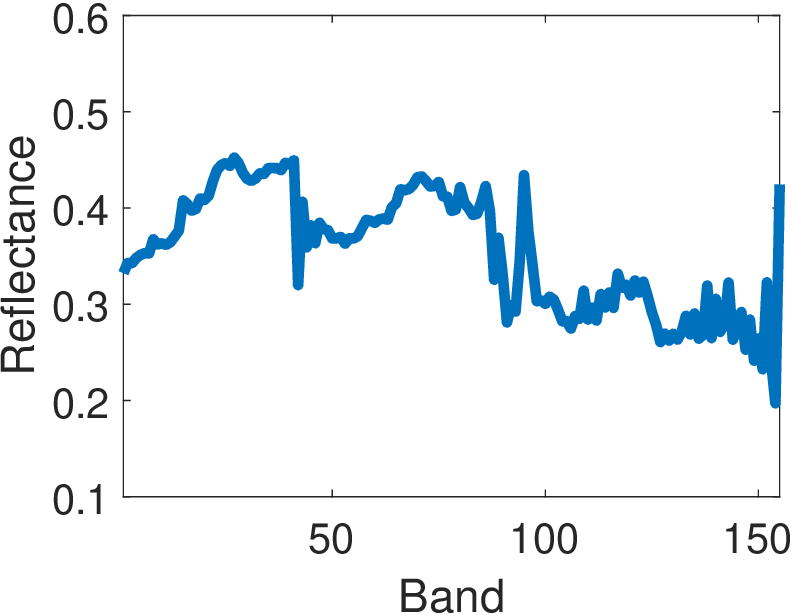}}
        \hspace{-1.1mm}
        \subfloat[]{\label{fig:capitalairport_NGMeet_reflectance}\includegraphics[width=0.1420\linewidth]{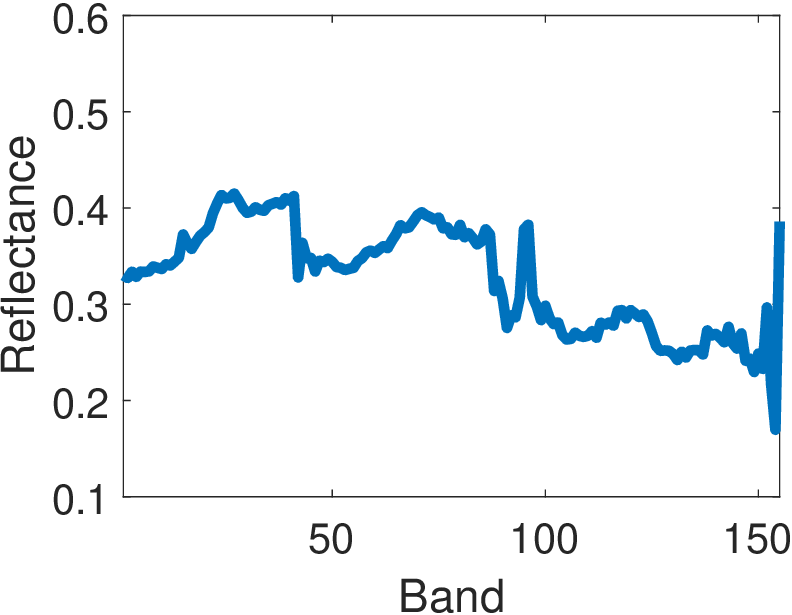}}
        \hspace{-1.1mm}
        \subfloat[]{\label{fig:capitalairport_LRMR_reflectance}\includegraphics[width=0.1420\linewidth]{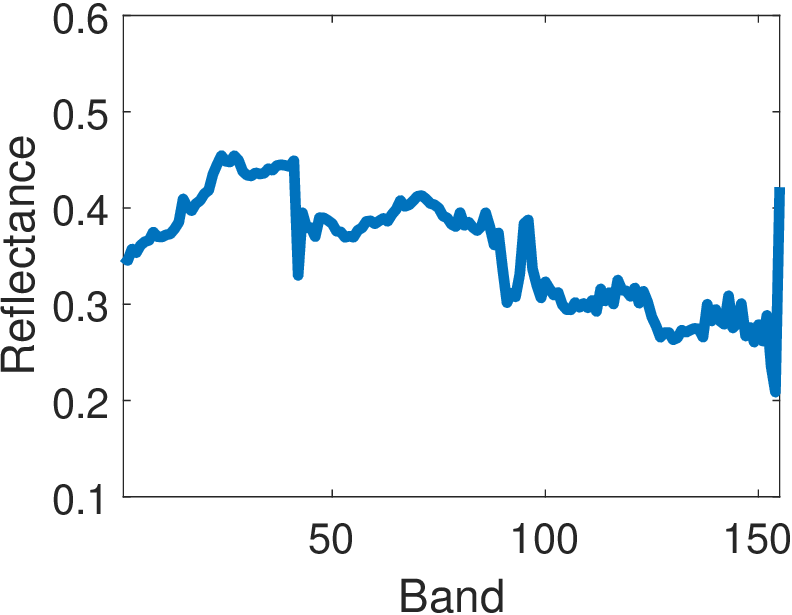}}
        \hspace{-1.1mm}
        \subfloat[]{\label{fig:capitalairport_E-3DTV_reflectance}\includegraphics[width=0.1420\linewidth]{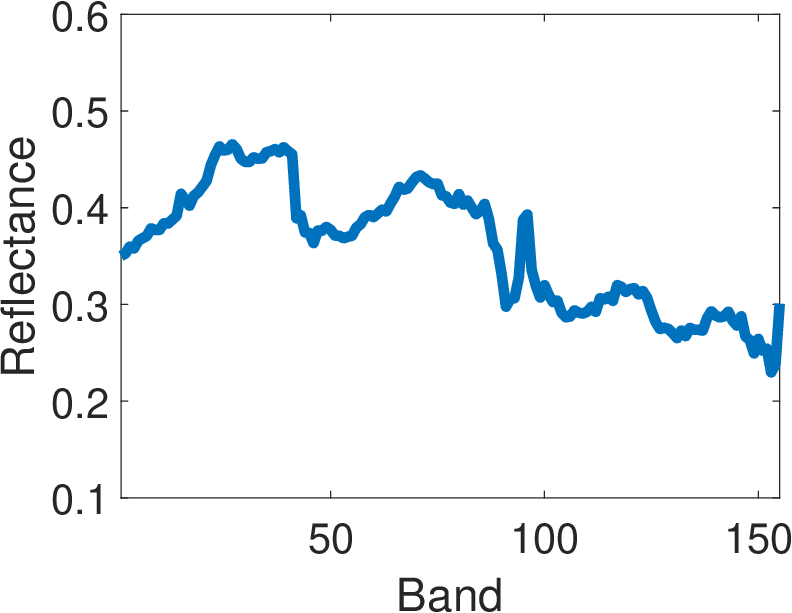}}
        \hspace{-1.1mm}
        \subfloat[]{\label{fig:capitalairport_3DlogTNN_reflectance}\includegraphics[width=0.1420\linewidth]{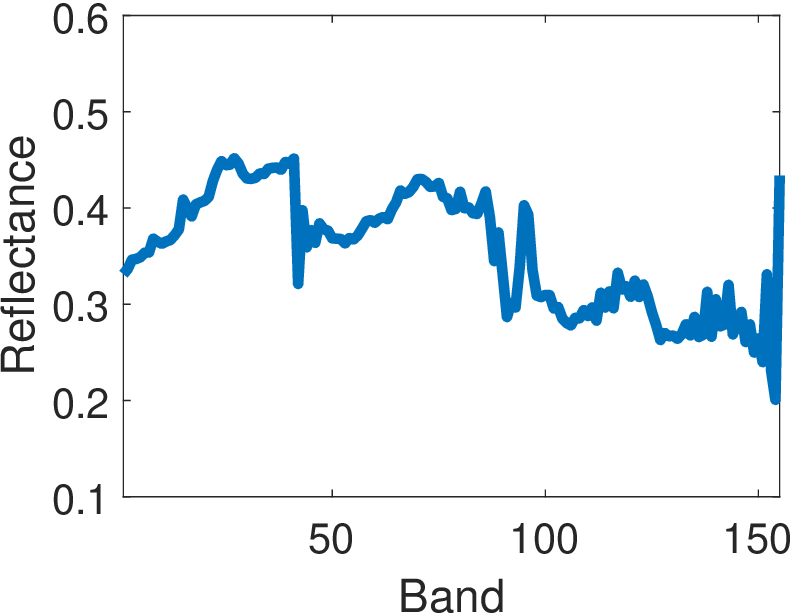}}
        \hspace{-1.1mm}
        \subfloat[]{\label{fig:capitalairport_T3SC_reflectance}\includegraphics[width=0.1420\linewidth]{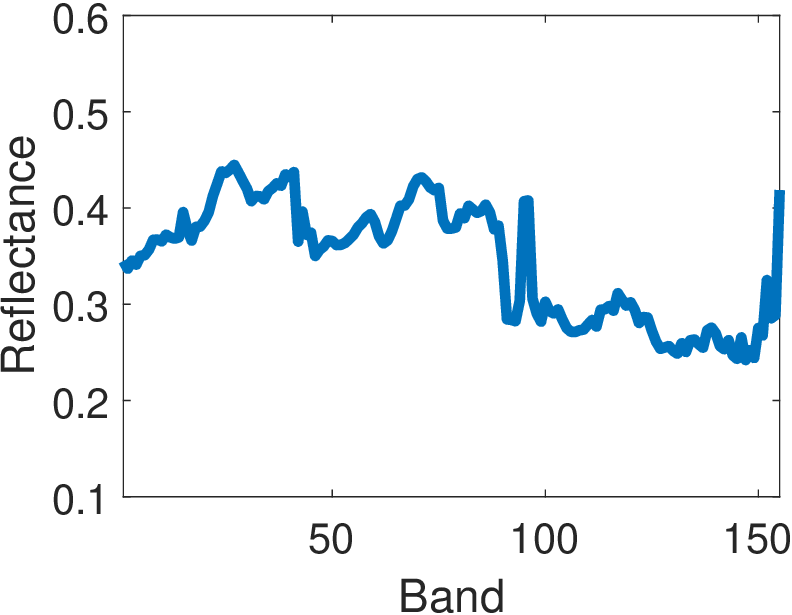}}
        \hspace{-1.1mm}
        \subfloat[]{\label{fig:capitalairport_MAC-Net_reflectance}\includegraphics[width=0.1420\linewidth]{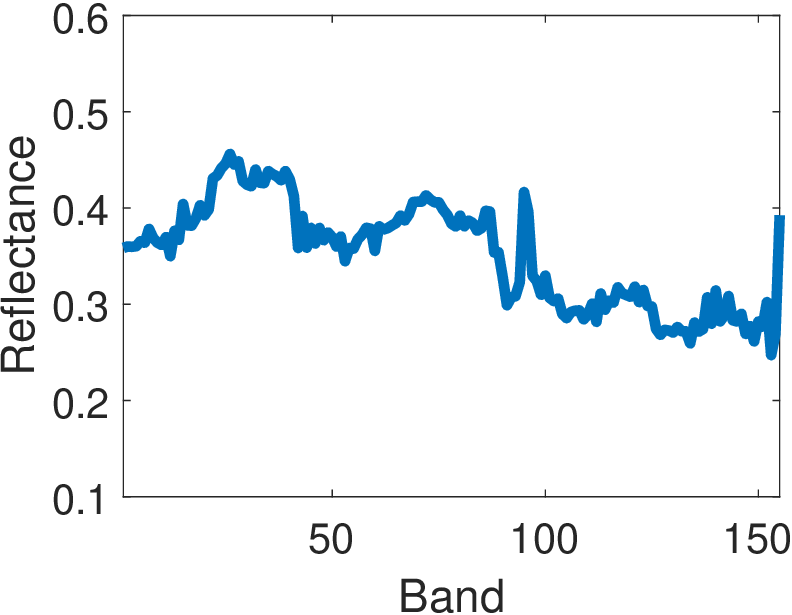}}
        \hspace{-1.1mm}
        \subfloat[]{\label{fig:capitalairport_TRQ3D_reflectance}\includegraphics[width=0.1420\linewidth]{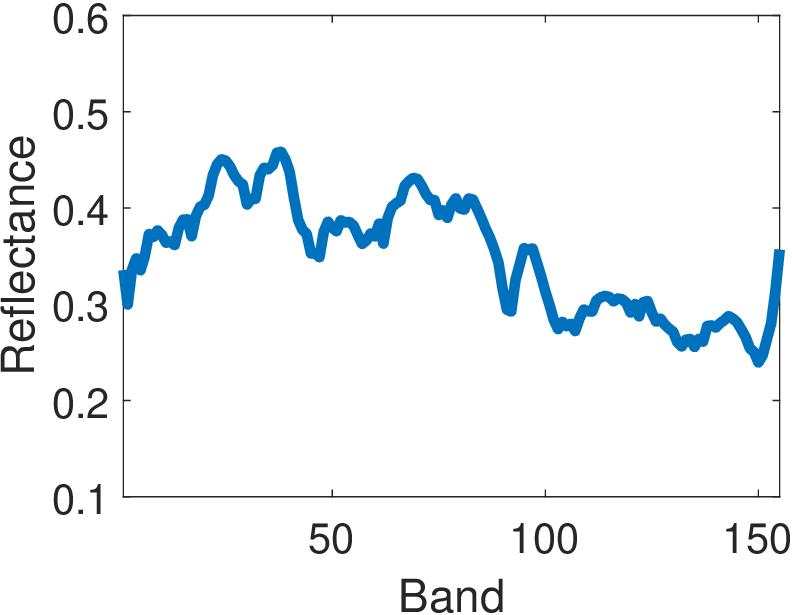}}
        \hspace{-1.1mm}
        \subfloat[]{\label{fig:capitalairport_SST_reflectance}\includegraphics[width=0.1420\linewidth]{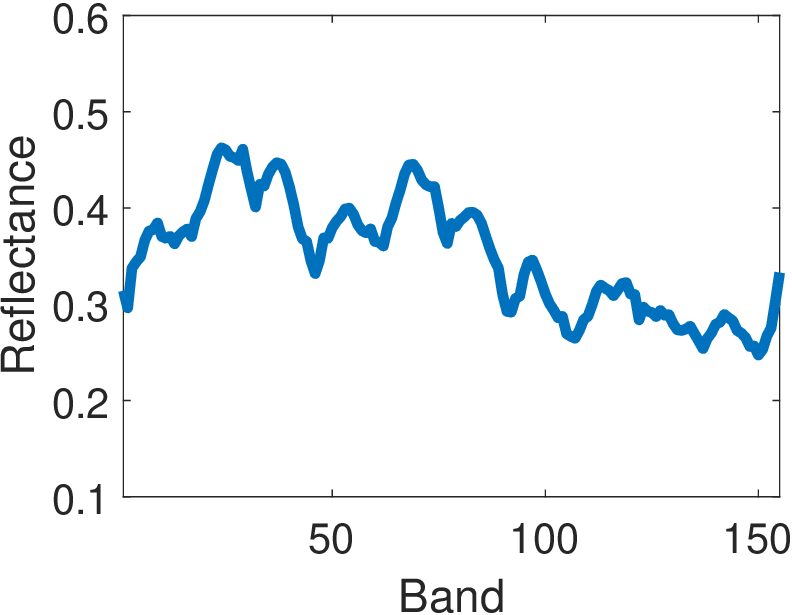}}
        \hspace{-1.1mm}
        \subfloat[]{\label{fig:capitalairport_DPNet_reflectance}\includegraphics[width=0.1420\linewidth]{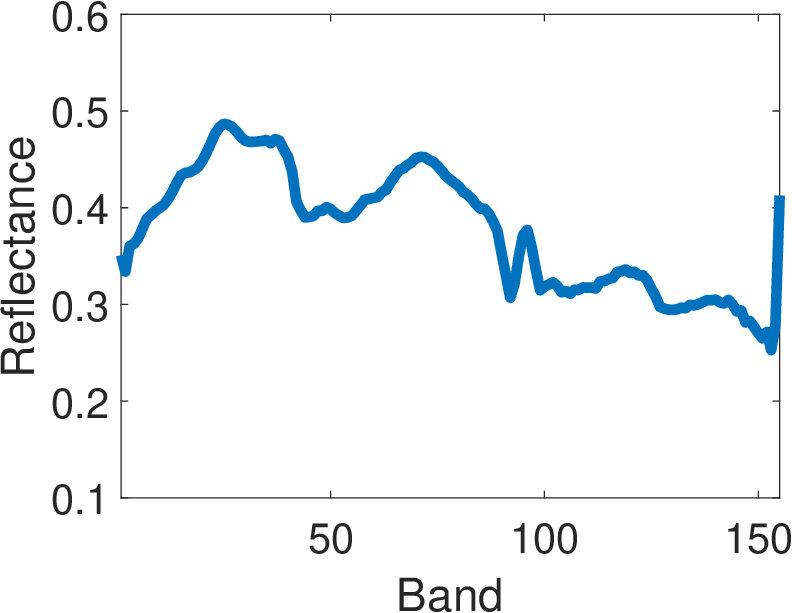}}
        \hspace{-1.1mm}
        \subfloat[]{\label{fig:capitalairport_ILRNet_reflectance}\includegraphics[width=0.1420\linewidth]{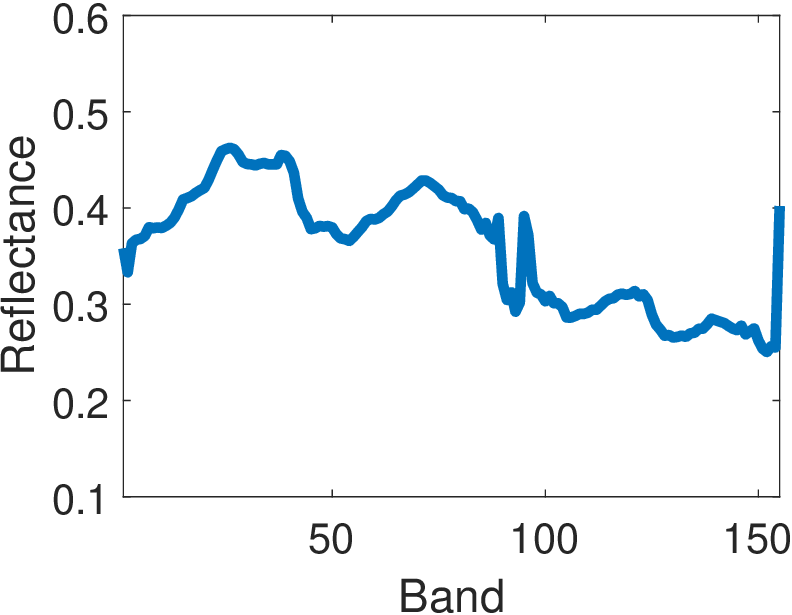}}
        \caption{Reflectance of pixel (90, 84) in the CapitalAirport HSI collected from the GF-5 satellite. (a) Noisy. (b) BM4D~\cite{Maggioni2013BM4D}. (c) MTSNMF~\cite{Ye2015MTSNMF}. (d) LLRT~\cite{Chang2017LLRT}. (e) NGMeet~\cite{He2022NGMeet}. (f) LRMR~\cite{Zhang2014LRMR}. (g) E-3DTV~\cite{Peng2020E-3DTV}. (h) 3DlogTNN~\cite{Zheng20203DlogTNN}. (i) T3SC~\cite{bodrito2021T3SC}. (j) MAC-Net~\cite{Xiong2022MAC-Net}. (k) TRQ3D~\cite{Pang2022TRQ3DNet}. (l) SST~\cite{li2022spatialspectral}. (m) DPNet-S~\cite{Xiongdpnet}. (n) \textbf{ILRNet}.} \label{fig:capitalairport_reflectance}
\end{figure*}

The second real-world noisy HSI was captured by the GF-5 satellite, including 330 spectral channels with a spectral range from 400 to 2500 nm. In this paper, we crop the sub-image of size $300 \times 300 \times 155$ for testing.

Fig.~\ref{fig:capitalairport_visual} provides a visual comparison of the denoising results obtained by different methods on the CapitalAirport HSI. Fig.~\ref{fig:capitalairport_visual}\subref{fig:capitalairport_noise_visual} was severely degraded by a larger range of strip noise, presenting a greater challenge for noise removal. Similarly, the denoising results of model-driven methods show noticeable residue of strip noise, with E-3DTV exhibiting less noise residue and better visual quality. Unlike the recovered results obtained from the EO-1 HSI, data-driven methods such as T3SC, MAC-Net, TRQ3D, and SST fail to effectively remove the extensive strip noise. Compared to the comparative methods, DPNet-S and ILRNet exhibit less noise residue and higher visual quality. Furthermore, according to the spectral reflectance curves plotted in Fig.~\ref{fig:capitalairport_reflectance}, ILRNet and DPNet-S produce more continuous curves. In summary, the real-world noise removal experiment demonstrates the strong generalization capability and considerable restoration quality of ILRNet on real-world data.

\subsection{Ablation Study}

In this section, we investigated the impact of iteration count, the RMM and Wavelet Transform on the ILRNet. All the experiments were performed on the ICVL dataset.

\subsubsection{Impact of the Number of Iterations}

\begin{figure*}[ht]
        \centering
        \subfloat{\includegraphics[width=\linewidth]{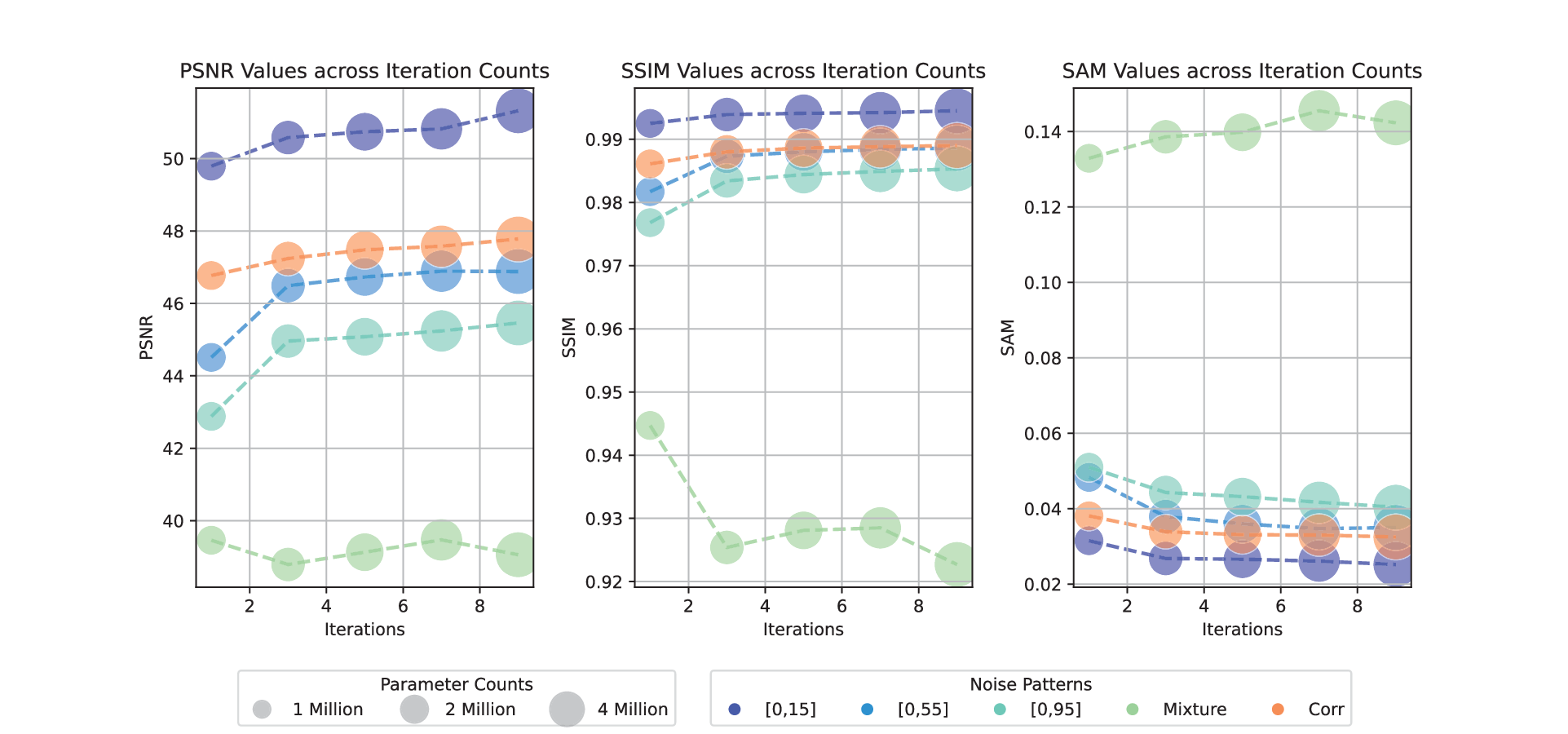}}
        \caption{ The impact of iteration count on the performance and parameter count of ILRNet. } \label{fig:iteration_ablution}
\end{figure*}

\begin{figure*}[!t]
        \centering
        \subfloat[Clean]{\label{fig:negev_0823-1005_clean}\includegraphics[width=0.16\linewidth]{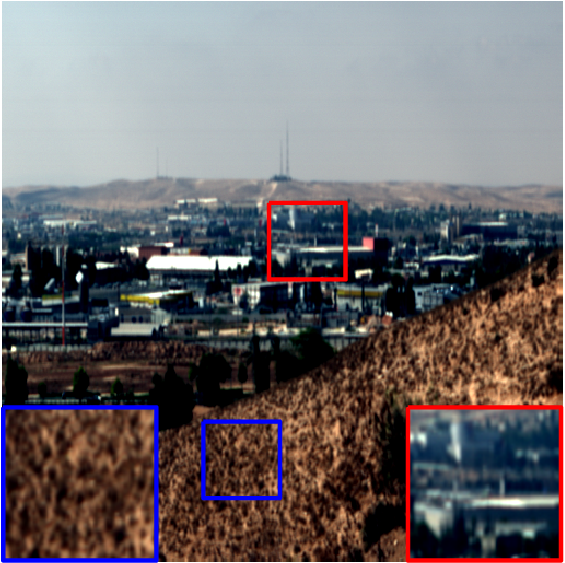}}
        \hspace{-1.1mm}
        \subfloat[Noisy]{\label{fig:negev_0823-1005_noise}\includegraphics[width=0.16\linewidth]{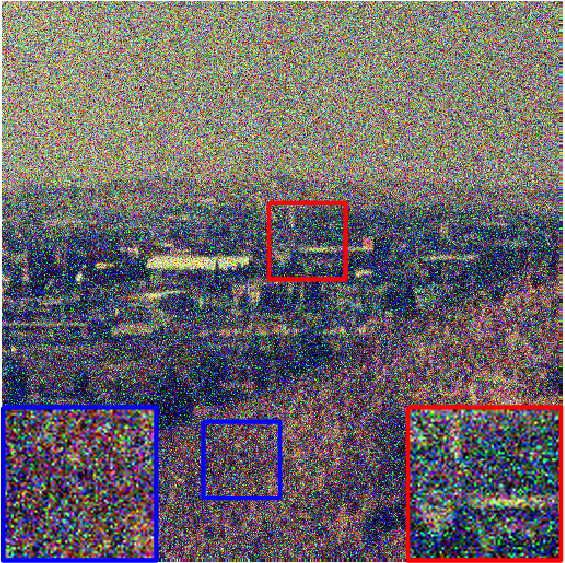}}
        \hspace{-1.1mm}
        \subfloat[$\X^{(0)}$]{\label{fig:negev_0823-1005_coarse}\includegraphics[width=0.16\linewidth]{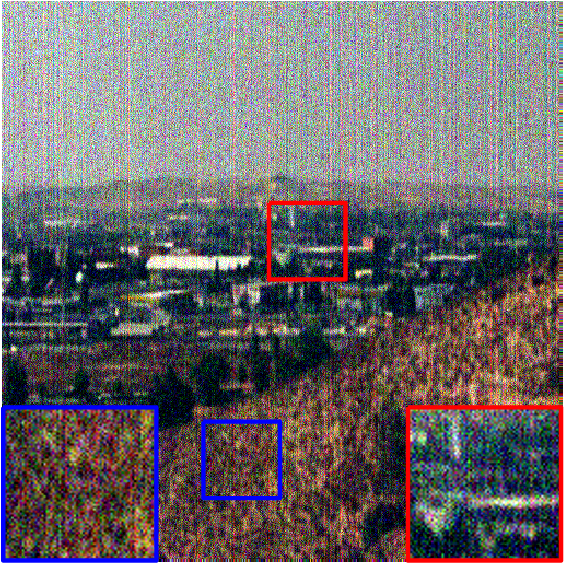}}
        \hspace{-1.1mm}
        \subfloat[$\X^{(1)}$]{\label{fig:negev_0823-1005_ref1}\includegraphics[width=0.16\linewidth]{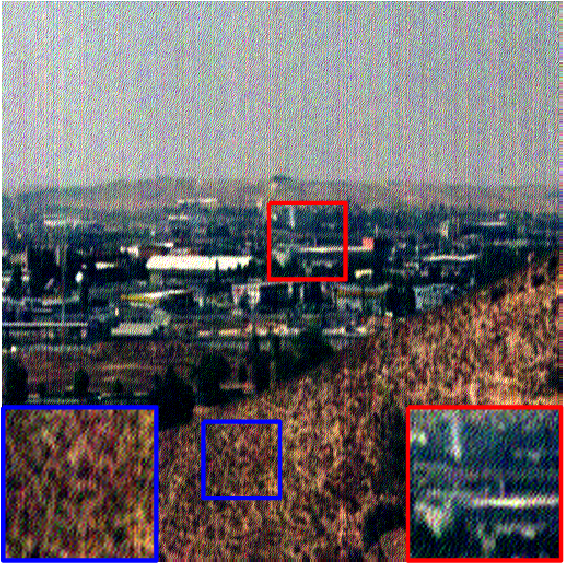}}
        \hspace{-1.1mm}
        \subfloat[$\X^{(2)}$]{\label{fig:negev_0823-1005_ref2}\includegraphics[width=0.16\linewidth]{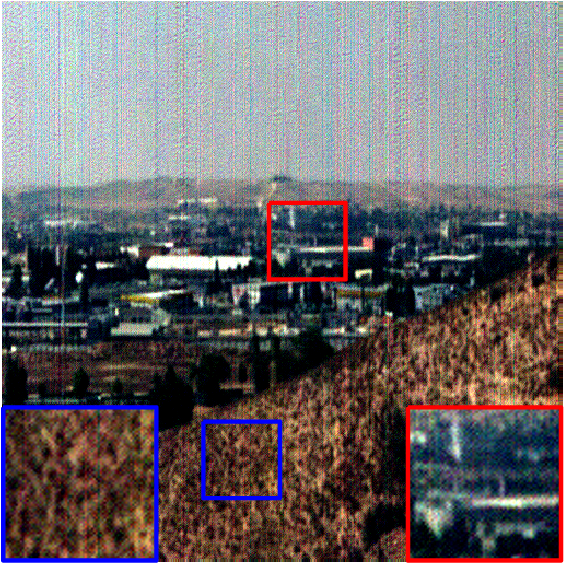}}
        \hspace{-1.1mm}
        \subfloat[$\X^{(3)}$]{\label{fig:negev_0823-1005_ref3}\includegraphics[width=0.16\linewidth]{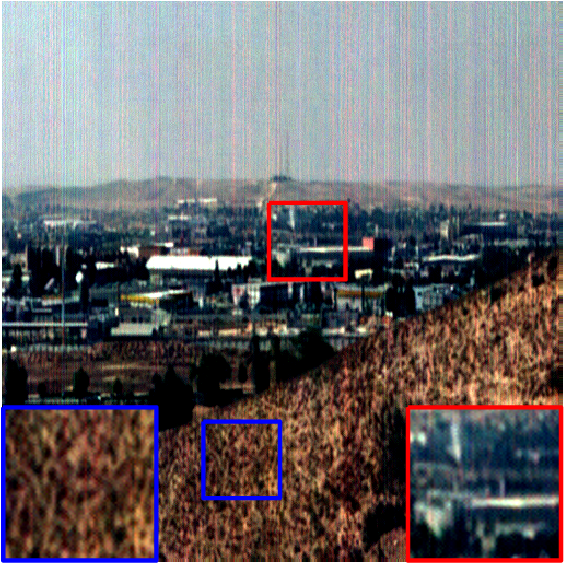}}
        \hspace{-1.1mm}
        \subfloat[$\X^{(4)}$]{\label{fig:negev_0823-1005_ref4}\includegraphics[width=0.16\linewidth]{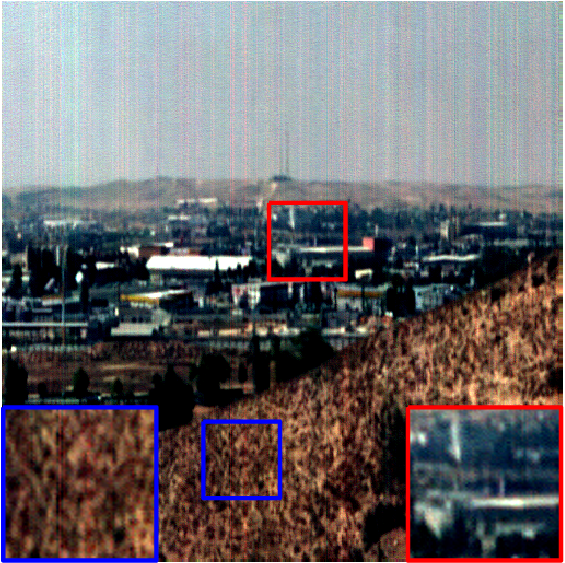}}
        \hspace{-1.1mm}
        \subfloat[$\X^{(5)}$]{\label{fig:negev_0823-1005_ref5}\includegraphics[width=0.16\linewidth]{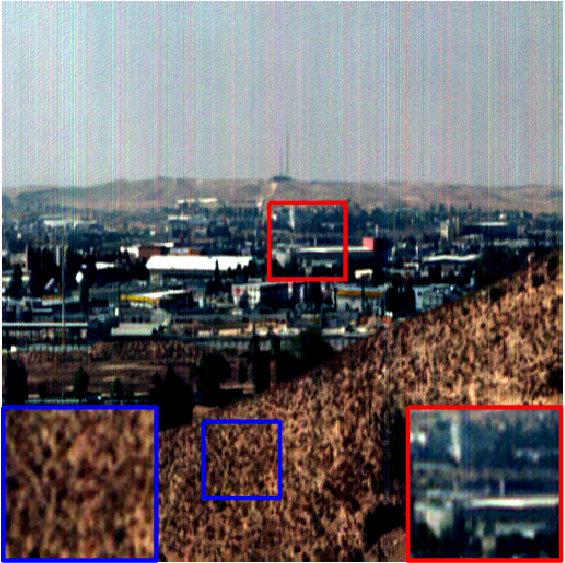}}
        \hspace{-1.1mm}
        \subfloat[$\X^{(6)}$]{\label{fig:negev_0823-1005_ref6}\includegraphics[width=0.16\linewidth]{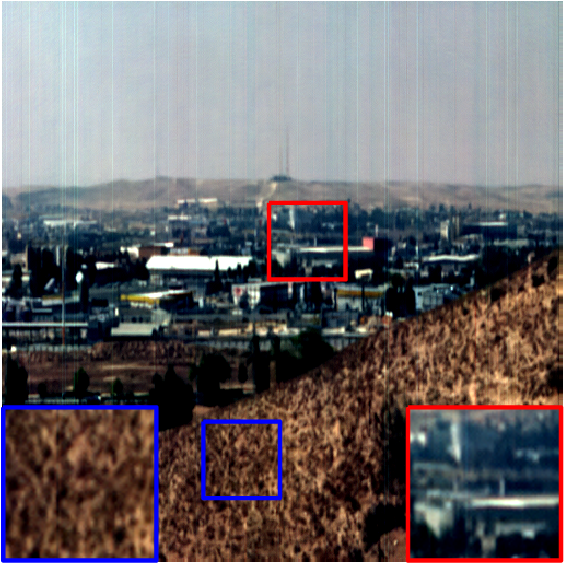}}
        \hspace{-1.1mm}
        \subfloat[$\X^{(7)}$]{\label{fig:negev_0823-1005_ref7}\includegraphics[width=0.16\linewidth]{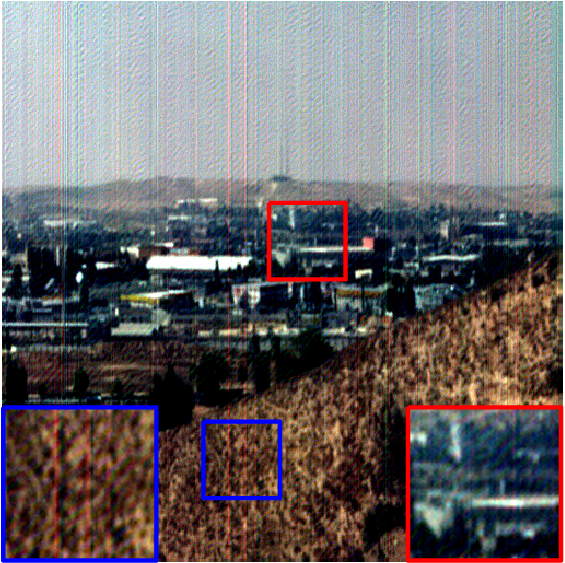}}
        \hspace{-1.1mm}
        \subfloat[$\X^{(8)}$]{\label{fig:negev_0823-1005_ref6}\includegraphics[width=0.16\linewidth]{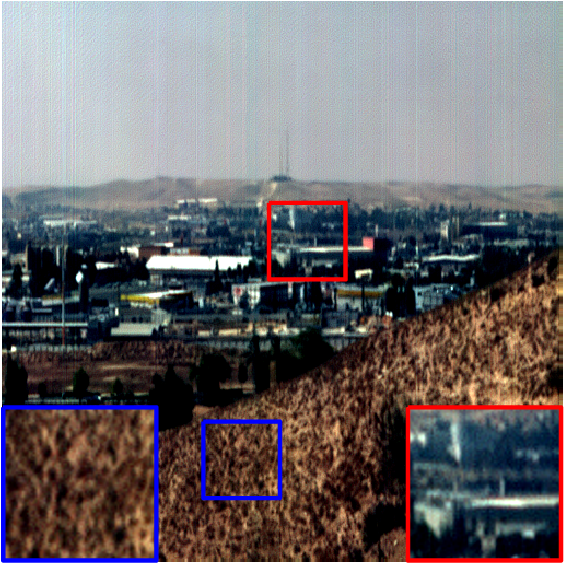}}
        \hspace{-1.1mm}
        \subfloat[$\X^{(9)}$]{\label{fig:negev_0823-1005_ref7}\includegraphics[width=0.16\linewidth]{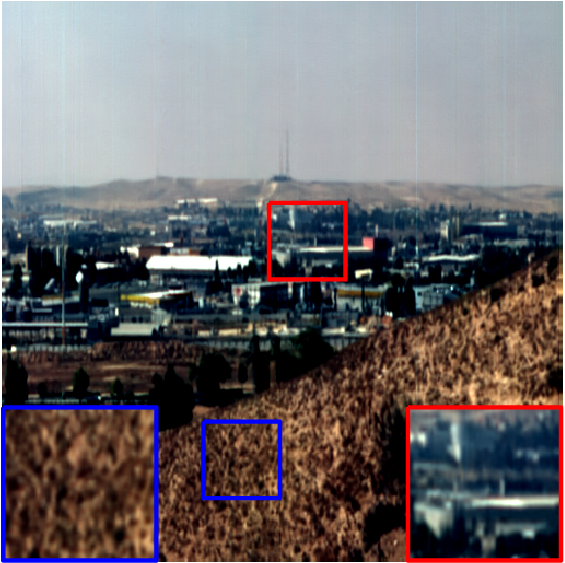}}

           \caption{A demonstration of the iterative refinement process on the negev\_0823-1005 HSI from the ICVL dataset under the mixture noise. The false-color images are generated by combining bands 23, 15, and 9.} \label{fig:refinement_visual}
\end{figure*}

\begin{figure}[!t]
        \centering
        \subfloat[$\lambda_1$]{\label{fig:analysis_lambda1}\includegraphics[width=0.48\linewidth]{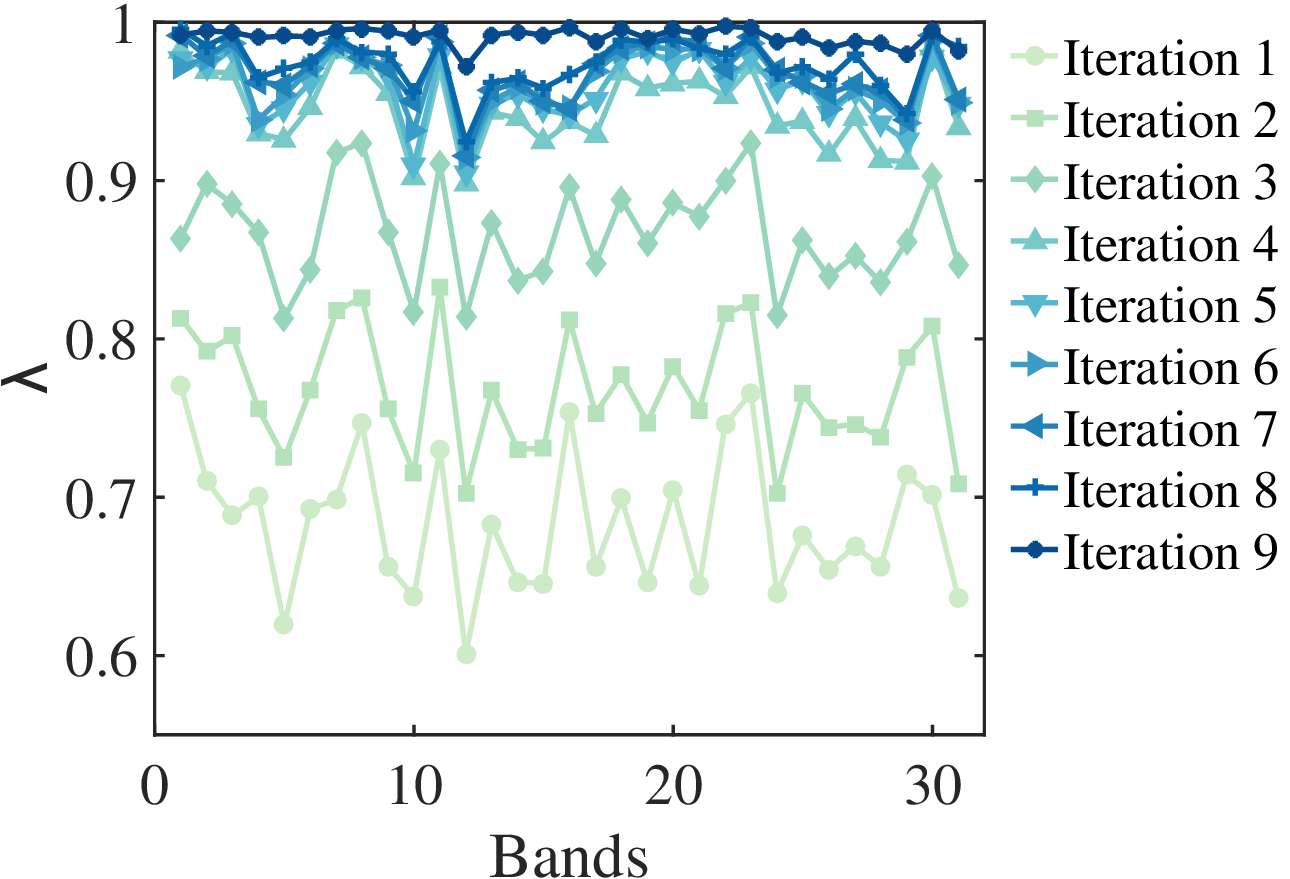}}
        \hspace{-1.1mm}
        \subfloat[$\lambda_2$]{\label{fig:analysis_lambda2}\includegraphics[width=0.48\linewidth]{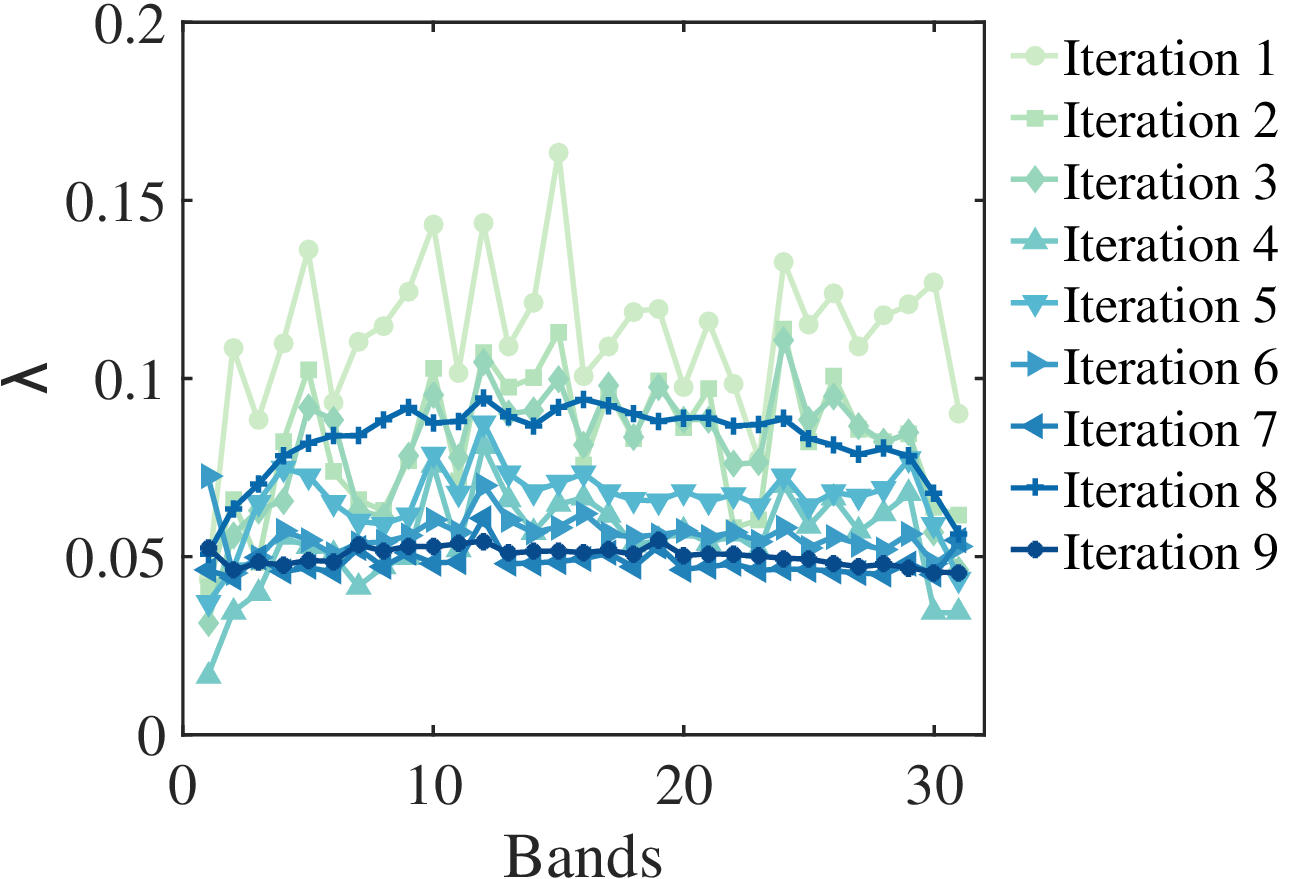}}
        \caption{ The average value of $\lambda1$ and $\lambda2$  under the mixture noise during the iterative refinement process.} \label{fig:analysis_lambda}
\end{figure}


To explore the impact of iteration count on the performance of ILRNet, we compared the indexes and parameter counts of ILRNet under different iteration counts in Fig.~\ref{fig:iteration_ablution}. According to the presented results, as the iteration count increases, the model's parameter count increases, its representation capacity improves gradually,  gradually improving the performance. Under most noise patterns, the performance improvement shows a gradual trend towards stabilization as the number of iterations increases. Considering both model performance and parameter count, we set the final iteration count to 9.

Furthermore, we visualize the intermediate results of the iteration process in Fig.~\ref{fig:refinement_visual}. Overall, as the refinement iterations progress, the visual quality of the recovered results gradually improves, further demonstrating the effectiveness of the iterative refinement process.

Finally, we conducted an analysis of the outputs of the $\Lambda(\cdot)$ module. Fig.~\ref{fig:analysis_lambda} illustrates the changes in $\lambda_1$ and $\lambda_2$ during the iterations. Fig.~\ref{fig:analysis_lambda}\subref{fig:analysis_lambda1} illustrates the variation of $\lambda_1$. According to Fig.~\ref{fig:analysis_lambda}\subref{fig:analysis_lambda1}, with the increase of iteration count, the value of $\lambda_1$ generally shows an increasing trend and gradually approaches 1. This indicates that as the refinement progresses, the quality of the HSI gradually improves, reducing the need to introduce details from the noisy observations $\Y$. This phenomenon is consistent with the original motivation behind designing ILRNet. Fig.~\ref{fig:analysis_lambda}\subref{fig:analysis_lambda2} presents the variation of $\lambda_2$. On one hand, the value of $\lambda_2$ remains consistently small, indicating that refining $\Z^{(k)}$ is a straightforward task. Even with the use of lightweight $f(\cdot)$, higher-quality HSI can be obtained, thus there is a relatively small demand for combining the results of the previous iteration. On the other hand, the value of $\lambda_2$ generally decreases with the increase in the number of iterations. This is because $\lambda_1$ gradually increases, leading to less noise introduced when combining with $\Y$, making the refinement of $\Z^{(k)}$ easier. Additionally, there are significant differences in the values of $\lambda_1$ and $\lambda_2$ for different bands, which further confirms our hypothesis that the sensor has different sensitivities in different bands, resulting in differences in $\lambda$ values between bands.

\subsubsection{Effectiveness of the RMM}
\begin{table}[!t]
        \caption{Denoising performance with/without the RMM Module.}\label{tab:lr_ablation}
        \centering
        \resizebox{\linewidth}{!}{
                \tablesize{
        \begin{tabular}{c|c|c|c|c|c|c}
                \Xhline{1.2pt}
        \multirow{2}*{RMM}&\multirow{2}*{Index}
        &\multicolumn{5}{c}{$\sigma$}\\
        \cline{3-7}
        &&\multirow{1}*{\makebox[0.06\textwidth][c]{\textbf{[0,15]}}}&\multirow{1}*{\makebox[0.06\textwidth][c]{\textbf{[0,55]}}}&\multirow{1}*{\makebox[0.06\textwidth][c]{\textbf{[0,95]}}}&\multirow{1}*{\makebox[0.06\textwidth][c]{Mixture}}&\multirow{1}*{\makebox[0.06\textwidth][c]{Corr}}\\
        \Xhline{1.2pt}  		
        \multirow{3}*{$\checkmark$}
        & PSNR$\uparrow$  & 51.22 & 46.88 & 45.46 & 39.06 & 47.80 \\
        & SSIM$\uparrow$  & .9945 & .9886 & .9853 & .9227 & .9890 \\
        & SAM$\downarrow$ & .0255 & .0350 & .0404 & .1423 & .0323 \\
        \Xhline{1.2pt}	
        \multirow{3}*{$\times$}
        & PSNR$\uparrow$  & 50.85 & 46.09 & 45.40 & 38.68 & 47.46 \\
        & SSIM$\uparrow$  & .9943 & .9875 & .9854 & .9184 & .9887 \\
        & SAM$\downarrow$ & .0258 & .0367 & .0408 & .1488 & .0330 \\
        \Xhline{1.2pt}
        \end{tabular}}}
\end{table}
\begin{figure}[!t]
        \centering
        \includegraphics[width=1\linewidth]{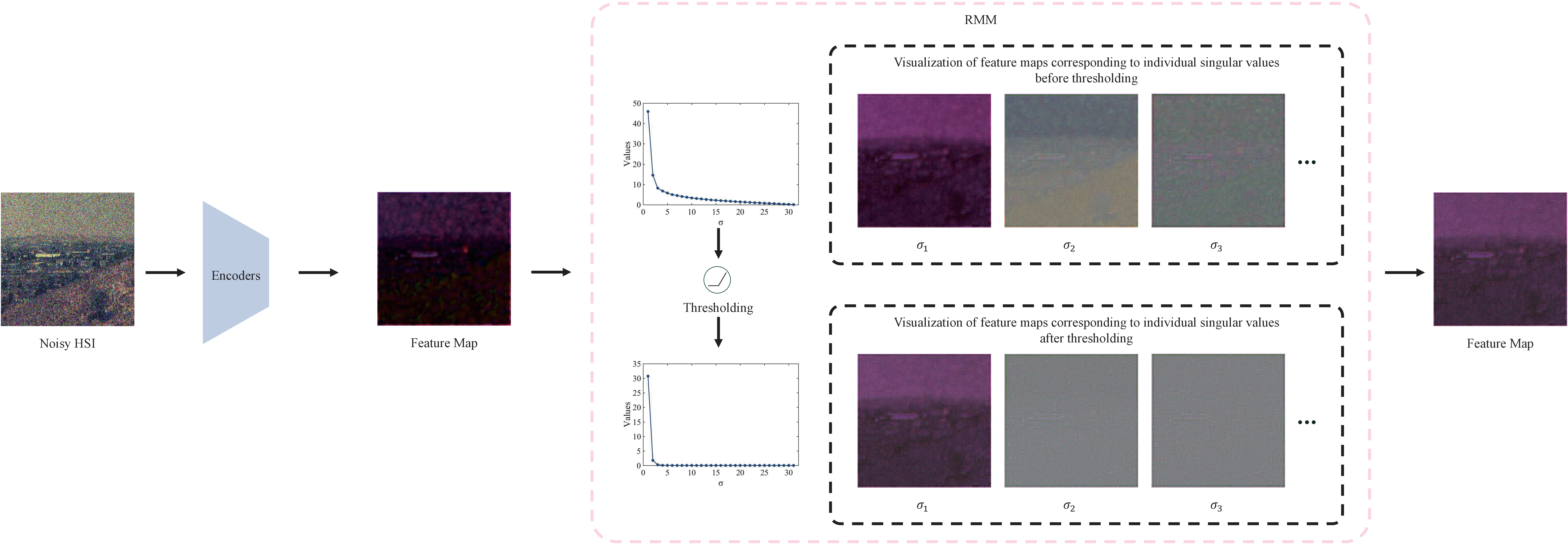}
        \caption{Visualization of the effectiveness of the RMM module.}\label{fig:svd_visual}
\end{figure}


To confirm the effectiveness of the RMM, we present the performance of ILRNet with/without the RMM in the coarse estimation module in Table~\ref{tab:lr_ablation}. According to the results shown, the inclusion of the RMM improves the performance of ILRNet, validating its effectiveness.

To provide a more profound illustration of the inherent low-rank characteristics of HSI, we present the singular values $\sigma$ obtained by RMM in Fig.~\ref{fig:svd_visual}. Using the RMM, the singular values present the obvious decay with more values approaching zero, indicating that our RMM can achieves the low-rank representation of the HSI. Additionally, the feature map processed by RMM exhibits  less noise, confirming the effectiveness of RMM in denoising.

\subsubsection{Effectiveness of the Wavelet Transform}

\begin{figure}[!t]
        \centering
	\subfloat[]{\label{fig:clean}\includegraphics[width=0.23\linewidth]{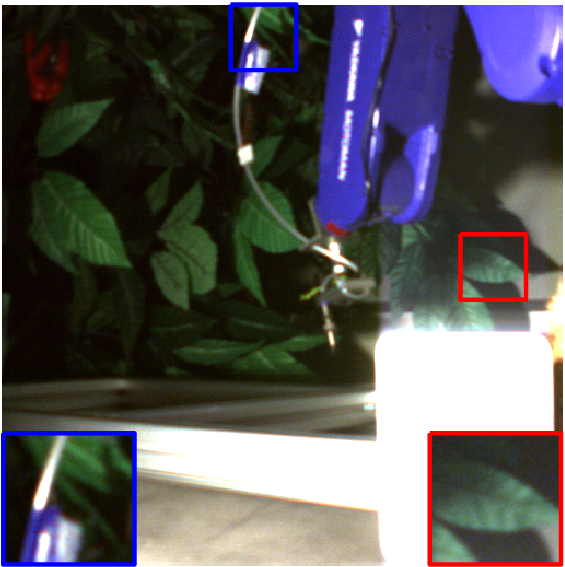}}
	\hspace{-2.1mm}
        \subfloat[]{\label{fig:with_wavelet}\includegraphics[width=0.23\linewidth]{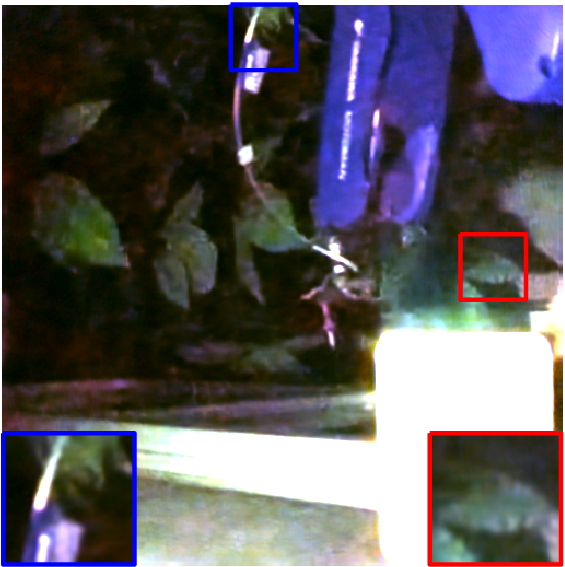}}
	\hspace{-2.1mm}
        \subfloat[]{\label{fig:without_wavelet}\includegraphics[width=0.23\linewidth]{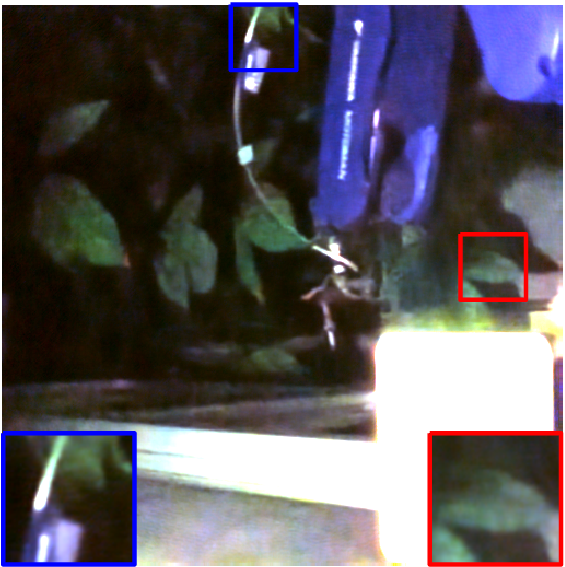}}
	\hspace{-2.1mm}
	\subfloat[]{\label{fig:wavelet_res}\includegraphics[width=0.23\linewidth]{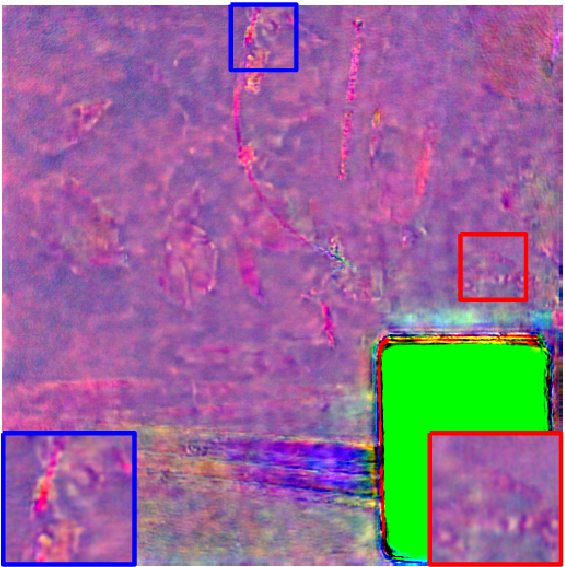}}
  \caption{ Visualization of the impact of wavelet transform on on the Master2900K from the ICVL Dataset under the non-i.i.d. Gaussian noise with $\sigma \in [0,95]$. (a) Clean. (b) With wavelet transform. (c) Without wavelet transform. (d) Residual of (b) and (c).} \label{fig:analysis_wavelet}
\end{figure}

To verify the advantages of combining wavelet transform, we conduct the experiment under the non-i.i.d. Gaussian noise with $\sigma \in [0,95]$. As can  be seen from Table~\ref{tab:wavelet}, with wavelet transform the PSNR improved by 0.15dB. Moreover, to more intuitively demonstrate the benefits of combining wavelet transform and RMM, we visualize the residual of the denoising results with and without wavelet transform in Fig.~\ref{fig:analysis_wavelet}. The results show that combining wavelet transform preserves the sharpness of object edges and enhances texture details such as leaf veins, validating the effectiveness.

\begin{table}[t]
        \caption{Denoising performance with/without Wavelet transform.}\label{tab:wavelet}
        \centering
        \begin{tabular}{c|c|c|c}
                \Xhline{1.2pt}
        \multirow{2}*{Wavelet Transform}&\multicolumn{3}{c}{Index}\\
        \cline{2-4}
        &\multicolumn{1}{c}{PSNR$\uparrow$}&\multicolumn{1}{c}{SSIM$\uparrow$}&\multicolumn{1}{c}{SAM$\downarrow$}\\
        \Xhline{1.2pt}  		
        $\times$     & \multicolumn{1}{c}{45.31} & \multicolumn{1}{c}{.9852} & \multicolumn{1}{c}{.0418} \\
        \hline
        $\checkmark$ & \multicolumn{1}{c}{\textbf{45.46}} & \multicolumn{1}{c}{\textbf{.9853}} & \multicolumn{1}{c}{\textbf{.0404}} \\
        \Xhline{1.2pt}
        \end{tabular}
\end{table}

\section{Conclusion}\label{conc}

The paper presents  an  iterative low-rank network based on a progressive refinement process and spectral low-rankness prior for effective HSI denoising. The learnable RMM combines the strengths of both data-driven and model-driven methods and flexibly learns thresholding parameters from data, allowing for adaptively capturing global spectral correlations across various scenarios. The iterative refinement process combines the intermediate results with noisy images adaptively to supplement details, followed by further denoising, resulting in denoised outcomes with richer texture details. Experiments on both synthetic and real-world noisy data demonstrate  the superior performance of ILRNet in noise removal and detail restoration. Ablation experiments further demonstrate the effectiveness of the RMM and iterative refinement process. In future work, we plan to integrate the nonlocal self-similarity prior into DNNs to further improve  the denoising  capabilities.

\appendices
\bibliographystyle{IEEEtran}
\bibliography{IEEEabrv,ilrnet}
\begin{IEEEbiography}[{\includegraphics[width=1in,height=1.25in,clip,keepaspectratio]{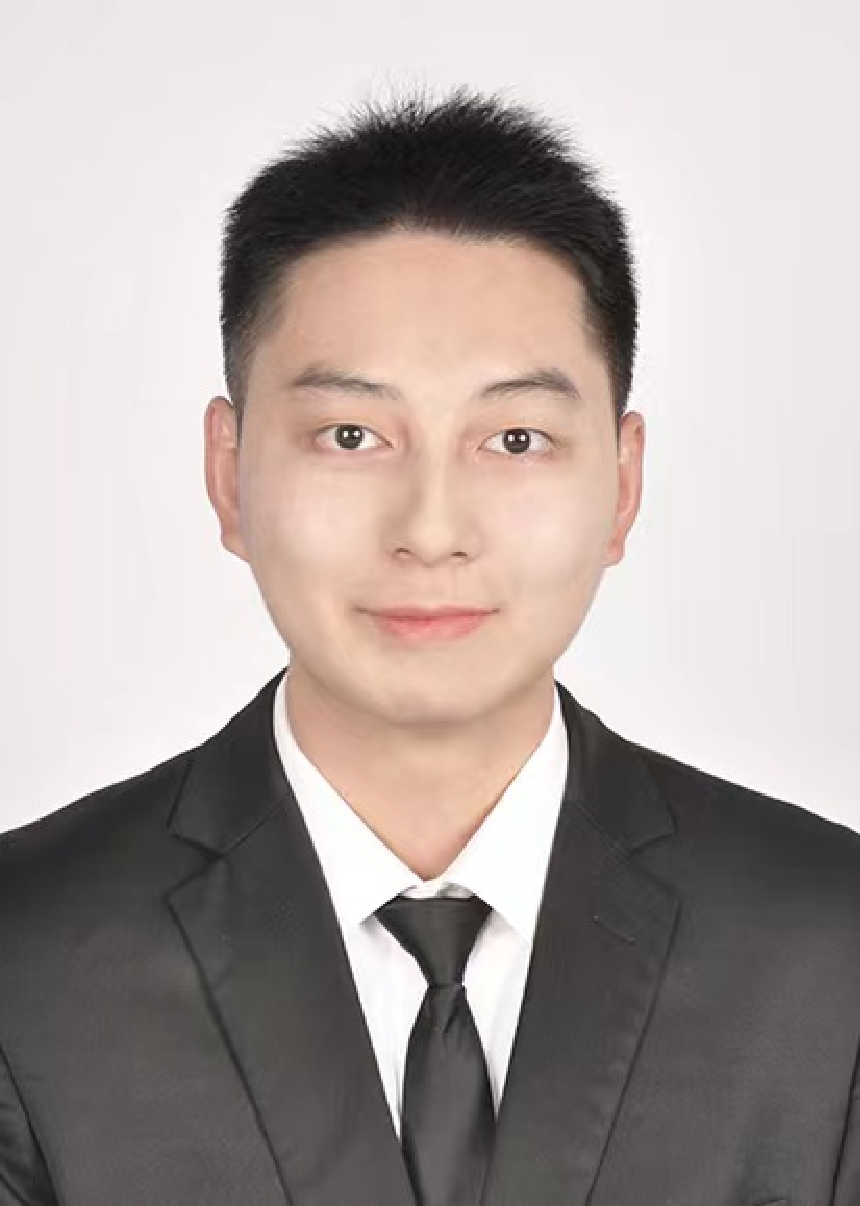}}]
{Jin Ye} rreceived the B.E. degree in software engineering from Nanjing University of Science and Technology, Nanjing, China, in 2023, where he is currently working towards the M.S. degree. His research interests include machine learning and hyperspectral low-level image processing.
\end{IEEEbiography}

		\begin{IEEEbiography}[{\includegraphics[width=1in,height=1.25in,clip,keepaspectratio]{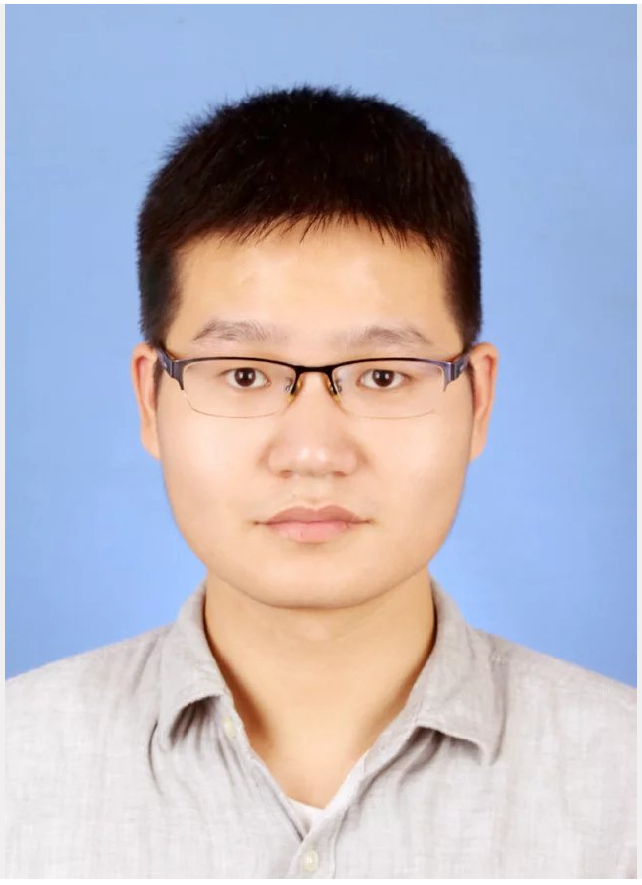}}]{Fengchao Xiong}(S'18-M'20) received the B.E. degree in software engineering from Shandong University, Jinan, China, in 2014, and the Ph.D. degree from the College of Computer Science, Zhejiang University, Hangzhou, China, in 2019.

He visited Wuhan University, Wuhan, China; Griffith University, Nathan, QLD, Australia; and the University of Macau, Taipa, Macau, China, in 2011-2012, 2017-2018, and 2021-2023, respectively. He is currently an Associate Professor with the School of Computer Science and Engineering, Nanjing University of Science and Technology, Nanjing, China. His research interests include hyperspectral image processing, machine learning, and pattern recognition.

Dr. Xiong serves as a Topical Associate Editor for IEEE Transactions on Geoscience and Remote Sensing.
\end{IEEEbiography}

\begin{IEEEbiography}[{\includegraphics[width=1in,height=1.25in,clip,keepaspectratio]{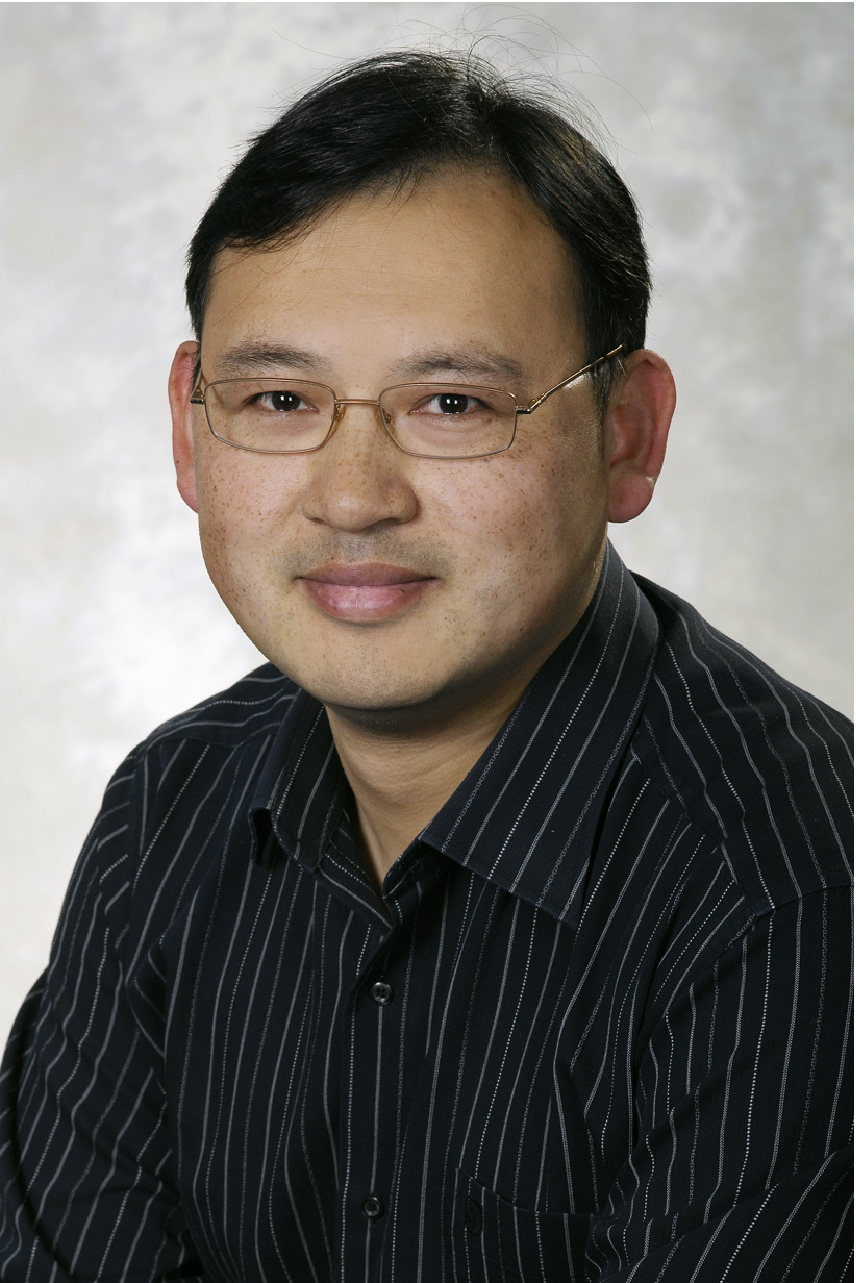}}]{Jun Zhou} (M'06-SM'12) received the B.S. degree in computer science and the B.E. degree in international business from Nanjing University of Science and Technology, Nanjing, China, in 1996 and 1998, respectively. He received the M.S. degree in computer science from Concordia University, Montreal, Canada, in 2002, and the Ph.D. degree from the University of Alberta, Edmonton, Canada, in 2006.

In June 2012, he joined the School of Information and Communication Technology, Griffith University, Nathan, Australia, where he is currently a Professor. Prior to this appointment, he was a Research Fellow with the Research School of Computer Science, Australian National University, Canberra, Australia, and a Researcher with the Canberra Research Laboratory, NICTA, Canberra. His research interests include pattern recognitions, computer vision, and spectral imaging with their applications in remote sensing and environmental informatics. He is an Associate Editor of IEEE Transactions on Geoscience and Remote Sensing and Pattern Recognition journal.
\end{IEEEbiography}

\vspace{-0.55cm}

\begin{IEEEbiography}[{\includegraphics[width=1in,height=1.25in,clip,keepaspectratio]{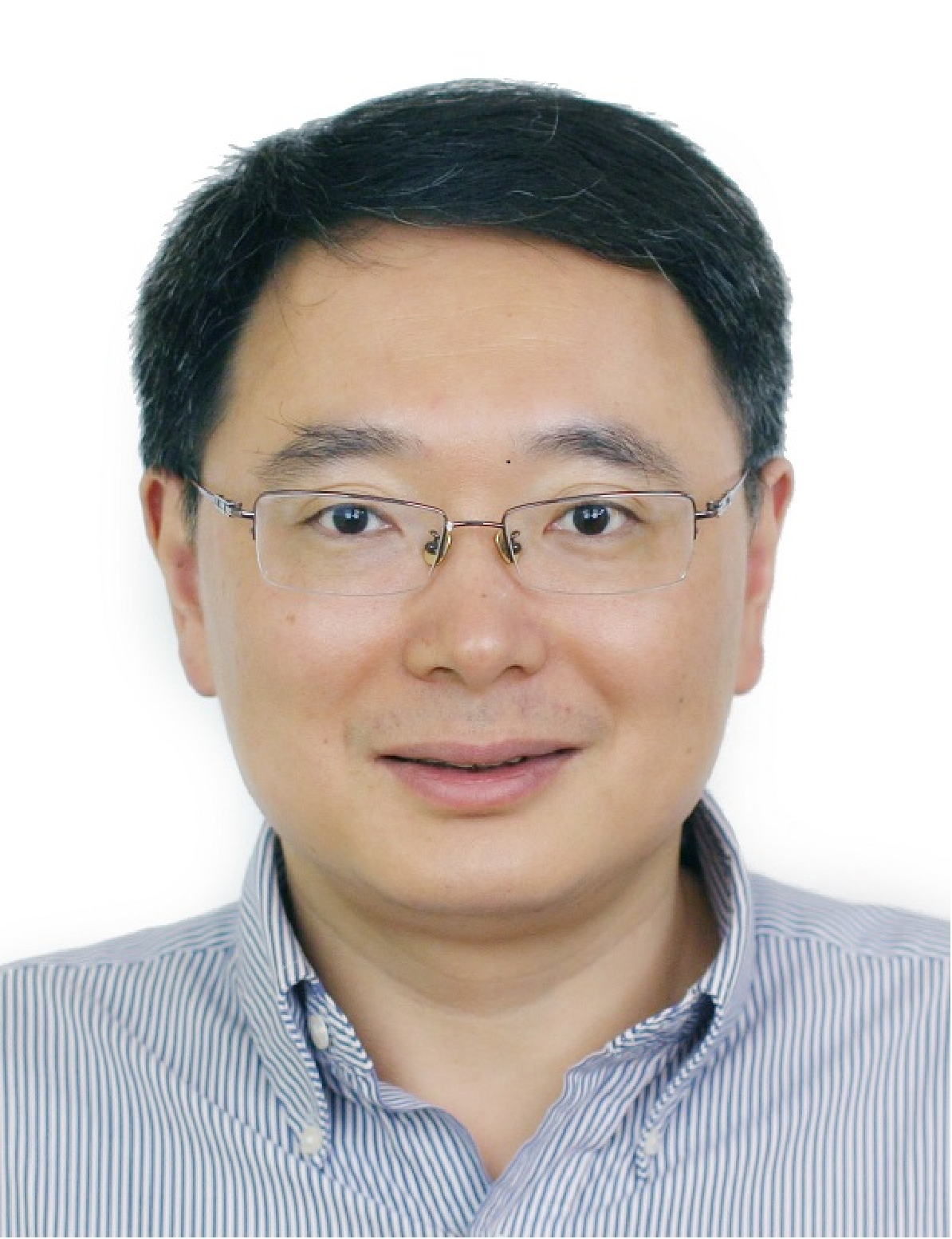}}]{Yuntao Qian} (M'04-SM'21) received the B.E. and M.E. degrees in automatic control from Xi'an Jiaotong University, Xi'an, China, in 1989 and 1992, respectively, and the Ph.D. degree in signal processing from Xidian University, Xi'an, China, in 1996.

During 1996--1998, he was a Postdoctoral Fellow with the Northwestern Polytechnical University, Xi'an, China. Since 1998, he has been with the College of Computer Science, Zhejiang University, Hangzhou, China, where he became a Professor in 2002. During 1999-2001, 2006, 2010, 2013, 2015-2016, and 2018 he was a Visiting Professor at Concordia University, Hong Kong Baptist University, Carnegie Mellon University, the Canberra Research Laboratory of NICTA, Macau University, and Griffith University. His current research interests include machine learning, signal and image processing, pattern recognition, and hyperspectral imaging. He is currently an Associate Editor of the \textsc{IEEE Journal of Selected Topics in Applied Earth Observations and Remote Sensing.}
\end{IEEEbiography}
\end{document}